\documentclass{MITcsail}
\usepackage{amsmath}

  \DeclareMathAlphabet\mathbfcal{OMS}{cmsy}{b}{n}

  \def\0{{\bf 0}}
  \def\1{{\bf 1}}

\def\eg{\emph{e.g.}} 
\def\ie{\emph{i.e.}} 
 
\def\etc{\emph{etc.}} \def\vs{\emph{vs.}}
 
\def\etal{\emph{et al.}}

\def\revision{\textcolor{black}}
\def\revise{\textcolor{black}}
\def\rtwo{\textcolor{black}} 
\def\rthree{\textcolor{black}} 

\usepackage{tikz}
\usepackage{booktabs}
\usepackage{multirow}
\usepackage{array}
\usepackage{tabularx}
\usepackage{makecell}

\usepackage[most]{tcolorbox}
\definecolor{LabGreen}{HTML}{14532D}
\definecolor{shadecolor}{RGB}{243,244,246}
\renewenvironment{abstract}{%
  \begin{tcolorbox}[%
    enhanced jigsaw,
    breakable,
    colback=shadecolor,
    opacityback=1.0,
    opacityfill=1.0,
    frame hidden,
    boxrule=0pt,
    boxsep=0pt,
    arc=0pt,
    outer arc=0pt,
    borderline west={2pt}{0pt}{LabGreen},
    left=1em,
    right=1em,
    top=1em,
    bottom=1em,
    before skip=12pt,
    after skip=0pt
  ]%
  \noindent{\large\sffamily\bfseries Abstract}\\[0.15cm]%
  \noindent\ignorespaces%
}{%
  \end{tcolorbox}%
  \vspace{0.3cm}%
}

\title{A Survey on Self-Improving Test-Time Intelligence: Feedback-Driven Adapting, Learning, and Scaling at Inference}
\shorttitle{Test-Time Intelligence: A Survey}

\author
{{\fontsize{10.5pt}{10pt}\selectfont Shuaicheng Niu$^{1,2,\star}$, Guohao Chen$^{1,2,\star}$, Yaofo Chen$^{1}$, Zhiquan Wen$^{1}$, Jinwu Hu$^{1}$, Zeshuai Deng$^{1}$, Deyu~Chen$^{1}$, Shuhai Zhang$^{1}$, Renjie Chen$^{1}$, Zihao Lian$^{1}$, Shoukai Xu$^{1}$, Gang Dai$^{1}$, Yunbei Zhang$^{3}$, \\ Wei Luo$^{4}$, Yifan Zhang$^{5}$, Mingkui Tan$^{1,4,\dagger}$, Cheng Deng$^{6,\dagger}$ } \\
\vspace{1em} 
\normalfont{\small $^{1}$South China University of Technology},~
\normalfont{\small $^{2}$Nanyang Technological University},~ 
\normalfont{\small $^{3}$Tulane University},~ \\
\normalfont{\small $^{4}$Pazhou Laboratory},~
\normalfont{\small $^{5}$National University of Singapore},~
\normalfont{\small $^{6}$Hohai University} \\
\normalsize{\small $^{\star}$Equal Contribution, $^{\dagger}$Corresponding Author}\\
}

\begin{document}

\maketitle
\thispagestyle{firstpagestyle}

\begin{abstract}
The ability of AI systems to improve their behavior during deployment is becoming increasingly important. As inference moves beyond the static execution of a fixed trained model, a growing body of work studies how models can refine their behavior on the fly by exploiting test-time information and additional computation. These developments have largely evolved along two directions: methods that modify the model’s state using test-time signals, and methods that \revision{improve predictions through extra inference-time resources such as more sampling and tool use.} However, these directions are often studied in separate communities with different terminology, making their connections harder to see. In this survey, we present \revision{feedback-driven \emph{Test-Time Intelligence} (TTI)} as a unified perspective for understanding such deployment-time improvement. We use this view to relate test-time adaptation, test-time learning, and test-time scaling, highlighting both their distinctions and their growing overlap in hybrid systems. This unified framework helps connect previously fragmented ideas and provides a clearer conceptual foundation for studying inference-time self-improvement. We review major methodological paradigms, representative applications, and open challenges across vision, language, multimodal learning, generative models, robotics, and healthcare. Our goal is to provide a coherent foundation and research roadmap for the study of self-improving AI systems at test time.
A collection of related resources is available at \url{https://github.com/mr-eggplant/awesome_test_time_intelligence}.
\end{abstract}

\tableofcontents

\section{Introduction}\label{sec:intro}

Artificial intelligence (AI) systems are traditionally developed under a train-then-infer paradigm~\cite{he2016deep}: a model is optimized during training, frozen after deployment, and then expected to perform reliably at inference time without further modification. Yet real-world intelligence is rarely so static: humans continue to adapt~\cite{cui2021toward}, draw on past experience, and allocate more effort to difficult situations while acting in the world. As AI systems are increasingly deployed in dynamic open world~\cite{yuan2023robust,niu2023sar,bashkirova2022visda,gao2024unified,lee2023towards} and personalized settings~\cite{karpowicz2025stabilizing,sun2026tard,tian2025continual}, the assumption that a fixed model can handle all test-time scenarios is becoming less tenable. In real deployment, models are often exposed to changing environments with potential distributional shifts~\cite{hendrycks2019benchmarking,koh2021wilds,SDV21acdc,wang2022continual}, user-specific demands~\cite{karpowicz2025stabilizing,shen2024speaker,gu2025metawriter}, and tasks whose difficulty varies substantially across instances~\cite{han2025token,jali2026not,zhang2025adaptthink}. As a result, good performance at test time can no longer always be achieved by simply executing a fixed model once and returning its output.

Instead, inference is increasingly becoming a dynamic process. A deployed system may need to exploit test-time feedback, update part of its internal state~\cite{wang2021tent,shu2022test}, retrieve external knowledge~\cite{asai2023self}, interact with tools~\cite{chen2023program,gou2024critic}, allocate more computation to difficult cases~\cite{han2025token,zhang2025adaptthink}, or refine its outputs through multi-step reasoning and verification~\cite{madaan2023self}, \etc~This shift is driving growing interest in AI systems that are capable of self-improvement during deployment, rather than remaining completely static after training.

\paragraph{What is Test-Time Intelligence?} One major route beyond static inference is \textbf{Test-Time Learning (TTL)}, introduced by Sun \etal~\cite{sun2020test}, which shows that models can improve at inference time by updating \revision{model states} using unlabeled test data. This idea has led to a large body of work on \textbf{Test-Time Adaptation (TTA)}~\cite{wang2021tent,niu2022eata,niu2023sar,lee2024deyo,wang2022continual,bartler2022mt3,liang2025comprehensive}, where models adapt through updating normalization layers~\cite{wang2021tent,hu2025beyond}, prompts~\cite{shu2022test,zhang2025dpcore}, low-rank adapters~\cite{imam2025test,zhangavoiding,liu2023vida}, inputs~\cite{gao2023back,yang2026diffusion}, or full parameters~\cite{sun2020test,liu2021ttt++,zhang2022memo,wang2022continual} with objectives such as entropy minimization~\cite{wang2021tent}, consistency regularization~\cite{niu2025spa}, or reconstruction-based supervision~\cite{gandelsman2022tmae,chen2025ttt3r}. Although much of this literature focuses on robustness under distribution shift, test-time learning is not restricted to robustness alone. It can more broadly serve as a general pipeline that uses test-time signals to improve prediction quality or task capability~\cite{niu2022cli,Hu2025TLM,hardt2023test,liao2026tool}.

\begin{figure}[t]
\centering
\includegraphics[width=0.8\linewidth]{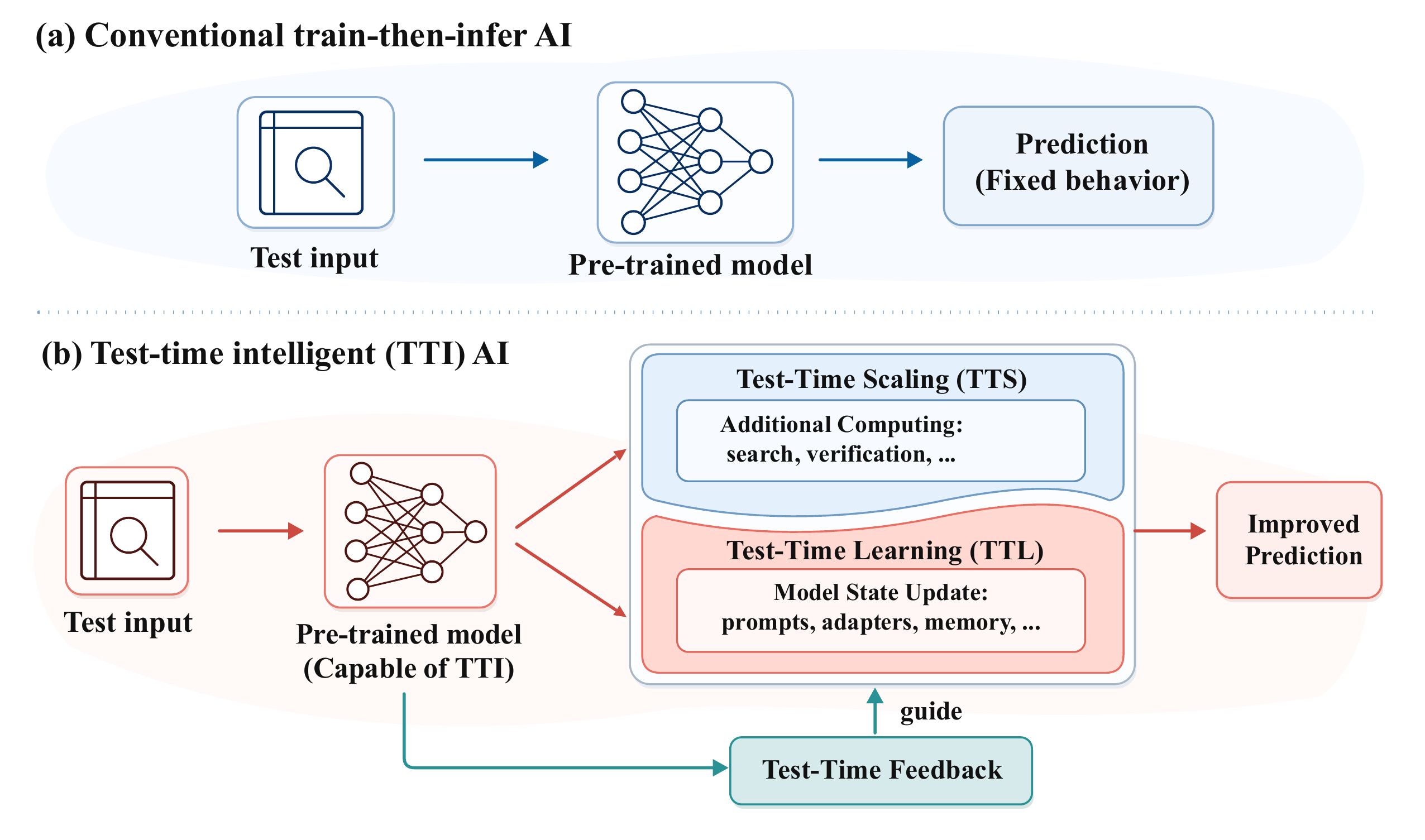}
\caption{\revision{Comparison between feedback-driven self-improving test-time intelligent AI and conventional static train-then-infer AI.}}
\label{fig:tti_vs_static_ai}
\end{figure}

Another major route is \textbf{Test-Time Scaling (TTS)}, where performance improves through additional inference-time computation without necessarily changing \revision{some model states}. Examples include \revision{repeated decoding~\cite{brown2024large}, self-consistency~\cite{wang2023self}, tool use~\cite{chen2023program,gou2024critic}, search-based inference~\cite{yao2023tree}, and inference-time optimization in generative models~\cite{ma2025inference}.} Similar ideas also appear in robotics and embodied systems, where planning and interaction can improve decision quality even without persistent model parameter updates~\cite{kwok2025robomonkey,kwok2026scaling}. These examples suggest that learning alone is a bit narrow to capture the full range of inference-time intelligence.

\begin{figure}[h!]
    \centering
    \makebox[\linewidth][c]{%
        \includegraphics[width=1.03\linewidth]{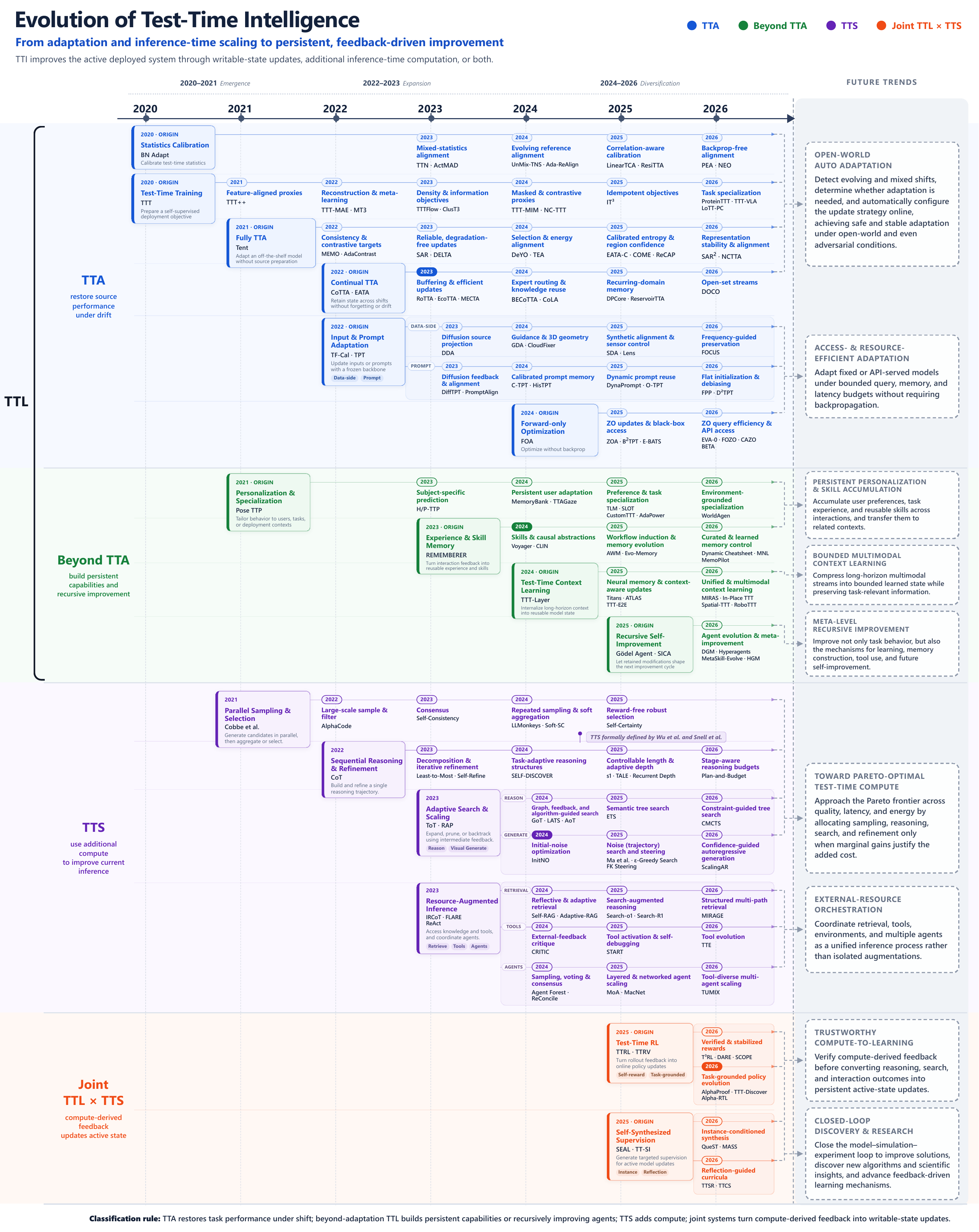}
    }
    \caption{Timeline of representative self-improving test-time intelligence methods.}
    \label{fig:roadmap}
\end{figure}

\begin{figure}[t]
    \centering
    \includegraphics[width=0.7\linewidth]{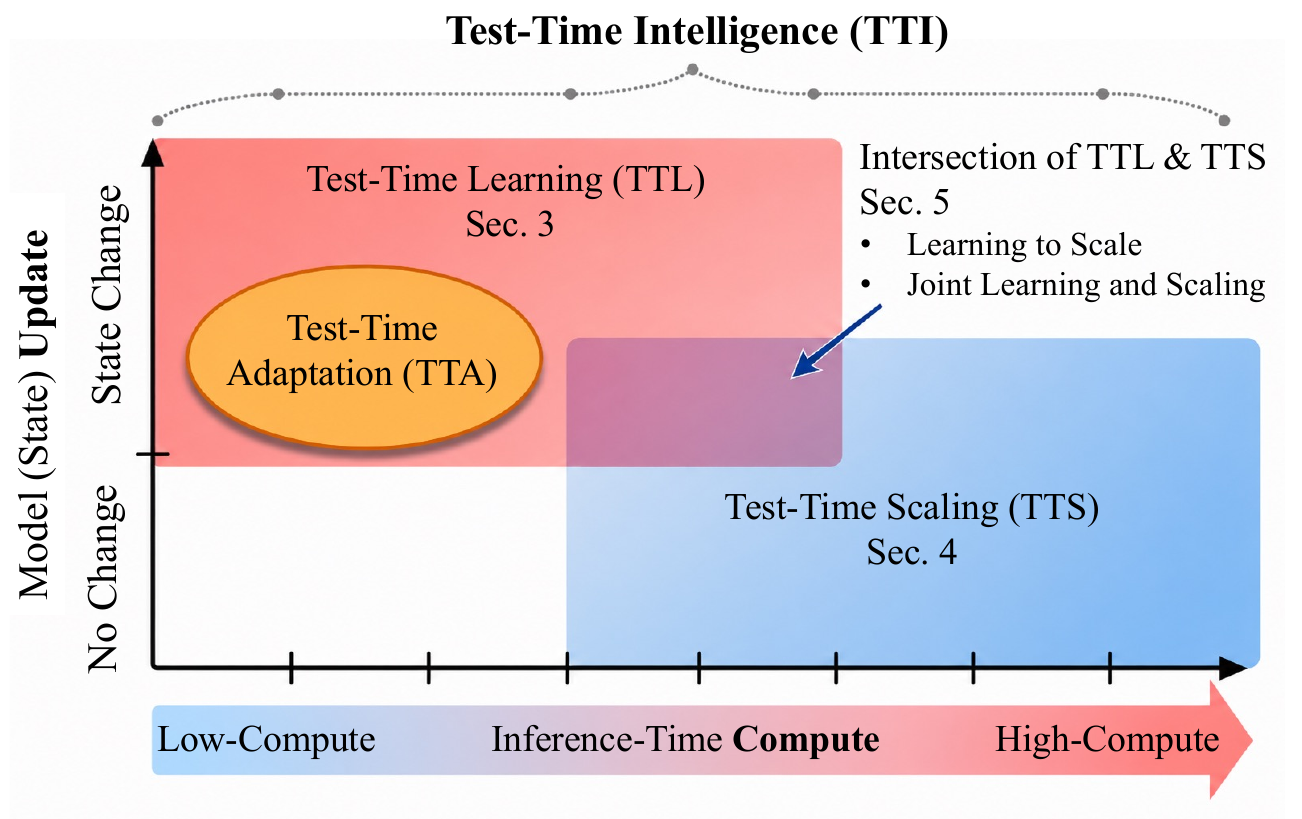}
    \caption{\revision{Taxonomy of feedback-driven self-improving test-time intelligence. Representative model states include norm layers, hidden states, prompts, adapters, and external memory.}}
    \label{fig:tti_taxonomy}
\end{figure}

This motivates a broader concept, which we call \revision{feedback-driven self-improving} \textbf{Test-Time Intelligence (TTI)}: the capability of a model or system to improve its behavior during deployment \revision{by leveraging test-time feedback for state updates or additional computation}, as shown in Fig.~\ref{fig:tti_vs_static_ai}. Under this view, test-time learning and test-time scaling are two major and complementary routes toward intelligent inference, \revision{as illustrated in Fig.~\ref{fig:tti_taxonomy} and detailed in Section~\ref{sec:definition}}. Learning emphasizes persistent or semi-persistent state change, \revision{while scaling emphasizes performance gains from additional inference-time resources such as search-based compute, often without state change.} In practice, these two routes also interact: a model may exploit scaling to produce reliable pseudo-labels for better learning, \textit{namely} Scaling for Learning (c.f. Section~\ref{sec:scale_for_learn}), or one can learn a policy to guide the reasoning model for when to scale, \textit{namely} Learning to Scale (c.f. Section~\ref{sec:learn_to_scale}). Therefore, \revision{TTI is best understood as a general paradigm for inference-time self-improvement through additional computing and/or model state updates based on test-time feedback signals.}

\textbf{Scope and Contributions} This survey presents a unified view of TTA, TTL, and TTS under the broader concept of TTI. The core contributions of this survey are: 
\textbf{1)} \revision{We introduce test-time model intelligence as a unifying conceptual framework for understanding how modern AI systems improve themselves during deployment through additional feedback with the joint roles of compute and state update.}
\textbf{2)} We clarify the conceptual relationship among adaptation, learning, and scaling: adaptation is best understood as a subset of test-time learning; scaling is not equivalent to learning, but overlaps with it in important ways; and some emerging systems lie precisely at their intersection.
\textbf{3)} We provide a comprehensive review of the key paradigms, representative applications, and open challenges in this rapidly growing area, with particular emphasis on the emerging shift from isolated adaptation or scaling techniques toward more general \revision{feedback-driven self-improving AI systems at test time.}

\textbf{Comparison with Prior Surveys}
Existing literature often studies related abilities under different names in separate communities with different assumptions and vocabularies, making it difficult to see their common structure and interaction. In this survey, we \revision{bring TTA, TTL, and TTS together} under a unified perspective that we call TTI. Compared to prior TTA surveys~\cite{liang2025comprehensive,wang2024search,xiao2024beyond} \revision{that are primarily organized around distribution shift and adaptation}, we broaden the scope from shift-centric adaptation to a more general perspective of deployment-time self-improvement. Compared to TTS surveys~\cite{zhang2025survey,chung2025revisiting}, which are often largely centered on LLMs and reasoning-time compute, our treatment is broader in both methodology and application scope. We position scaling as part of a larger landscape of test-time computation and self-improvement, encompassing both learning-free and learning-based mechanisms, as well as hybrid systems that combine the two.

This broader view allows us to connect ideas that are often studied separately across communities, including adaptation algorithms in machine learning, inference-time scaling strategies in foundation models, search and planning in embodied systems, and optimization-based test-time methods in generative models such as diffusion. We also review applications beyond language models, covering vision, multimodal learning, generative modeling, agentic AI, robotics, and healthcare. \revision{Across these areas, the update--compute view distinguishes whether improvement arises from state updates, additional inference-time computation, or both, and provides a common language for describing shared elements such as feedback signals. This places TTL and TTS within the same design space and helps reveal transferable mechanisms, hybrid design opportunities, and unresolved research gaps.}

\section{Conceptual Foundations}\label{sec:definition}

\subsection{Test-Time Intelligence}

\revision{We define Test-Time Intelligence (TTI) as the capability of an AI system to improve itself during deployment by exploiting test-time feedback. In this sense, inference is not merely a fixed execution process, but a dynamic stage in which the system can adapt, refine, or enhance its behavior while operating in the real world.} Under this view, as demonstrated in Fig. ~\ref{fig:tti_taxonomy}, there are two major and complementary routes toward intelligent inference, \ie, test-time learning (TTL) and test-time scaling (TTS).

\textbf{TTL and TTS: Distinction}
The core difference between TTL and TTS lies in whether performance improvement arises from \revision{changing the model state, such as norm layers, hidden states, prompts, adapters, and memory. TTL improves predictions by acquiring information/feedback from test data into the persistent or semi-persistent model state changes through explicit learning.} Test-time adaptation \textit{(TTA) falls within this paradigm as a special case of TTL}, primarily concerned with robustness under distribution shift. TTS, by contrast, improves predictions by \revision{spending more compute on search, verification, or tool-use resources during inference,} typically without leaving lasting changes to the model once the current inference ends.

\textbf{TTL and TTS: Interplay}
The boundary between the two, however, is not strict. Scaling and learning \revision{may} interact in both directions. On the one hand, learning can enable more effective scaling: a model may learn when to reason longer, how much computation to allocate, or which tools to invoke. On the other hand, \rthree{scaling can enable learning by providing stronger supervision signals through consensus or verification, which can then support self-training or policy updates.}  This interplay gives rise to a rich design space of hybrid systems that combine persistent self-improvement with dynamic inference-time computation. Accordingly, a central argument of this survey is that TTL and TTS should not be studied as isolated threads. Rather, they are best understood as two fundamental and interacting axes within the \revision{broader paradigm of feedback-driven self-improving test-time intelligence}.

\textbf{\revision{TTI and Training-Time Amortization}}
\revision{TTI and training-time amortization differ in when the resulting improvement becomes available. TTI therefore denotes
improvements realized within the active deployment process---either immediately for the current instance or cumulatively across the episode or deployment stream---through online state updates, additional inference-time computation, or both. By contrast, some pipelines retain deployment-time signals for a later offline learning pipeline, such as fine-tuning, distillation, or training-data construction, rather than using them to improve the active
system. When the improvement becomes available only through this separate, offline pipeline, we attribute it to training-time amortization rather than TTI.}

\textbf{\rtwo{TTL and Ordinary Inference-Time State Evolution}}
\rtwo{The key boundary is whether test-time information triggers a feedback-driven update to a designated writable state. TTL does not include ordinary transient states that naturally evolve during inference, such as KV caches in autoregressive decoding, hidden states in sequential models, or diffusion latents along a predefined denoising trajectory. Although such states may affect the current output, they are part of the standard inference procedure and do not constitute TTL unless they are explicitly optimized, written to memory, or otherwise updated using test-time feedback. In this survey, we reserve TTL for cases where test-time data or external feedback modifies a writable state, such as parameters or memory, and the modified state is reused by the active deployed system.}

\subsection{\revision{A Unified View: Update and Compute}}

To unify the diverse methods in this area, we organize TTI around the following \revision{two} fundamental ingredients:
\begin{itemize}
    \item \revision{\textbf{Update} refers to whether a system changes its state with test-time feedback. This may involve updating internal parameters, tuning prompts, adjusting normalization statistics, optimizing latent variables, or writing to memory. Updating is the core mechanism underlying TTL, whereas most TTS methods do not modify the model state. Nevertheless, some TTS methods maintain and update external memory, thereby lying at the intersection of TTL and TTS, as indicated by the shaded area in Fig.~\ref{fig:tti_taxonomy}.}
    \item \revision{\textbf{Compute} refers to the amount and structure of inference-time computation devoted to solving a test instance. Prior TTS studies have primarily focused on large models, such as LLMs and MLLMs, and improve performance by allocating additional computation through repeated decoding, extended reasoning, search, verification, or multi-agent interaction. Compute is therefore the key mechanism underlying TTS. In contrast, TTL has mainly been studied on relatively lightweight models, such as visual recognition models, and typically requires less overall computation. Nevertheless, some methods combine additional inference-time computation with model-state updates and thus lie in the intermediate region of the compute axis.}
\end{itemize}

\revision{This view helps place different methods in a common space. As illustrated in Fig.~\ref{fig:tti_taxonomy}, classical TTL methods primarily rely on state updates, typically with limited extra overall computation, and therefore occupy the upper-left region. In contrast, TTS methods primarily rely on additional inference-time computation, usually with no persistent state update, and thus occupy the lower-right region. Hybrid methods combine both mechanisms, for example, by using search or self-consistency to produce better learning signals, or by learning when and how to allocate additional inference-time resources. These methods lie in the shaded intersection region. From this perspective, TTI arises not from learning or scaling alone, but from different ways of coordinating state updates and inference-time computation.}

\section{Test-Time Learning}\label{sec3}

Test-time learning (TTL) refers to any paradigm in which a deployed model \revision{updates writable state using information derived from test inputs and, when available, external feedback, and uses the updated state for inference within the same deployment process.} The central motivation draws from the ubiquity of distribution shift: real-world data encountered at deployment frequently departs from the distribution seen during training due to sensor changes, environmental variation, domain differences, or temporal drift, and a model frozen at training time inevitably
degrades~\cite{liang2025comprehensive,wang2022continual}. Beyond adaptation to distribution shifts, there is also an emerging trend that extends the scope of TTL for memorization, personalization, self-improvement, \etc~We depict them below.

\subsection{Problem Settings and Design Dimensions}
\label{sec:problem-settings}

Let $p(x, y)$ define a joint distribution on the input-output space $\mathcal{X} \times \mathcal{Y}$ and $\mathcal{D}_{\mathrm{tr}} = \{(x_i, y_i)\}_{i=1}^{n}$ be a labeled training dataset drawn from a source distribution $p_{\mathcal{S}}(x, y)$. 
The training phase produces a model $f_{\theta}: \mathcal{X} \to \mathcal{Y}$ with parameters $\theta^* = \arg\min_{\theta} \, \mathcal{L}_{\mathrm{tr}}(\theta; \mathcal{D}_{\mathrm{tr}})$, where $\mathcal{L}_{\mathrm{tr}}$ is a supervised objective (e.g., cross-entropy), and $\theta^*$ is fixed under the traditional deployment assumption.

\textbf{Definition 1 (Test-Time Learning)} Let $\mathcal{D}_{\mathrm{te}}=\{x_j\}_{j=1}^{m}$ denote the test inputs encountered during deployment, and let $\mathcal{F}_{\mathrm{te}}$ denote any external feedback received at test time (\eg, from human users, tools, or the environment), which may be empty. \revision{Test-time learning (TTL) refers to any paradigm in which the active deployed model updates writable state $\phi$ using information derived from $\mathcal{D}_{\mathrm{te}}$ and, when available, external feedback $\mathcal{F}_{\mathrm{te}}$, and uses the updated state for inference within the same deployment process. The writable state $\phi$ may include the input or its latent representation, model parameters or statistics, auxiliary modules, and writable external memory such as caches or prototype banks.} Formally, TTL induces a deployment-time state transition:
\begin{equation}\label{eq:ttl_overall}
    \phi^{+} = \operatorname{Update}\!\left(\phi;\, \mathcal{D}_{\mathrm{te}}, \mathcal{F}_{\mathrm{te}}\right)
\end{equation}
\revision{Here, $\operatorname{Update}$ may represent optimization of a test-time objective $\mathcal{L}_{\mathrm{ttl}}$, a closed-form or running-statistics update, a memory insertion/replacement, or input/latent refinement.} The settings of TTL cover both: 1) $p_{\mathcal{S}} \neq p_{\mathcal{T}}$, where TTL focuses on addressing distribution shifts; and 2) $p_{\mathcal{S}} = p_{\mathcal{T}}$, which extends TTL to adaptive memorization and self-improvement.
\revise{According to Eqn.~\ref{eq:ttl_overall}, the design space of TTL methods can be organized along four design dimensions:}

\begin{enumerate}
    \item \textbf{Feedback signal} (Section~\ref{sec:feedback-signals}): \revision{the information that drives the update, ranging from entropy, consistency, and pseudo-labels to feedback from human users, tools, or the environment.}
    \item  \textbf{Update target $\phi$} (Section~\ref{sec:what-updated}): \revise{the writable state that is modified, ranging from inputs and normalization statistics to auxiliary modules, full parameter sets, and external memory.}
    \item \textbf{Temporal horizon} (Section~\ref{sec:temporal-horizons}): the granularity at which updates occur, from single-sample to batch adaptation at the data level, and from episodic to continual learning across a deployment lifetime at the model level.
    \item \textbf{Objective scope} (Sections~\ref{sec:tta_adaptation}--\ref{sec:beyond_adaptation}): whether the primary goal is distributional alignment, memory augmentation, personalization, or iterative self-improvement.
\end{enumerate}

\revise{This provides a \textit{multi-view description} of the TTL design space. Specifically, we describe each method along the four axes: which feedback it uses, which state it updates, how information is accumulated over time, and which objective it serves. A compound method may therefore appear under multiple second-level headings, while the third-level categories specify its role within each axis. For example, pseudo-labels are classified as feedback even when used in continual, source-free, or reinforcement-learning settings.}
\rthree{For clarity, Sections~\ref{sec:feedback-signals}--\ref{sec:temporal-horizons} describe methods along their feedback, update-target, and temporal attributes, whereas Sections~\ref{sec:tta_adaptation}--\ref{sec:beyond_adaptation} situate them according to their primary objectives. When a method spans multiple dimensions, we discuss its mechanism where it is most informative and use cross-references elsewhere to clarify its other roles.}

\subsection{Feedback Signals for Test-Time Learning}
\label{sec:feedback-signals}

\revise{The choice of feedback signal---the internally derived or externally provided information that drives writable-state updates---is arguably the most consequential design decision in any TTL system. When ground-truth labels are unavailable, a TTL method must construct a surrogate objective that correlates with downstream task performance using the available test-time information.}

\revise{The five signal families surveyed below make different assumptions about what constitutes reliable supervision at deployment. \textbf{Entropy and confidence objectives} require the source model to remain sufficiently calibrated on target inputs; they are inexpensive when applied to a small parameter subset, but can reinforce confident mistakes or collapse under severe or label-distribution shifts. \textbf{Consistency and augmentation} trade additional forward passes and a validity assumption on the chosen transformations for a less direct dependence on any single prediction. \textbf{Reconstruction and other self-supervised signals} provide dense, input-grounded feedback, yet often require a training-prepared auxiliary objective or decoder and may optimize structure that is only weakly related to the downstream task. \textbf{Pseudo-label and self-training methods} exploit class structure and historical neighbors, but their gains depend on reliable initial labels and can be reversed by confirmation bias. \textbf{External tool, environment, or user feedback} can be more task-aligned, while requiring interfaces, latency budgets, permissions, or human effort. Consequently, lightweight internal signals are preferable for modest shifts and autonomous low-cost deployment, whereas external feedback is preferable when errors are consequential and reliable feedback is available.}

\begin{table}[t!]
\centering
\caption{Taxonomy of feedback signals for test-time learning.}
\label{tab:feedback-signals}
\renewcommand{\arraystretch}{1.0}
\tiny
\setlength{\tabcolsep}{2pt}
\resizebox{1.0\textwidth}{!}{%
\begin{tabular}{@{} p{1.8cm} p{2.7cm} p{3.6cm} p{2.8cm} p{2.8cm} p{2.8cm} @{}}
\toprule
\textbf{Signal}\newline\textbf{Family} & \textbf{Strategy} & \textbf{Representative Works} & \revise{\textbf{Required Feedback/}}\newline\revise{\textbf{Resources}} & \revise{\textbf{Applicable}}\newline\revise{\textbf{Scenarios}} & \revise{\textbf{Failure Modes}} \\
\midrule
\multirow{5}{*}{\parbox{1.8cm}{Entropy \&\\Confidence}}
  & entropy minimization          & Tent~\cite{wang2021tent}; EATA~\cite{niu2022eata} & \multirow{5}{*}{\parbox{2.8cm}{\revise{Unlabeled predictions, confidence scores, and gradient access to the selected state}}} & \multirow{5}{*}{\parbox{2.8cm}{\revise{Modest covariate shifts with a reasonably calibrated source model}}} & \multirow{5}{*}{\parbox{2.8cm}{\revise{Confirmation bias, overconfidence, class collapse, and sensitivity to label shift}}} \\
  & collapse prevention           & SAR~\cite{niu2023sar}; COME~\cite{zhang2025come}; DeYO~\cite{lee2024deyo}; ReCAP~\cite{hu2025beyond} & & & \\
  & non-saturating confidence     & TEA~\cite{yuan2024tea}; LSCD-TTA~\cite{liang2025low} & & & \\
  & confidence gating             & SoTTA~\cite{gong2023sotta}; RoTTA~\cite{yuan2023robust} & & & \\
  & feature regularization        & SAR$^2$~\cite{niu2025adapt}; NCTTA~\cite{chen2026neural} & & & \\
\midrule
\multirow{4}{*}{\parbox{1.8cm}{Consistency \&\\Augmentation}}
  & augmentation consistency      & MEMO~\cite{zhang2022memo}; CoTTA~\cite{wang2022continual} & \multirow{4}{*}{\parbox{2.8cm}{\revise{Semantics-preserving augmentations, multiple views or model copies, and agreement objectives}}} & \multirow{4}{*}{\parbox{2.8cm}{\revise{Inputs with known invariances or streams where teacher--student agreement is meaningful}}} & \multirow{4}{*}{\parbox{2.8cm}{\revise{Invalid transformations, correlated model errors, teacher drift, or trivial agreement}}} \\
  & \revise{model/branch consistency}       & RMT~\cite{dobler2023robust}; ZeroSiam~\cite{chenzerosiam} & & & \\
  & self-bootstrapping            & SPA~\cite{niu2025spa}; EATA-C~\cite{tan2025uncertainty} & & & \\
  & feature statistics alignment  & ActMAD~\cite{mirza2023actmad}; Ada-ReAlign~\cite{zhang2024test} & & & \\
\midrule
\multirow{5}{*}{\parbox{1.8cm}{Reconstruction \&\\Self-Supervision}}
  & autoencoder reconstruction    & TTA-AE~\cite{he2021autoencoder}; TTA-DAE~\cite{karani2021test} & \multirow{5}{*}{\parbox{2.8cm}{\revise{Input structure plus a reconstruction, masking, contrastive, or discriminative pretext objective}}} & \multirow{4}{*}{\parbox{2.8cm}{\revise{Large shifts where intrinsic structure remains informative and an auxiliary task is available}}} & \multirow{5}{*}{\parbox{2.8cm}{\revise{Pretext--task mismatch, shortcut reconstruction, or auxiliary-head domain bias}}} \\
  & masked patch prediction       & TTT-MAE~\cite{gandelsman2022tmae}; \revise{Continual-MAE~\cite{liu2024continual}} & & & \\
  & transformation pretexts
& TTT~\cite{sun2020test}; IT$^3$~\cite{durasov20243} & & & \\

& contrastive pretexts
& TTT++~\cite{liu2021ttt++}; Rec-TTA~\cite{colussi2025rec} & & & \\

& discriminative pretexts
& NC-TTT~\cite{osowiechi2024nc}; TTTFlow~\cite{osowiechi2023tttflow} & & & \\

\midrule
\multirow{4}{*}{\parbox{1.8cm}{Pseudo-Label \&\\Self-Training}}
  & model prediction as target    & Goyal \etal~\cite{goyal2022test}; ECL~\cite{zeng2024rethinking} & \multirow{4}{*}{\parbox{2.8cm}{\revise{Predicted labels, prototypes, neighbors, consensus samples, or an external teacher}}} & \multirow{4}{*}{\parbox{2.8cm}{\revise{Class-structured tasks with reliable initial predictions or repeated related samples}}} & \multirow{4}{*}{\parbox{2.8cm}{\revise{Noisy-label reinforcement, prototype contamination, imbalance, and stale teachers}}} \\
  & prototypes \& neighbors       & T3A~\cite{iwasawa2021test}; AdaContrast~\cite{chen2022contrastive} & & & \\
  & consensus \& majority voting   & MEMO~\cite{zhang2022memo}; MM-TTA ~\cite{shin2022mm} & & & \\
  & external-model pseudo-labels   & TTT-KD~\cite{weijler2025ttt}; T$^2$ARD~\cite{gong2025cross} & & & \\
\midrule
\multirow{3}{*}{\parbox{1.8cm}{External Feedback\\(Tools/Env/\\User)}}
  & \revise{environment interaction}           & PAD~\cite{hansen2021self}; FeedTTA~\cite{kim2025test} & \multirow{3}{*}{\parbox{2.8cm}{\revise{Environment interaction, executable tools, evaluators, or human/user responses}}} & \multirow{3}{*}{\parbox{2.8cm}{\revise{Interactive, verifiable, or high-stakes tasks where task-aligned feedback is available}}} & \multirow{3}{*}{\parbox{2.8cm}{\revise{Reward hacking, tool failure, delayed/sparse feedback, annotation bias, unsafe explore}}} \\
  & tool-augmented feedback       & RLCF~\cite{zhaotest}; Reward-Adaptation~\cite{song2025reward} & & & \\
  & human-in-the-loop             & SimATTA~\cite{gui2024active}; COPR~\cite{zhang2025copr} & & & \\
\bottomrule
\end{tabular}%
}
\end{table}

\subsubsection{Entropy and Confidence-Based Signals}
\label{sec:entropy}

\revise{Prediction sharpness provides a widely used signal for test-time adaptation. We distinguish entropy minimization as a direct adaptation signal from alternative confidence estimates used for filtering or gating updates.}

\textbf{\textit{Entropy as an Adaptation Signal.}}
\revise{Entropy-based methods directly minimize the Shannon entropy of predictions on unlabeled test inputs. Tent~\cite{wang2021tent} updates only normalization-layer parameters, while subsequent work improves the reliability and stability of entropy-driven adaptation. This baseline is simple and inexpensive, but assumes that confident target predictions are usually correct.}
\revise{One line improves gradient reliability through \textit{sample selection and reweighting}: EATA~\cite{niu2022eata} filters unreliable and redundant samples; SAR~\cite{niu2023sar} combines entropy-based filtering with sharpness-aware updates; DeYO~\cite{lee2024deyo} reweights samples using a shape-based PLPD score, and \rthree{DELTA~\cite{zhao2023delta} dynamically reweights online updates to reduce their bias toward dominant classes}. Marsden \etal~\cite{marsden2024universal} further use weight ensembling and diversity weighting. This strategy is preferable in noisy streams, although its effectiveness depends on reliable selection scores.}
\revise{Another line addresses \textit{uncertainty calibration and collapse prevention}: SHOT~\cite{liang2020we} combines per-sample entropy minimization with batch-level marginal entropy maximization; COME~\cite{zhang2025come} uses a Dirichlet prior as a collapse-resistant surrogate; EATA-C~\cite{tan2025uncertainty} combines entropy optimization with model uncertainty estimation and calibration; and ZeroSiam~\cite{chenzerosiam} uses asymmetric stop-gradient optimization. These safeguards are most useful when naive entropy minimization risks collapse, but require additional calibration, diversity, or asymmetry constraints.}
\revise{Feature-level regularization further limits representation degradation: SAR²~\cite{niu2025adapt} regularizes feature redundancy and class inequity over a centroid bank, while NCTTA~\cite{chen2026neural} aligns features and classifiers based on the neural collapse phenomenon. This strategy helps preserve representation structure, but relies on suitable feature-level priors.}

\textbf{\textit{\revise{Alternative Confidence Signals for Adaptation.}}}
\revise{Beyond direct entropy minimization, methods also use alternative confidence signals either as adaptation objectives or as reliability scores for gating updates.}
\revise{Likelihood-ratio objectives avoid gradient saturation as predictions become increasingly peaked: Mummadi \etal~\cite{mummadi2021test} and LSCD-TTA~\cite{liang2025low} use such losses to maintain stable gradients at high confidence. TEA~\cite{yuan2024tea} instead applies contrastive divergence to a free-energy objective, while ReCAP~\cite{hu2025beyond} replaces sample-wise entropy with a region-level confidence proxy. These objectives retain informative gradients at high confidence, but their effectiveness depends on the reliability of the chosen confidence proxy.}
\revise{A complementary line uses confidence scores as a \textit{gating or filtering mechanism}, selecting reliable samples without changing the underlying adaptation objective.} SoTTA~\cite{gong2023sotta} and RoTTA~\cite{yuan2023robust} gate memory-bank admission via MSP confidence and uncertainty scores, respectively, while OSTTA~\cite{lee2023towards} further filters out OOD samples via post-adaptation confidence drop before gradient updates. \revise{This strategy is lightweight and broadly compatible with existing objectives, but may discard useful hard samples when confidence is poorly calibrated.}

\subsubsection{Consistency and Augmentation-Based Signals}
\label{sec:consistency}
\revise{Consistency signals encourage agreement either across target views or model branches, or between test-time features and retained reference statistics. We therefore organize this subsection into prediction-level consistency and feature-level alignment.}

\textbf{\textit{Prediction Consistency as a Signal.}}
\revise{At the input level, \textit{augmentation consistency} assumes that semantics-preserving transformations should yield stable predictions. MEMO~\cite{zhang2022memo} minimizes the entropy of predictions marginalized over augmented views, while TPT~\cite{shu2022test} applies the same objective to prompt tuning. CoTTA~\cite{wang2022continual} uses averaged teacher--student predictions over augmented views as pseudo-targets, whereas MIC~\cite{sojka2025adaptive} minimizes the discrepancy between masked and unmasked predictions. This strategy is most useful when reliable semantics-preserving transformations are available, at the cost of additional forward passes.}
\revise{Beyond input transformations, \textit{model/branch consistency} enforces agreement across models or network branches. RMT~\cite{dobler2023robust} aligns an EMA teacher and student via symmetric cross-entropy, while ZeroSiam~\cite{chenzerosiam} aligns a predictor branch with a stop-gradient branch through asymmetric divergence. This strategy avoids relying on hand-designed input transformations, but requires stable model or branch targets.}
\revise{A related strategy, \textit{full-to-part consistency}, provides a self-bootstrapping signal by comparing a full-information prediction with a partial-information counterpart. SPA~\cite{niu2025spa} aligns predictions from the original input and a Fourier-domain deteriorated view, whereas EATA-C~\cite{tan2025uncertainty} measures model uncertainty through consistency between the full network and stochastic-depth sub-networks. REM~\cite{han2025rem} instead uses progressive masking to align prediction distributions across difficulty levels while preserving their entropy ranking. This strategy is effective when partial views preserve task-relevant information, but can become unreliable when they remove discriminative evidence.}

\textbf{\textit{Feature Consistency as a Signal.}}
\revise{Beyond prediction-level agreement, consistency can also be imposed on intermediate features by aligning test-time activations with retained source statistics, including under evolving streams.}
\revise{\textit{Source-statistics alignment} matches test activations to references retained from source training. TTT+~\cite{liu2021ttt++}, ActMAD~\cite{mirza2023actmad}, and ViTTA~\cite{lin2023video} align global statistics through moment matching, per-layer activation alignment, and EMA, respectively, while TTN~\cite{lim2023ttn} interpolates source and test BN statistics according to shift sensitivity. Class-aware and covariance-level methods include CAFe~\cite{adachi2023covariance} and CAFA~\cite{jung2023cafa}, while DA-TTA~\cite{wang2024distribution} and ResiTTA~\cite{zhou2024resilient} perform layer-wise or soft BN alignment. FOA~\cite{niu2024foa} instead shifts activations without backpropagation. These methods are effective when retained source references remain representative, but can be brittle under large or evolving shifts. For non-stationary streams, \textit{evolving-reference alignment} is represented by Ada-ReAlign~\cite{zhang2024test}, which aligns evolving target representations to a source sketch using adaptively combined base learners. Evolving references better track non-stationary streams, although errors in the online reference may accumulate over time.}

\subsubsection{Reconstruction and Self-Supervision Signals}
\label{sec:reconstruction}
\revise{This family derives test-time objectives from the structure of unlabeled inputs, often using auxiliary heads, transformations, or decoders prepared during training. These signals either reconstruct observed or masked content or apply other pretext objectives to guide adaptation.}

\textbf{\textit{Reconstruction Error as a Signal.}}
\revise{\textit{Full-input reconstruction} directly provides the adaptation signal. TTA-AE~\cite{he2021autoencoder} combines pixel-level and multi-level feature reconstruction to update adaptor parameters, while TTA-DAE~\cite{karani2021test} and DeTTA~\cite{wen2024denoising} use denoising residuals to capture corruption-sensitive structure. It provides dense input-level feedback, but depends on reconstruction error remaining aligned with downstream performance.}
\revise{Beyond full-input reconstruction, \textit{masked prediction} adapts the encoder by reconstructing missing content. TTT-MAE~\cite{gandelsman2022tmae} and TTT-MIM~\cite{mansour2024ttt} use masked-image objectives with different masking and reconstruction designs, while Hybrid-TTA~\cite{park2025hybrid} couples masked reconstruction with the primary segmentation objective. Continual-MAE~\cite{liu2024continual} further uses distribution-aware masking and feature reconstruction for continual adaptation. Masked prediction also extends beyond images: MATE~\cite{mirza2023mate} reconstructs masked 3D point-cloud tokens, Wang \etal~\cite{wang2025test} apply per-frame MAE to video streams, and T4P~\cite{park2024t4p} masks temporal trajectory tokens under driving shifts. This strategy is preferable when missing-content recovery captures transferable structure, although it often requires a training-prepared auxiliary objective or decoder.}

\textbf{\textit{Other Pretext Tasks as Signals.}}
\revise{Beyond reconstructing observed or masked content, \textit{other pretext objectives} derive test-time gradients from transformations, invariances, or learned relations. TTT~\cite{sun2020test} uses four-way rotation prediction; IT$^3$~\cite{durasov20243} enforces test-time idempotence; TLM~\cite{Hu2025TLM} applies next-token prediction to input prompts; and Clust3~\cite{hakim2023clust3} maximizes mutual information through per-layer projectors. These objectives are useful when a suitable pretext task is available, but their gains depend on its correlation with the primary task.}
\revise{\textit{Contrastive and discriminative objectives} provide representation-level feedback: TTT++~\cite{liu2021ttt++} combines SimCLR with source-statistics alignment, AdaContrast~\cite{chen2022contrastive} combines MoCo-style learning with pseudo-labeling, and MT3~\cite{bartler2022mt3} meta-learns a BYOL-style objective. Rec-TTA~\cite{colussi2025rec} combines contrastive learning with feature reconstruction, while NC-TTT~\cite{osowiechi2024nc} trains a noise-contrastive discriminator over feature views. These objectives can better preserve discriminative feature structure, but typically require multiple views or training-prepared components. A \textit{likelihood-based alternative}, TTTFlow~\cite{osowiechi2023tttflow}, adapts the encoder by maximizing likelihood under a fixed normalizing-flow head. This alternative provides a direct density signal, although its reliability depends on the fit of the fixed density model.}
\revise{Another line performs \textit{auxiliary--main task alignment} so that self-supervised gradients are more likely to benefit the primary task. SR-TTT~\cite{lyu2022learning} feeds segmentation outputs into the auxiliary reconstruction branch; CTA~\cite{barbeau2025cta} aligns self-supervised and supervised encoders to encourage compatible gradients; and S4T~\cite{jeong2025synchronizing} predicts inter-task relations across multiple downstream tasks. Beyond vision, this paradigm extends to graph~\cite{zhang2024fully,pirhayatifard2025cross}, speech~\cite{behera2025test,dumpala2025test}, and time-series~\cite{christou2024test} settings. This strategy directly addresses pretext--task mismatch, but requires additional source-time design or known relations between tasks.}

\subsubsection{Pseudo-Label and Self-Training Signals}
\label{sec:pseudolabel}

\revise{Pseudo-label and self-training methods construct supervision from the model's own predictions or related test-time signals. This supervision may take the form of discrete or soft targets for supervised updates, or pseudo-rewards that identify preferable outputs within a reinforcement-learning objective. We organize these methods according to how such supervision is generated: from the adapted model itself, prototypes or neighboring samples, consensus across multiple predictions, or external models.}

\textbf{\textit{Predictions from the Adapted Model.}}
\revise{The most direct strategy uses the model's own prediction as a \textit{self-generated target}.} Goyal \etal~\cite{goyal2022test} show theoretically that this recovers the optimal TTA loss for cross-entropy-trained classifiers under a conjugate framework, unifying pseudo-labeling with entropy minimization. \revise{This strategy requires no external supervision, but can reinforce incorrect predictions through confirmation bias. Because self-generated targets can be noisy, another line develops \textit{pseudo-label correction}.} ECL~\cite{zeng2024rethinking} targets less probable categories via complementary labels to reduce incorrect pseudo-labeling, while IST~\cite{ma2024improved} corrects pseudo-labels via feature-similarity graph construction and stabilizes updates with parameter moving average. \revise{Correction improves robustness to noisy targets, although its effectiveness depends on reliable complementary or neighborhood structure.}

\textbf{\textit{Prototypes and Nearest Neighbors.}}
\revise{One strategy derives pseudo-labels from \textit{class prototypes} constructed online from high-confidence test features. Representative methods include} SHOT~\cite{liang2020we}, which freezes the source classifier for encoder alignment; T3A~\cite{iwasawa2021test}, which replaces the linear classifier with backpropagation-free pseudo-prototypes; and more recent works via decoupled learning~\cite{wang2025decoupled} and prototypical contrastive anchoring~\cite{karimi2025adapac}. \revise{Prototype-based targets are effective when target features form stable class clusters, but are sensitive to imbalance and prototype contamination. When a single class center cannot capture local target structure, \textit{nearest-neighbor targets} instead draw supervision from a memory bank:} Yang \etal~\cite{yang2021exploiting} exploit neighborhood structure for source-free adaptation; AdaContrast~\cite{chen2022contrastive} refines pseudo-labels by soft voting among neighbors; DLTTA~\cite{yang2022dltta} extends this to medical imaging. \revise{Neighbor-based targets preserve local structure, but require additional memory and can propagate errors through unreliable neighbors.}

\textbf{\textit{Consensus and Majority Voting.}}
\rthree{A third line constructs self-generated supervision by aggregating multiple predictions. At the view level, MEMO~\cite{zhang2022memo} minimizes the entropy of predictions marginalized over augmentations, while CoTTA~\cite{wang2022continual} averages teacher predictions across augmented views to form pseudo-targets. DPLOT~\cite{yu2024dplot} further uses paired original and horizontally flipped views to avoid the distribution gap introduced by stronger transformations. Consensus can also combine complementary evidence: TeSLA~\cite{tomar2023tesla} ensembles weakly augmented predictions and refines them with nearest neighbors, whereas MM-TTA~\cite{shin2022mm} combines intra-modal pseudo-label generation with cross-modal refinement for 3D segmentation. View-based consensus is effective when transformations preserve semantics, while hybrid or cross-modal consensus benefits from complementary errors across information sources. Both remain vulnerable when their constituent predictions share correlated biases. Consensus can also be converted into rollout-derived rewards in joint learning-and-scaling systems, whose update mechanisms are discussed in Section~\ref{sec:test-time-rl}.}

\textbf{\textit{External Model for Pseudo-labels.}}
\revise{External models can provide pseudo-supervision beyond the adapted classifier. Through \textit{foundation-model supervision}, DINOv2~\cite{oquab2023dinov2} and GPT-4o serve as external teachers for test-time knowledge distillation into the adapted model~\cite{weijler2025ttt,gong2025cross}. External teachers provide richer supervision, but require additional model access and may transfer teacher bias or domain mismatch. Yang \etal~\cite{yang2026diffusion} instead use \textit{diffusion-generated references} to support pseudo-label construction under distribution shift. Such references can be useful under input corruption, but introduce additional inference cost and depend on the quality of the generative prior.}

\subsubsection{External Feedback from Tools, Environments, or Users}
\label{sec:external-feedback}

The signals surveyed so far are self-contained within the model's computation. This section covers exogenous feedback that arises from the model's deployment context: rewards from an environment, outputs from invoked tools, and corrections from human users. These closed-loop interactions introduce a qualitatively different supervision structure unavailable to purely self-supervised approaches.

\textbf{\textit{Environment Interaction as a Signal.}}
\revise{Environment interaction can provide either self-supervised signals from observed transitions or explicit task outcomes. When explicit rewards are unavailable, \textit{interaction-derived self-supervision} constructs adaptation objectives from observed transitions or actions:} PAD~\cite{hansen2021self}, AugWM~\cite{ball2021augmented}, and MoVie~\cite{yang2023movie} adapt encoders or policies via inverse-dynamics prediction, dynamics-augmentation context, and latent dynamics modeling, respectively. \rthree{AdaJEPA~\cite{wang2026adajepa} instead executes an action chunk and uses the observed next-state transition to adapt its latent world model.} \revise{These signals are frequent and label-free, but their auxiliary objectives may not directly reflect task success. When task outcomes are observable, \textit{outcome-based feedback} provides more task-aligned supervision.} \rthree{FeedTTA~\cite{kim2025test} adapts a navigation agent from episode success or failure, whereas TT-VLA~\cite{liu2026fly} uses step-wise task-progress rewards to adapt a VLA policy. However, episode-level outcomes can be sparse or delayed, while denser progress rewards require a reliable measure of task advancement.}

\textbf{\textit{Tool-augmented Feedback as a Signal.}}
\rthree{External evaluators and executable tools can assess the quality or validity of test-time outputs and turn the resulting evidence into structured feedback. Through \textit{executable feedback for active updates}, T$^3$RL~\cite{liao2026tool} uses code execution to verify sampled rollouts and correct consensus-derived rewards. TTT-Discover~\cite{yuksekgonul2026learning} updates a policy from executable scientific objectives, while Alpha-RTL~\cite{zhou2026alphartl} uses EDA feedback for per-design optimization. Together, these methods show that executable feedback can support test-time learning by validating self-generated supervision or directly evaluating progress on the deployment task.. For tasks without executable correctness criteria, \textit{model-based evaluation} provides softer feedback: RLCF~\cite{zhaotest} uses CLIP image--text similarity as a test-time reward, while Reward-Adaptation~\cite{song2025reward} uses a reward model to assess pseudo-label reliability before updating. Learned evaluators apply more broadly, but inherit evaluator bias, calibration errors, and domain mismatch.}

\textbf{\textit{Human-in-the-loop as a Signal.}}
\rthree{Human feedback varies in bandwidth and specificity, ranging from binary evaluation to selectively queried labels and pairwise preferences. \textit{Binary evaluative feedback} indicates whether a prediction is correct without providing a replacement label. BiTTA~\cite{lee2025test} uses such judgments to correct uncertain predictions while retaining self-adaptation on confident samples. Binary judgments reduce feedback bandwidth, but do not specify the correct target when a prediction is wrong. More informative supervision can instead be acquired selectively.} \revise{When a limited annotation budget is available, \textit{active annotation} queries labels only for samples or regions expected to provide the greatest adaptation benefit:} SimATTA~\cite{gui2024active} uses entropy-diversity clustering, Li \etal~\cite{liexploring} apply uncertainty-diversity margin, EATTA~\cite{wang2025effortless} detects source-target border samples, ATASeg~\cite{yuan2023few} selects uncertain pixels within a click budget, and CPATTA~\cite{shi2025annotation} applies conformal coverage scores. \rthree{Queried labels provide explicit correction, but incur human cost and depend on a reliable acquisition criterion. For subjective or open-ended tasks where exact labels are difficult to provide, \textit{preference-based feedback} expresses which output or behavior is preferred:} Li \etal~\cite{liprovably}, COPR~\cite{zhang2025copr}, and Dueling RL~\cite{saha2023dueling} update policies online from pairwise human preferences. \rthree{Thus, binary judgments minimize feedback bandwidth, queried labels provide more explicit correction, and preferences support objectives without unique targets, but all three depend on reliable users and careful feedback acquisition.}

\subsection{What Is Updated at Test Time?}
\label{sec:what-updated}

The central question of this section is: during \revise{test-time learning} (TTL), which \revise{component or deployment state} should be updated in order to improve adaptability under distribution shift, task variation, or environmental perturbation, while controlling computational cost, avoiding catastrophic forgetting, and maintaining stable inference? Although existing methods take diverse forms, they largely follow the same high-level principle: they treat the modifiable component at test time as the core design choice, and balance adaptation capacity, stability, and efficiency by restricting or expanding the update scope. From this perspective, current methods can be broadly divided into five categories: input adaptation, partial parameter updates, auxiliary parameter updates, \revise{backbone or full-parameter updates}, and external memory or cache updates. \revise{These categories represent different update targets rather than a strict progression in adaptation strength.}

\revise{Across these categories, the update target determines both adaptation capacity and deployment burden, but a larger writable state does not imply a universal empirical advantage. \textbf{Input adaptation} preserves a frozen task model and is suitable when parameter access is unavailable, although iterative restoration may add latency and can remove task-relevant content. \textbf{Partial-parameter updates} usually offer a favorable stability--efficiency compromise when gradients are available, but may underfit large shifts or depend on fragile batch statistics. \textbf{Prompts and adapters} keep the backbone frozen and reduce optimizer state, making them attractive for large foundation models; nevertheless, they still require suitable insertion points and often backpropagation, and their restricted capacity can be sensitive to initialization or a single sample. \textbf{Backbone and full-parameter updates} provide the greatest flexibility for severe or task-level change, at the cost of peak memory, latency, catastrophic forgetting, and collapse, and therefore require reliable feedback and rollback controls. \textbf{Memory and cache updates} can be forward-only and reuse recurring experience, but assume local feature or temporal relevance and introduce contamination, staleness, privacy, and storage-growth risks. Thus, deployments should prefer the smallest update scope that captures the expected shift and avoid persistent, high-capacity updates when feedback quality or recovery mechanisms are weak.}

\begin{table*}[!t]
\centering
\caption{Taxonomy of methods by what is updated at test time.}
\label{tab:what-updated}
\renewcommand{\arraystretch}{1.2}
\footnotesize
\setlength{\emergencystretch}{3em}
\setlength{\tabcolsep}{2pt}
\resizebox{\textwidth}{!}{%
\begin{tabularx}{1.18\textwidth}{@{}
>{\raggedright\arraybackslash}p{0.11\textwidth}
>{\raggedright\arraybackslash}p{0.17\textwidth}
>{\raggedright\arraybackslash}p{0.20\textwidth}
>{\raggedright\arraybackslash}p{0.23\textwidth}
>{\raggedright\arraybackslash}p{0.20\textwidth}
>{\raggedright\arraybackslash}X@{}}
\toprule
\textbf{Update Target} & \textbf{Strategy} & \textbf{Representative Works} & \revise{\textbf{Required Access/}}\newline\revise{\textbf{Resources}} & \revise{\textbf{Applicable}}\newline\revise{\textbf{Scenarios}} & \revise{\textbf{Failure Modes}} \\
\midrule
\multirow{2}{*}{\parbox{0.11\textwidth}{Input}}
  & diffusion restoration   & DDA~\cite{gao2023back}; GDA~\cite{tsai2024gda} & \multirow{2}{*}{\parbox{0.23\textwidth}{\revise{Writable input/latent state; current input; optionally a generative restoration prior}}} & \multirow{2}{*}{\parbox{0.20\textwidth}{\revise{Inaccessible task models under recoverable input corruption}}} & \multirow{2}{*}{\parbox{0.21\textwidth}{\revise{Semantic distortion, projection to an incorrect source mode}}} \\ 
  & frequency calibration   & TF-Cal~\cite{zhao2022test} & & & \\[8pt]
\midrule
\multirow{3}{*}{\parbox{0.11\textwidth}{Partial\\Parameters}}
  & normalization statistics      & AdaBN~\cite{li2016revisiting}; Tent~\cite{wang2021tent} & \multirow{3}{*}{\parbox{0.23\textwidth}{\revise{Internal-activation access plus writable statistics or gradient access to selected modules}}} & \multirow{3}{*}{\parbox{0.20\textwidth}{\revise{Online shifts with limited memory and latency budgets and a backbone that should be preserved}}} & \multirow{3}{*}{\parbox{0.21\textwidth}{\revise{Insufficient capacity, batch-statistic noise, or selection of the wrong layers or channels}}} \\
  & channel \& layer selection    & GALA~\cite{sahoo2025layer}; AdaShadow~\cite{fang2024adashadow} & & & \\
  & module-level split            & \rtwo{In-Place TTT~\cite{feng2026place}} & & & \\[4pt]
\midrule
\multirow{2}{*}{\parbox{0.11\textwidth}{Auxiliary\\Parameters}}
  & prompt tuning                 & TPT~\cite{shu2022test}; FOA~\cite{niu2024foa} & \multirow{2}{*}{\parbox{0.23\textwidth}{\revise{Writable prompt/adapter, compatible insertion points, backbone-activation access, optimizer state}}} & \multirow{2}{*}{\parbox{0.20\textwidth}{\revise{Large pretrained backbones whose original parameters must remain frozen}}} & \multirow{2}{*}{\parbox{0.21\textwidth}{\revise{Prompt overfitting, initialization sensitivity, limited capacity, or interface mismatch}}} \\
  & adapters \& low-rank          & TTL~\cite{imam2025test}; BECoTTA~\cite{lee2024becotta} & & & \\[16pt]
\midrule
\multirow{2}{*}{\parbox{0.11\textwidth}{\revise{Backbone/}\\Full\\Parameters}}
  & \rthree{backbone-only}          & TTT~\cite{sun2020test};  \rthree{SHOT~\cite{liang2020we}} & \multirow{2}{*}{\parbox{0.23\textwidth}{\revise{Full weight and gradient access, reliable feedback, optimizer memory, and preferably reset or rollback state}}} & \multirow{2}{*}{\parbox{0.20\textwidth}{\revise{Severe or task-level changes with sufficient compute and operational controls}}} & \multirow{2}{*}{\parbox{0.21\textwidth}{\revise{Catastrophic forgetting, collapse, source-performance loss, and update instability}}} \\
  & \rthree{full-model}             & MEMO~\cite{zhang2022memo}; \rthree{AdaContrast~\cite{chen2022contrastive};} CoTTA~\cite{wang2022continual} & & & \\[22pt]
\midrule
\multirow{3}{*}{\parbox{0.11\textwidth}{External\\Memory}}
  & class prototypes       & T3A~\cite{iwasawa2021test}; \revise{DPE~\cite{zhang2024dpe}} & \multirow{2}{*}{\parbox{0.23\textwidth}{\revise{Feature and prediction access and writable persistent storage for prototypes, key--value entries, or interaction feedback}}} & \multirow{2}{*}{\parbox{0.20\textwidth}{\revise{Repeated related samples or tasks, forward-only adaptation, or reusable personalized experience}}} & \multirow{2}{*}{\parbox{0.21\textwidth}{\revise{Stale or contaminated entries, unbounded growth, privacy leakage, and retrieval mismatch}}} \\
  & \rthree{instance retrieval}   & AdaNPC~\cite{zhang2023adanpc}; TDA~\cite{karmanov2024efficient} & & & \\[8pt]
  & \rthree{semantic experience} & MemoryBank~\cite{zhong2024memorybank}; PhysMem~\cite{li2026learning} & & & \\[8pt]
\bottomrule
\end{tabularx}%
}
\end{table*}

\subsubsection{Input Adaptation}
\label{sec:input-latent}

Input adaptation methods address distribution shift by modifying the test input itself rather than any model parameter, so that the input is projected closer to the source domain before prediction. \rthree{Lens~\cite{baek2025lens} moves this intervention before image capture by adapting camera-sensor parameters to the current model and scene.} \revise{Because input-level adaptation can operate independently per sample, it avoids some sensitivity to batch size and data order, although its stability still depends on the restoration objective and optimization process.} \revise{In \textit{diffusion-based restoration},} DDA~\cite{gao2023back} projects each target input toward the source domain via a source-trained unconditional diffusion model with image guidance and classifier self-ensembling. GDA~\cite{tsai2024gda} extends this to broader OOD types by combining marginal entropy guidance with style and content preservation during reverse sampling. FDD~\cite{tjiobeautifying} conditions diffusion on input-predicted high-pass and low-pass frequency filters to preserve shape and color cues for dense prediction tasks. \rthree{FOCUS~\cite{tjio2026focus} instead guides reverse diffusion with learned spatially adaptive frequency priors to better preserve task-relevant semantics.} \revise{Rather than modifying the restoration process itself, SDA~\cite{guo2025everything} finds that diffusion-restored inputs remain misaligned with the source model and uses source-time fine-tuning on mixed-diffusion synthetic data to prepare the model for this synthetic test domain. This strategy is suitable when corruption can be projected toward a useful source-like domain, but iterative sampling adds latency and may alter task-relevant content.} \revise{As a lighter \textit{frequency-domain calibration} strategy,} TF-Cal~\cite{zhao2022test} calibrates target-domain amplitude features using a source-domain prototype at test time to reduce the style gap while preserving semantic information. \revise{This avoids generative restoration, but relies on suitable frequency-domain priors.}

\subsubsection{Partial Parameter Updates}
\label{sec:partial-params}

Partial parameter update methods address distribution shift by adapting only a small, designated subset of model parameters at test time rather than the full network. \revise{Their shared premise is that useful corrections can often be achieved through localized updates, thereby preserving the remaining weights, mitigating forgetting, and keeping adaptation cost tractable.} Existing work can be organized by how the update subset is identified.

\textbf{\textit{Normalization-based adaptation.}} \revise{Early methods such as AdaBN~\cite{li2016revisiting} replace} source BN statistics with target statistics, while Tent~\cite{wang2021tent} further updates affine parameters in BN layers through entropy minimization. However, directly relying on target batch statistics can be brittle in practice. To address this, $\alpha$-BN~\cite{you2021test} mixes source and target BN statistics with a fixed ratio to preserve discriminative structure, while TTN~\cite{lim2023ttn} learns channel-wise interpolating weights based on domain-shift sensitivity to balance source and target statistics. Meanwhile, MixNorm~\cite{hu2021mixnorm}, DCN~\cite{jiang2022domain}, and TEMA~\cite{su2024unraveling} are designed for arbitrary-batch-size, single-sample, and realistic mini-batch BN adaptation settings, respectively. \revise{UnMix-TNS}~\cite{tomar2024mixing}, RoTTA~\cite{yuan2023robust}, and CycleTTA~\cite{jiang2024advancing} further improve BN adaptation under temporally correlated, dynamic, and continuous cyclic test-time shifts, respectively. \revise{Overall, normalization-based adaptation is inexpensive, but its reliability depends on representative target statistics and sufficiently stable test streams.}

\textbf{\textit{Selective updates on designated subsets.}} A channel selection method~\cite{vianna2024channel} selectively adapts channels to improve robustness under label distribution shift, while GALA~\cite{sahoo2025layer} and AdaShadow~\cite{fang2024adashadow} identify beneficial or adaptation-critical layers for more reliable or efficient test-time updates. More explicit parameter-level selection is explored in FIESTA~\cite{honarmand2025fiesta} and PSMT~\cite{tian2024parameter}, where Fisher information is used to identify adaptation-critical parameters in FIESTA and to preserve crucial parameters during continual adaptation in PSMT. \rtwo{A module-specific design, In-Place TTT~\cite{feng2026place}, repurposes the final projection matrix of MLP blocks as fast weights and updates it in place during inference.} \revise{Such localized updates provide a flexible capacity--stability trade-off, but depend on reliably identifying which parameters should adapt.}

\subsubsection{Auxiliary Parameter Updates}
\label{sec:prompts-adapters}

Auxiliary parameter methods leave the pretrained backbone entirely frozen and instead introduce a small set of newly added, trainable parameters to absorb distribution shift at test time. \revise{This design separates pretrained knowledge from task- or domain-specific adjustments, making adaptation easier to reverse and store than full-model updates.} Existing methods can be grouped by the form of the auxiliary parameters they introduce, namely prompts, adapters, or \revise{low-rank} modules, each offering a different \revise{trade-off} between expressivity, placement in the network, and adaptation cost.

\textbf{\textit{Prompt tuning.}} TPT~\cite{shu2022test} introduces test-time prompt tuning for vision-language models by optimizing prompts on a single test sample via marginal entropy minimization over augmented views. Subsequent works extend TPT to broader test-time settings: B2TPT~\cite{meng2025black} addresses the black-box MaaS scenario by optimizing low-dimensional intrinsic prompts with a derivative-free evolution algorithm, while Active TPT~\cite{sarkar2024active} incorporates active learning by querying true labels for uncertain test samples. Prompt tuning has also been extended beyond image classification: DTS-TPT~\cite{yan2024dts} performs temporal-synchronized prompt tuning for zero-shot video activity recognition, while Shao et al.~\cite{shao2024adaptive} use CLIP-guided prompt representations as degradation fuzzy sets and introduce test-time self-supervised prompt fine-tuning for image restoration. \rthree{Beyond extending prompt tuning to new access conditions and application settings, recent work has also improved the reliability of the prompt update itself. For each test instance, DiffTPT~\cite{feng2023diverse} enriches the available evidence with filtered diffusion-generated views, while PromptAlign~\cite{samadh2023align} constrains adaptation using source feature statistics. When samples arrive as a stream, HisTPT~\cite{zhang2024histpt} retrieves historical knowledge and DynaPrompt~\cite{xiaodynaprompt} dynamically maintains a prompt buffer to reuse information across related inputs. Other methods address calibration and optimization bias: O-TPT~\cite{sharifdeen2025otpt} orthogonalizes textual features, FPP~\cite{jang2026fpp} provides a data-free flatness-aware initialization, and D$^2$TPT~\cite{song2026d2tpt} combines retrieved knowledge with reliability-aware regularization. Together, these methods strengthen the evidence used for each update, reuse information across test samples, or constrain unreliable optimization while retaining the frozen backbone.} \revise{However, its limited capacity can constrain adaptation.}

\textbf{\textit{Adapters and low-rank modules.}} 
\rthree{The central design choice is how auxiliary adaptation capacity is parameterized and managed. \textit{Low-rank parameterization} constrains updates to a compact parameter subspace. TTL~\cite{imam2025test} adapts low-rank parameters in transformer attention through confidence maximization, LoTT-PC~\cite{ye2026lott} updates low-rank modulation parameters through decoder-free masked-feature alignment, and VDS-TTT~\cite{moradi2025continuous} uses verifier-selected pseudo-labels to update LoRA parameters. This parameterization reduces optimization and storage costs, but limits adaptation to a predefined update space. \textit{Task-structured module design} instead places writable state at a functionally relevant point in the prediction pipeline. GOAT~\cite{zhangavoiding} introduces low-rank node feature transformations for graph adaptation, while PETSA~\cite{medeiros2025accurate} combines lightweight input and output calibration modules with dynamic gating. These modules can target the part of the prediction process affected by the shift, but their design is more architecture- and task-dependent. \textit{Modular composition and routing} distributes adaptation capacity across multiple components when a single auxiliary module cannot represent heterogeneous or recurring shifts. ViDA~\cite{liu2023vida} dynamically combines high- and low-rank adapters, whereas BECoTTA~\cite{lee2024becotta} and MoE-TTA~\cite{iftee2024moe} route inputs across domain-specific adapter experts. Composition increases adaptation capacity and routing helps separate knowledge associated with different test conditions, but both increase module-storage and selection requirements. Overall, low-rank parameterization constrains the update space, task-structured module design determines where adaptation acts, and modular composition and routing determine how adaptation capacity is distributed across reusable auxiliary states.}

\subsubsection{\revise{Backbone and Full-Parameter Updates}}
\label{sec:full-params}

\rthree{These methods update much more of the model than the preceding approaches. Backbone-only methods change the learned representation while keeping the task head fixed, whereas full-model methods change both. This difference determines what can be corrected and how widely an erroneous update can affect later predictions.}

\textbf{\textit{\rthree{Backbone-only updates.}}} \rthree{TTT~\cite{sun2020test} establishes the basic route by preparing an auxiliary objective during source training and using it to update the shared backbone at deployment. TTT++~\cite{liu2021ttt++} and TTT-MAE~\cite{gandelsman2022tmae} retain this update target but use contrastive learning with feature alignment and masked reconstruction, respectively. Source-time preparation is not always available. SHOT~\cite{liang2020we} instead freezes the source classifier and updates the target feature extractor through information maximization and pseudo-labeling. In both settings, the adapted features must remain meaningful to the fixed task head or improving the auxiliary objective can otherwise degrade task predictions. To address this, SR-TTT~\cite{lyu2022learning} feeds segmentation outputs into its auxiliary reconstruction branch, while CTA~\cite{barbeau2025cta} aligns self-supervised and supervised encoders to preserve this compatibility. Continual-MAE~\cite{liu2024continual} further extends encoder adaptation to continual updates, illustrating that a fixed task head alone does not prevent representation drift. Backbone-only adaptation is therefore more expressive, but also more expensive and disruptive, than updating selected layers, prompts, or adapters.}

\textbf{\textit{\rthree{Full-model updates.}}} \rthree{Full-model adaptation also allows the task head to change. MEMO~\cite{zhang2022memo} adapts all parameters by minimizing marginal entropy across augmented views of the current sample, while AdaContrast~\cite{chen2022contrastive} combines contrastive learning and pseudo-label learning on target data. Updating the entire model can correct a broader range of errors, but unreliable feedback can now degrade both the representation and the final decision boundary. The risk becomes more consequential when these changes are retained across later inputs and are accumulated. CoTTA~\cite{wang2022continual} combines prediction and weight averaging with stochastic restoration to preserve a stable reference. ABR~\cite{wang2026adaptive} and PLATO-TTA~\cite{xie2025plato} use adaptive re-initialization and consistency-based backtracking to undo harmful updates, while AWMC~\cite{lee2023awmc} distributes adaptation across parameter-shared models to reduce mode collapse. Thus, full-model adaptation removes the fixed-head constraint, but requires feedback that can guide both representations and decision boundaries. Its broader capacity also makes feedback errors more destructive, so persistent updates should incorporate restoration or rollback.}

\subsubsection{External Memory and Cache Updates}
\label{sec:memory-cache}
\rthree{External-memory methods update a model-external state from deployment-time observations or feedback and directly reuse that state for subsequent prediction or decision making. The memory itself therefore carries acquired test-time knowledge and remains part of the inference mechanism. Many such methods are gradient-free, although compound systems may update memory together with other writable states.}

\textbf{\textit{\rthree{Aggregated class-prototype memory.}}} \rthree{These methods compress incoming evidence into compact class-level summaries. T3A~\cite{iwasawa2021test} constructs evolving class prototypes from test features and predicts by their similarity to the current input. DPE~\cite{zhang2024dpe} extends this idea by jointly evolving visual and textual prototypes, while PTA~\cite{huang2026pta} integrates historical features directly into weighted class prototypes to avoid retrieval over a growing exemplar cache. Prototype memory is compact and efficient when class structure remains coherent, but imbalance, multimodal classes, or incorrect assignments can distort a summary that affects many later predictions.}

\textbf{\textit{\rthree{Instance-retrieval memory.}}} \rthree{Rather than compressing each class into one representative, these methods preserve individual entries and retrieve locally relevant evidence. AdaNPC~\cite{zhang2023adanpc} stores feature--label pairs, predicts by nearest-neighbor voting, and writes each test feature and predicted label back to memory. TDA~\cite{karmanov2024efficient} instead maintains positive and negative key--value caches whose retrieved scores refine vision--language predictions, while BoostAdapter~\cite{zhang2024boostadapter} combines historical entries with samples generated around the current instance. Recent designs make the stored evidence more structured: MCP~\cite{chen2025mcp} coordinates entropy, alignment, and negative caches to construct more reliable prototypes, whereas Point-Cache~\cite{sun2025point} preserves both global and local-part information for point-cloud recognition. Instance memories retain richer target structure than prototypes, but require more storage and retrieval computation and remain sensitive to irrelevant or noisy neighbors.}

\textbf{\textit{\rthree{Semantic and structured memory.}}} \rthree{External memory can also retain deployment experience in a form directly reusable for later reasoning or action. MemoryBank~\cite{zhong2024memorybank} updates a retrievable interaction history across sessions, Dynamic Cheatsheet~\cite{suzgun2026dynamic} and ReasoningBank~\cite{ouyang2026reasoningbank} distill experience into reusable strategies, and PhysMem~\cite{li2026learning} stores verified physical principles for later robot decisions. Their objectives and long-term evolution are discussed in Section~\ref{sec:memory_augmented}. These methods illustrate that writable memory can improve capabilities beyond distribution-shift adaptation.}

\subsection{Temporal Horizons of Test-Time Learning}
\label{sec:temporal-horizons}

\rtwo{Temporal design has two independent axes. The \textbf{supervision horizon} specifies the range and structure of evidence used for one update, from a single input, token, or interaction to a batch, history, or trajectory. The \textbf{persistence horizon} specifies how long the resulting writable state---parameters, statistics, prompts, fast weights, modules, or memory---is retained. The former shapes signal reliability, latency, and data requirements. The latter determines whether useful experience and harmful errors remain local or accumulate. These axes should be chosen separately: a method may update from one observation but retain the result across a stream, or aggregate a trajectory within an episode and then reset. The same temporal design therefore applies whether TTL adapts to distribution shift, memorizes context, personalizes behavior, or learns through interaction.}

\begin{figure*}[!t]
\centering
\includegraphics[width=0.75\linewidth]{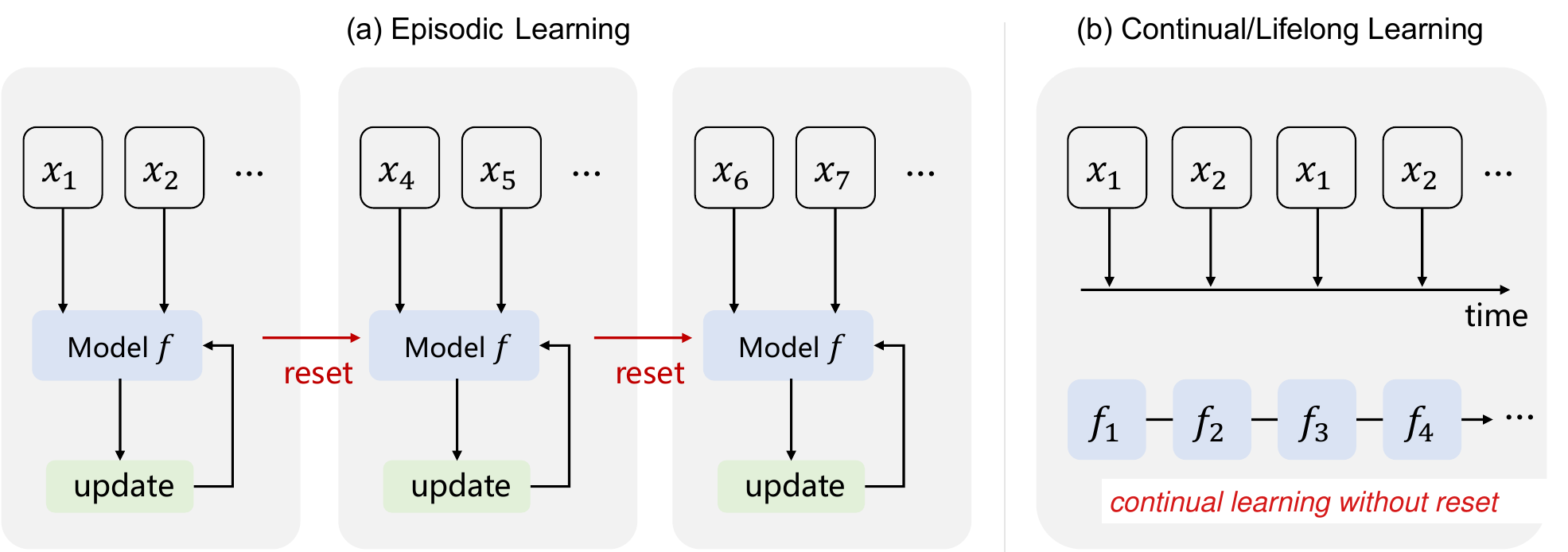}
\caption{\rtwo{Comparison of episodic and continual persistence in TTL. Episodic methods reset the updated state at a chosen boundary, whereas continual methods retain it across the deployment stream. The illustrated model state can also represent prompts, fast weights, modules, or external memory.}}
\label{fig:episode-vs-continue}
\end{figure*}

\subsubsection{\rtwo{Supervision Horizon: Single- vs. Multi-Sample Learning}}
\label{sec:supervision-horizon}

\rtwo{The supervision horizon concerns the evidence used for each update. \textbf{\textit{Single-sample supervision}} uses one local evidence unit and its derived views or structure. MEMO~\cite{zhang2022memo} and TPT~\cite{shu2022test} seek confident or augmentation-consistent predictions from one input. TTT~\cite{sun2020test} and MAE-TTT~\cite{gandelsman2022tmae} use prepared pretext tasks or reconstruction, whereas TTT-Layer~\cite{sun2024learning} uses each token's self-supervision to update fast-weight context memory. Thus, a local evidence unit may be an input or a token, and may update parameters, prompts, or memory. Prediction-derived signals require less task-specific preparation but inherit model errors, whereas structure-derived signals require a suitable auxiliary objective. A narrow horizon supports immediate, low-buffer updates, but the signal can be noisy and repeated optimization costly.}

\rtwo{\textbf{\textit{Multi-sample supervision}} uses relationships within an unordered batch, rolling history, or structured sequence. Tent~\cite{wang2021tent}, TTT++~\cite{liu2021ttt++}, and ActMAD~\cite{mirza2023actmad} aggregate batch-level predictions or feature statistics. RoTTA~\cite{yuan2023robust} and SAR$^2$~\cite{niu2025adapt} retain selected historical features or predictions, whereas ViTTA~\cite{lin2023video} and BayesTTA~\cite{cui2025bayestta} model temporal evolution. In interactive TTL, the sequence instead comprises actions, feedback, and outcomes: TT-VLA~\cite{liu2026fly} learns from step-wise environmental feedback, while ReflectivePlanning~\cite{hong2026learning} uses execution failures. Within these methods, batch aggregation requires a sufficiently coherent group, rolling history improves coverage but introduces buffering and staleness, and sequence modeling assumes that order conveys useful dynamics. Multi-sample supervision is therefore most useful when related observations jointly provide reliable evidence, whether they are target samples or interaction steps.}

\subsubsection{\rtwo{Persistence Horizon: Episodic and Continual Learning}}
\label{sec:persistence-horizon}

\rtwo{The persistence horizon concerns the lifetime of an update, independently of how its evidence is constructed. \textbf{\textit{Episodic persistence}} retains writable state only within a bounded deployment unit and resets it afterward. TTT~\cite{sun2020test}, MEMO~\cite{zhang2022memo}, TPT~\cite{shu2022test}, and MAE-TTT~\cite{gandelsman2022tmae} commonly restore the source state after each input. EFSA~\cite{huzaifa2024efsa} resets after a query-centered episode, while FCL~\cite{yun2026fair} and FairTPT~\cite{launayfairness} keep context or prompt updates task-local. For context memorization, TTT-Layer~\cite{sun2024learning} retains token-updated fast-weight memory throughout the current sequence. For interactive generation, SplatPainter~\cite{zheng2025splatpainter} trains a Gaussian representation within one asset-editing episode. Single-domain protocols instead reset before a new target domain~\cite{niu2023sar,lee2024deyo,hu2025beyond}. Although their objectives differ, these protocols prevent updated state from affecting unrelated deployment units. The reset interval is also distinct from the supervision horizon: Tent~\cite{wang2021tent} updates batch by batch, while its state may be reset or retained according to the deployment protocol. Parameter-based methods restore source parameters, whereas prompt-, buffer-, or memory-based methods can discard episode-local auxiliary state. Shorter intervals limit error propagation, whereas longer episodes reuse more context but rely more strongly on within-episode consistency~\cite{bennequin2021bridging}.}

\rtwo{\textbf{\textit{Continual persistence}} carries writable state across the deployment stream so later predictions or actions can reuse earlier experience. The retained state may be parameters or statistics, modules, or external memory. For model adaptation, CoTTA~\cite{wang2022continual} and EATA~\cite{niu2022eata} anchor updates to stable source or teacher knowledge, while SAR~\cite{niu2023sar} and RoTTA~\cite{yuan2023robust} control which samples and updates are retained. AETTA~\cite{lee2024aetta} instead estimates target accuracy without labels and can support recovery when adaptation degrades. CoLA~\cite{chen2024cross} and BECoTTA~\cite{lee2024becotta} isolate adaptation in modular state. For longer-term experience reuse, Cheatsheet~\cite{suzgun2026dynamic}, PhysMem~\cite{li2026learning}, and ReflectivePlanning~\cite{hong2026learning} respectively retain strategies, physical knowledge, and trial-and-error feedback. These mechanisms regulate different aspects of persistence: anchoring limits change, reliability control governs admission, and modular or memory-based designs determine where experience is stored. Persistent parameters can drift or collapse, whereas persistent memory can become stale, contaminated, or excessively large. User-specific state may also leak across contexts. Episodic persistence therefore isolates task-, user-, or asset-specific state, while continual persistence supports adaptation, personalization, and experience reuse. The appropriate horizon depends on how strongly evidence is related across deployment units and how costly an erroneous retained update would be. Figure~\ref{fig:episode-vs-continue} summarizes the distinction.}

\subsection{Test-Time Learning for Adaptation}
\label{sec:tta_adaptation}

This section discusses how test-time learning is instantiated to close the gap between the source and target distributions. We begin by formalizing domain shift and the goals of distribution alignment (Section~\ref{sec:domain_shift}), and review two foundational TTL paradigms: test-time training and fully test-time adaptation (Section~\ref{sec:ttt_ftta}). \revise{We then examine practical constraints along the data and temporal protocols (Section~\ref{sec:source_free_online}) and the optimization and access requirements (Section~\ref{sec:blackbox_adaptation}).} \revise{Finally, Section~\ref{sec:tta_theory} summarizes what current theory establishes about the reliability and limits of test-time adaptation.}

\subsubsection{Domain Shift and Distribution Alignment}
\label{sec:domain_shift}

Formally, domain shift indicates that the target distribution $p_{\mathcal{T}}(x, y)$ encountered at deployment departs from the source distribution $p_{\mathcal{S}}(x, y)$ on which the model was trained. Given that $p(x,y)=p(x)\,p(y\mid x)=p(y)\,p(x\mid y)$, domain shift can be taxonomized by which factor changes.

\textbf{Definition 2 (Covariate Shift).}
The marginal input distribution changes while the labeling function is preserved: $p_{\mathcal{S}}(x)\neq p_{\mathcal{T}}(x)$ and $p_{\mathcal{S}}(y\mid x)=p_{\mathcal{T}}(y\mid x)$. This is the most widely studied setting in test-time adaptation \cite{wang2021tent, liang2025comprehensive} and covers corruptions, sensor changes, and style variation.

\textbf{Definition 3 (Label Shift / Prior Shift).}
The class prior changes while the class-conditional input distribution is preserved: $p_{\mathcal{S}}(y)\neq p_{\mathcal{T}}(y)$ and $p_{\mathcal{S}}(x\mid y)=p_{\mathcal{T}}(x\mid y)$. Label shift arises naturally in medical screening, fraud detection, and any deployment where class prevalence varies over time \cite{niu2023sar,yuan2023robust}.

In practice, real-world distribution shifts are often compound, combining covariate and label shifts simultaneously \cite{niu2023sar,zhao2023ttapitfal}. The principle behind TTL-based adaptation methods is to minimize the divergence between the source and target representations. Let $g_\phi: \mathcal{X}\to\mathcal{Z}$ denote the feature extractor parameterized by the learnable subset $\phi$, and let $q_{\mathcal{S}}(z)$, $q_{\mathcal{T}}(z)$ be the induced feature distributions under source and target inputs, respectively. Distribution alignment seeks:
\begin{equation}\label{eq:alignment}
    \phi^{*} = \arg\min_{\phi}\; \mathcal{D}\!\left(q_{\mathcal{T}}^{(\phi)}(z)\;\|\;q_{\mathcal{S}}(z)\right) + \lambda\,\mathcal{R}(\phi),
\end{equation}
where $\mathcal{D}$ is a divergence measure (\eg, MMD \cite{Gretton2012mmd}, KL divergence, or optimal-transport cost \cite{Courty2016ot}), and $\mathcal{R}(\phi)$ is a regularizer that ensures a small change from $\theta^{*}$, \revise{\ie, the frozen source model in the traditional deployment assumption}.

\revise{The following sections organize adaptation along three design choices. The \textbf{objective/preparation choice} (Section~\ref{sec:ttt_ftta}) contrasts \textit{training-prepared TTT}, which adds a source-stage objective or architecture and depends on pretext-task relevance, with \textit{fully test-time adaptation}, which uses test-accessible signals on off-the-shelf models but faces unreliable self-supervision. The \textbf{data/temporal protocol choice} (Section~\ref{sec:source_free_online}) spans source-retaining/source-free, offline/online, and reset/continual adaptation: source access and offline processing provide stronger anchors, whereas source-free, online, or continual protocols fit privacy-constrained or streaming deployment but raise drift risk. The \textbf{optimization/access choice} (Section~\ref{sec:blackbox_adaptation}) contrasts gradient updates with \textit{forward-only, gradient-free, or black-box methods}, which fit restricted-access or memory-constrained deployment but sacrifice update flexibility or require extra queries. These choices combine; a method may be source-free, online, and forward-only, \eg, EVA-0~\cite{chen2026eva0}.}

\subsubsection{Test-Time Training and Fully Test-Time Adaptation}
\label{sec:ttt_ftta}

\begin{figure*}[h]
\centering
\includegraphics[width=0.75\linewidth]{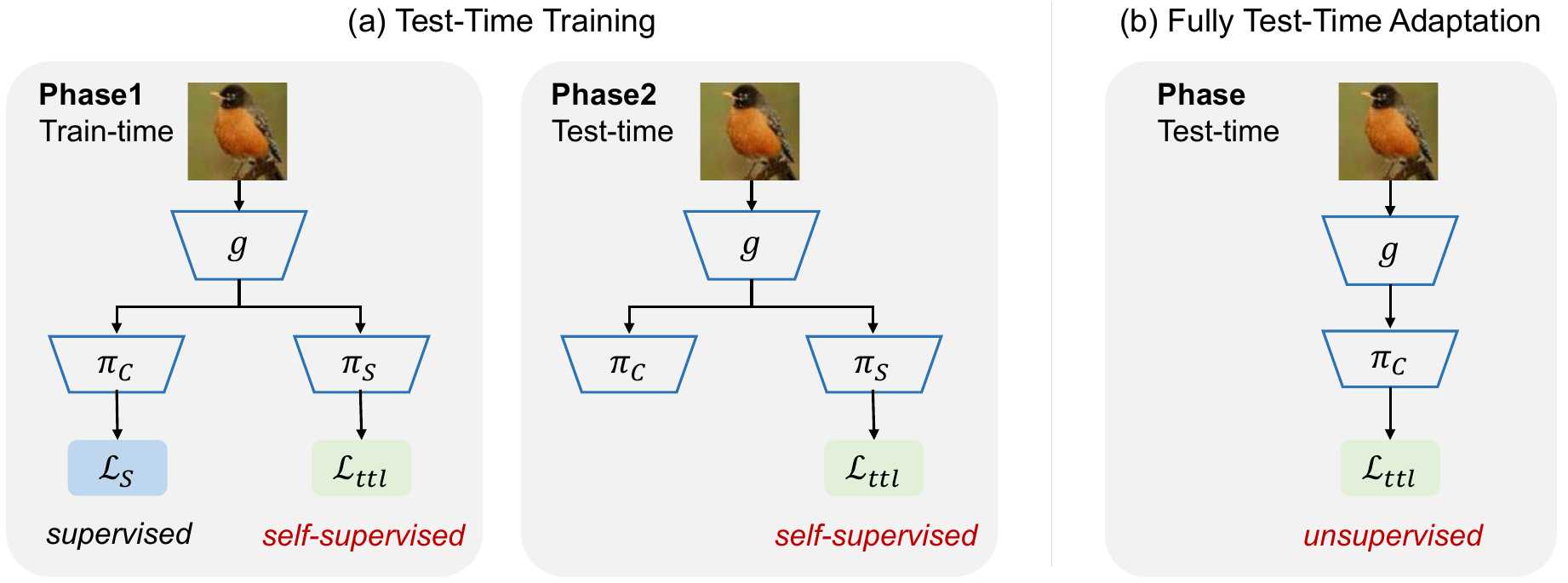}
\caption{Two representative types of test-time learning for model adaptation.}
\label{fig:ttt-ftta}
\end{figure*}

\rtwo{Test-time training (TTT)~\cite{sun2020test} and fully test-time adaptation (FTTA)~\cite{wang2021tent} are two foundational paradigms for adapting models under distribution shift. Both update a model using unlabeled test data, but differ in when the adaptation objective is prepared: TTT incorporates an auxiliary objective or architecture during source training, whereas FTTA constructs its objective directly from test-time signals available to an off-the-shelf pretrained model (see Figure~\ref{fig:ttt-ftta}).}

\textbf{\textit{Test-Time Training.}}
\rtwo{TTT uses a two-phase framework~\cite{sun2020test}. During source training, a shared encoder $g$ supports both a supervised task head $\pi_C$ and a self-supervised head $\pi_S$. At test time, the auxiliary objective $\mathcal{L}_{ttl}$ updates the shared encoder before the task prediction is produced. This preparation can provide task-relevant test-time gradients, but successful adaptation still depends on alignment between the auxiliary and primary tasks.}

\rtwo{Subsequent work mainly improves TTT through \textit{auxiliary-objective design} and \textit{auxiliary--main task alignment}. The former replaces the original rotation objective with signals such as masked reconstruction~\cite{gandelsman2022tmae} or next-token prediction~\cite{Hu2025TLM}, whereas the latter improves the relevance of test-time updates through feature-statistics alignment~\cite{liu2021ttt++} or meta-learning~\cite{bartler2022mt3,gu2025docttt}. These mechanisms are reviewed from the feedback-signal perspective in Section~\ref{sec:reconstruction}. Overall, TTT is most suitable when source-time preparation is possible and a task-relevant auxiliary objective can be designed.}


\textit{\textbf{Fully Test-Time Adaptation.}} \rtwo{Unlike TTT, FTTA adapts an off-the-shelf model without altering its source-training process. Tent~\cite{wang2021tent} provides a foundational example through \emph{test-time entropy minimization}:}
\begin{equation}
    \mathcal{H}(\hat{y}) = -\sum_c \hat{y}_c \log \hat{y}_c
\end{equation}
\rtwo{where $c$ is the number of classes. Minimizing prediction entropy encourages confident target predictions and pushes decision boundaries toward low-density regions~\cite{chapelle2005semi}, but assumes that confidence remains correlated with correctness. Subsequent work improves reliability through sample selection, representation regularization, and calibration or collapse-prevention mechanisms~\cite{niu2022eata,niu2023sar,lee2024deyo,zhang2025come,tan2025uncertainty}; these developments are detailed in Section~\ref{sec:entropy}.}

\rtwo{Beyond entropy minimization, FTTA can construct objectives from prediction consistency or pseudo-labels, whose mechanisms and representative methods are reviewed in Sections~\ref{sec:consistency} and~\ref{sec:pseudolabel}. These alternatives broaden the available test-time signals without requiring a redesigned source-training pipeline, but still depend on the accuracy and diversity of self-generated feedback. Overall, TTT is preferable when source training can be modified to prepare an aligned auxiliary objective, whereas FTTA is more practical when only an existing model is available. This lower preparation cost, however, brings greater exposure to confirmation bias, drift, and collapse under unreliable feedback.}

\subsubsection{Source-Free, Online, and Continual Adaptation}
\label{sec:source_free_online}

A TTL algorithm is defined by three practical design dimensions: how much source information it accesses, whether it adapts offline or online, and whether the test stream contains a single domain or a sequence of evolving domains.

\textbf{\textit{From Source-Dependent to Source-Free.}} The amount of source information used at test time ranges along a spectrum~\cite{su2024revisiting,liang2025comprehensive}. \revise{On one extreme, 1) \textit{Source-dependent} methods retain labeled source data during deployment to anchor adaptation, as in AR-TTA~\cite{sojka2023ar}. This deployment protocol differs from TTT methods that use source data before deployment to prepare an auxiliary objective or architecture but need not retain the dataset at test time~\cite{gu2025docttt,gandelsman2022tmae,liu2021ttt++}.} However, this incurs heavy storage, computation, and privacy costs.
Instead, 2) \textit{Source-light} methods leverage the unlabeled source data~\cite{mirza2023actmad,niu2022eata}, including to estimate the statistics of the source distribution for domain alignment~\cite{mirza2023actmad,su2024revisiting,niu2024foa}, or identify important parameters for anti-forget regularization~\cite{niu2022eata,deng2023efficient}. However, accessing source data still raises privacy concerns and is not available for third-party models. \revise{To adapt an arbitrary off-the-shelf model, 3) \textit{Source-free} methods use the pretrained model and current test data without retaining source samples; many fully TTA methods operate in this setting~\cite{wang2021tent,zhang2022memo,niu2023sar,wang2022continual}. This is the most privacy-preserving and deployment-friendly protocol, but the lack of source anchors makes stable adaptation more challenging~\cite{niu2023sar,gong2022note,chenzerosiam}.}

\textbf{\textit{Offline vs.\ Online adaptation.}} 1) \textit{Offline} TTA, sometimes called source-free domain adaptation, assumes that the entire target test set is collected in advance and the model is allowed multiple passes over this data before producing final predictions~\cite{liang2020we,liang2021source,li2021imbalanced}. This setting benefits from dataset-level statistics, stable gradient estimates, and the freedom to run longer optimization, but it fundamentally requires storing and re-accessing test samples, which conflicts with streaming real-world deployments~\cite{liang2025comprehensive}. 2) \textit{Online} TTA, by contrast, \revise{adapts on a per-batch or per-sample basis} as data arrives~\cite{wang2021tent,niu2022eata,niu2023sar} and emits predictions immediately after a single update step. This paradigm is necessary for latency-sensitive applications such as autonomous driving~\cite{wang2022continual} and embedded perception~\cite{deng2025zoa,niu2024foa}, and matches the data-access pattern of production systems. However, \revise{with small batches or highly correlated data streams}~\cite{niu2023sar,gong2022note,chenzerosiam}, the design scope of test-time objectives becomes more limited, and TTA suffers from a higher risk of collapse.

\textbf{\textit{Single-Domain vs.\ Continual Adaptation.}} 1) In the \textit{single-domain} setting, adaptation targets a single, fixed OOD distribution~\cite{wang2021tent}, and resets the model to its source weights before each new target domain~\cite{niu2023sar,lee2024deyo,hu2025beyond}. This isolates domain interfaces but is unrealistic for long-running deployments where the environment evolves continuously~\cite{wang2022continual}, \eg, seasonal shifts. 
2) \textit{Continual} TTA~\cite{wang2022continual,niu2022eata,marsden2024universal,zhang2025dpcore} addresses this by allowing a single model to adapt across a sequence of evolving domains without resets. While arguably a more realistic and high-potential protocol, continual TTA is also the most challenging due to \textit{catastrophic forgetting}~\cite{marsden2024universal,chen2024cross}, \revise{where adaptation to the current domain overwrites parameters needed for previously encountered domains and degrades generalization}, and \textit{error accumulation}~\cite{wang2022continual,dobler2023robust}, \revise{where erroneous pseudo-label signals accumulate and amplify during long-term adaptation}.
Existing works address these jointly via: 1) Reliable test-time objective design, including \textit{sample selection}~\cite{niu2022eata,niu2023sar,lee2024deyo} to remove noisy learning signals, \textit{uncertainty calibration}~\cite{tan2025uncertainty,zhang2025come} to mitigate overfitting, \textit{defining structures} on predictions~\cite{han2025rem}, features~\cite{niu2025adapt,ni2025maintaining}, and gradients~\cite{shin2024gradient} to enhance efficacy and robustness; 2) Stable update mechanisms, including \textit{Fisher-informed regularization}~\cite{niu2022eata,deng2023efficient} that prevents drastic changes in important parameters, \textit{teacher-anchoring restoration}~\cite{wang2022continual,marsden2024universal,song2023ecotta,chakrabarty2023santa} which aligns the predictions/parameters with the stable anchor model, \textit{gradient filterization}~\cite{lee2024continual,duan2025lifelong} which suppresses suspicious gradient updates within a window. 3) Multi-modeling~\cite{song2023test,chen2024cross,zhang2025dpcore}, which decouples and maintains a set of learned knowledge to address catastrophic forgetting and facilitates knowledge reuse, whereas knowledge reuse is guided by loss minimization in CoLA~\cite{chen2024cross}, and by domain similarity in~\cite{song2023test,zhang2025dpcore,vray2025reservoirtta}.
\rthree{DOCO~\cite{yang2026doco} further considers open-set continual streams by dynamically separating likely in- and out-of-distribution samples and learning a source-aligned compensation prompt.}

\subsubsection{Forward-only and Gradient-Free Adaptation}
\label{sec:blackbox_adaptation}

A practically important constraint in many deployment scenarios is that the model to be adapted is inaccessible for gradient computation, \eg, it is a quantized model on an edge device.
Even when the model is technically accessible, the memory overhead of storing intermediate activations for gradient computation can be several times larger than inference alone (\eg, 5,165\,MB for TENT \vs~832\,MB for inference on ViT-Base~\cite{niu2024foa}), limiting their practicality.
These constraints motivate a family of methods that achieve TTL/TTA using only forward passes, without relying on backpropagation.
We organize these methods into two categories: \emph{forward-optimization TTL} methods, which iteratively optimize learnable parameters using derivative-free optimizers, and \emph{optimization-free TTA} methods, which directly calibrate statistics, adjust inputs or outputs without any iterative optimization loop.

\textbf{\textit{Forward-Optimization TTL Methods.}} The core idea is to replace the gradient-based parameter updates with derivative-free optimization strategies that require only forward passes through the model. This reformulates TTA as: given a frozen model $f_{\boldsymbol{\theta}}(\cdot)$ and an unsupervised objective $\mathcal{L}_{ttl}$, find the optimal adaptation variable $\mathbf{z}^*$ by evaluating $\mathcal{L}_{ttl}$ through forward passes alone, without computing $\nabla_{\mathbf{z}} \mathcal{L}_{ttl}$.

FOA~\cite{niu2024foa} first pioneers this paradigm. FOA introduces a small set of learnable prompts to reduce the solution space from millions to a few thousand dimensions, and employs the Covariance Matrix Adaptation Evolution Strategy (CMA-ES)~\cite{auger2012tutorial} for prompt-based derivative-free optimization, with a designed source-target alignment fitness function $\mathcal{L}$ to stabilize the learning process. At each test batch $t$, CMA-ES samples a population of $K$ candidate prompts from:
\begin{equation}
\label{eq:cma_sampling}
    \mathbf{p}_k^{(t)} \sim \mathbf{m}^{(t)} + \tau^{(t)} \mathcal{N}(\mathbf{0}, \boldsymbol{\Sigma}^{(t)}),\quad k = 1, \ldots, K,
\end{equation}
where $\mathbf{m}^{(t)} \in \mathbb{R}^{dN_p}$ is the mean of the search distribution (\ie, the parameters of the model), $\tau^{(t)} \in \mathbb{R}^+$ is the step size, and $\boldsymbol{\Sigma}^{(t)}$ is the covariance matrix that captures the shape of the search ellipsoid. 
Each candidate prompt $\mathbf{p}_k^{(t)}$ is evaluated on the test batch to compute a fitness value. $(\mathbf{m}, \tau, \boldsymbol{\Sigma})$ are then updated based on the ranking of fitness values to favor successful candidates~\cite{auger2012tutorial}. ZOA~\cite{deng2025zoa} alternatively explores the Simultaneous Perturbation Stochastic Approximation (SPSA) for gradient estimation:
\begin{equation}
\label{eq:spsa}
    \hat{g}(\boldsymbol{\theta}) = \frac{\mathcal{L}(\mathbf{x}; \boldsymbol{\theta} + c\boldsymbol{\epsilon}) - \mathcal{L}(\mathbf{x}; \boldsymbol{\theta})}{c} \boldsymbol{\epsilon}^{-1},
\end{equation}
where $c > 0$ is the perturbation scale and $\boldsymbol{\epsilon}$ is a random perturbation vector sampled from a mean-zero distribution (\eg, Rademacher). ZOA demonstrates that SPSA achieves higher learning efficiency than CMA-ES when using only two forward passes per sample, and further introduces a domain knowledge accumulation scheme for zeroth-order continual learning. \rtwo{EVA-0~\cite{chen2026eva0} further stabilizes this strict two-forward setting through a scale-invariant objective, anchor-guided optimization, and sample-wise symmetric perturbations.}

Several works extend the forward-optimization paradigm along various axes. 1) To enhance convergence, FOZO~\cite{wang2026fozo} exploits gradually decaying perturbation scales based on SPSA, whereas PACE~\cite{sojka2026subspace} and ZOTTA~\cite{zhang2026zotta} define subspaces to reduce the optimization dimensionality. \rtwo{CAZO~\cite{zhang2026cazo} complements these approaches by reducing zeroth-order gradient variance through curvature-aware anisotropic perturbation sampling.} 2) To enhance efficiency, SepAMP~\cite{wu2025sepamp} introduces a mixed-precision forward calculation upon FOA. 3) Other studies extend the application scope. B$^2$TPT~\cite{meng2025black} applies black-box prompt tuning to vision-language models by using pseudo-labeling and evolutionary algorithms. E-BATs~\cite{dong2025bats} and BFT~\cite{li2026backpropagation} demonstrate that the forward-optimization paradigm can also be applied to speech and EEG models. Across most methods, the activation discrepancy regularizer introduced by FOA has become a common ingredient for the loss function.

\textbf{\textit{Optimization-Free TTA Methods.}} \revise{Another family adapts writable state without iterative gradient-based optimization and is often categorized into statistics calibration, input purification, and output adjustment.}

For \textit{statistics calibration}, 1) BN adaptation~\cite{schneider2020improving} pioneers replacing the source mean and variance in \textit{BN layers} with statistics computed from test data. Subsequent methods explore strategies such as interpolating source and test statistics~\cite{you2021test}, distribution-aware update~\cite{hong2023mecta,jiang2025feature}, augmentations~\cite{khurana2021sita, gong2022note} to stabilize test-time statistics calibration in diverse settings. Instead, 2) several methods explore statistics calibration in the \textit{feature space}. FOA~\cite{niu2024foa} proposes a back-to-source activation shifting method that replaces the mean of test activations with the source. PEA~\cite{xiao2026architecture} further calibrates the testing mean and variance of each layer's embeddings with source statistics. \rthree{NEO~\cite{murphy2026neo} removes the need for source statistics and iterative optimization by re-centering target embeddings at the latent origin.}

For \textit{input purification}, 1) OST~\cite{termohlen2021continual} and TAF-CAL~\cite{zhao2022test} exploit the \textit{Fast Fourier transform} (FFT)~\cite{brigham2009fast} to decompose inputs into amplitude and phase of frequencies, and replace the test amplitude with the source before reconstruction. Instead, 2) DDA~\cite{gao2023back} explores the ability of \textit{pre-trained diffusion models} to iteratively denoise corrupted test images for removing domain-shift artifacts. Decorruptor-CM~\cite{oh2024efficient} further introduces with a latent diffusion model and trains a distilled variant with consistency distillation to enable faster and fewer denoising steps. CloudFixer~\cite{shim2024cloudfixer} extends diffusion-driven input adaptation to 3D point clouds via geometric transformations guided by a diffusion prior. In contrast, 3) Purge-Gate~\cite{yazdanpanah2025purge} \textit{removes input tokens} most affected by domain shifts based on source statistics to ensure robust inference.

Regarding \textit{output adjustment}, prototype-based methods maintain online-updated class prototypes and replace the original classifier head with a nearest-prototype decision rule at test time~\cite{iwasawa2021test,sui2025just,zhang2025backpropagation}. In contrast, LAME~\cite{boudiaf2022parameter} directly adjusts the output probability distribution to ensure samples with similar features are assigned with consistent pseudo labels.

\subsubsection{\revise{Theoretical Understanding of Test-Time Adaptation}}
\label{sec:tta_theory}

\rtwo{Existing theory does not provide a universal guarantee for test-time adaptation; instead, it characterizes its effectiveness and limitations under specific objectives, feedback mechanisms, update rules, and deployment streams.}

\rtwo{\textit{Proxy--task alignment and objective design.} The original TTT analysis~\cite{sun2020test} gives a local sufficient condition: a small self-supervised update reduces main-task loss when the auxiliary- and main-loss gradients on shared parameters align. Conjugate pseudo-labeling constructs an unlabeled objective from the source loss~\cite{goyal2022test}. In a binary Gaussian model, gradient descent with conjugate pseudo-labels and the squared loss can approach an optimal predictor, whereas hard pseudo-labels can fail~\cite{wang2023understanding}. TIPI~\cite{nguyen2023tipi} bounds target loss through invariance to label-preserving transformations. Stronger guarantees are possible when the test-time signal has task-specific structure: data consistency recovers the optimal estimator under a noise-variance shift in linear denoising~\cite{darestani2022test}, while linear-transformer analysis links TTT gains and sample complexity to pretraining--target alignment~\cite{gozeten2025test}. Together, these results show that a useful proxy must align with the task, source objective, and deployment shift.}

\rtwo{\textit{Mechanism- and feedback-specific bounds.} AdaNPC~\cite{zhang2023adanpc} derives target-error bounds and shows that inserting online target instances into memory can tighten them under locality and label-quality assumptions. ATTA~\cite{gui2024active} studies limited labeled feedback: its VC-based analysis shows that selected test labels can tighten the bound under weighting and budget conditions. These guarantees apply to the analyzed memory- and label-assisted mechanisms, not arbitrary memory updates or unsupervised adaptation.}

\rtwo{\textit{Failure, stability, and protection.} Optimizing a plausible surrogate need not improve task performance. DeYO~\cite{lee2024deyo} shows that entropy can be unreliable under spurious latent factors, whereas the Entropy Enigma~\cite{press2024entropy} shows that prolonged entropy minimization can reverse early gains and degrade accuracy. PeTTA~\cite{hoang2024persistent} relates long-term collapse to pseudo-label error, class structure, and update rate. Complementary analyses use entropy lower bounds~\cite{zhang2025come} or asymmetric updates~\cite{chenzerosiam} to exclude overconfident or collapsed solutions. Another line controls whether and when adaptation proceeds: Protected TTA~\cite{bar2024protected} provides a regret guarantee for the online betting update in its entropy-shift detector, while risk monitoring~\cite{schirmer2025monitoring} uses confidence sequences to control false alarms over time under proxy-informativeness assumptions. Together, these results show that reliable adaptation requires informative signals, controlled updates, and mechanisms for detecting harmful shifts.}

\rtwo{\textit{Adaptation over non-stationary streams.} Gradual-domain-adaptation theory studies pseudo-label updates along ordered domains. Under small consecutive shifts and regularity assumptions, gradual self-training admits a target-error bound~\cite{kumar2020gradual}; later analysis improves its dependence on the number of domains through accumulated-shift terms~\cite{wang2022gradual}. However, this ordered-domain assumption does not cover general TTA streams. For less structured settings, Ada-ReAlign~\cite{zhang2024test} establishes dynamic regret governed by environment variation, while recent work~\cite{zhou2026ttalearnability} formalizes post-shift recovery and long-term reliability using recovery complexity and TTA learnability. These stream-level results expose an adaptivity--information trade-off: reliable recovery requires informative test-time signals and sufficient proxy--task alignment.}

\subsection{Beyond Adaptation: Memory, Personalization, and Self-Improvement}
\label{sec:beyond_adaptation}

\revise{Beyond-adaptation TTL uses deployment experience to improve future behavior rather than only recover source-domain performance. Evolving memory (Section~\ref{sec:memory_augmented}) reuses context or experience for later inputs, personalization (Section~\ref{sec:personalization}) adapts behavior for repeated individual users, environment- or task-specific learning (Section~\ref{sec:env_task_specialization}) specializes a model for recurring operating conditions, and interactive improvement (Section~\ref{sec:interactive_feedback}) \rthree{uses direct user feedback to correct outputs and shape subsequent behavior}. }
\rthree{Recursive self-improvement (Section~\ref{sec:recursive_self_improvement}) extends this progression by modifying the agent, its scaffold, or part of the mechanism that produces subsequent improvements.}
These families emphasize what persists, whose behavior or context is specialized, and which external feedback is used.

\subsubsection{Test-Time Learning with Evolving Memory}
\label{sec:memory_augmented}
This paradigm formulates \rtwo{TTL as the process of storing, organizing, and retrieving information accumulated from test-time observations and feedback signals during inference}, \revise{so that later predictions or actions can reuse information acquired during deployment}. Relevant methods are categorized by their design objectives: memorizing long context or accumulating experience for future action refinement.

\textbf{\textit{TTL for Efficient Context Memorization.}}
Standard Transformers process long contexts via self-attention with quadratic complexity $\mathcal{O}(n^2)$ in sequence length, which becomes prohibitive for long documents and extended conversations~\cite{xiao2023efficient}. TTT-Layer~\cite{sun2024learning} addresses this bottleneck through TTL by proposing that the hidden state of a sequence model can itself be a \emph{learnable model} for context memorization, updated as each new token arrives. Formally, at position $t$, the hidden state $W_t$ is a parameterized model (e.g., a linear model or a small MLP) that is updated by one gradient step:
\begin{equation}
    W_t = W_{t-1} - \eta \nabla_W \ell_{\text{ssl}}(x_t;\, W_{t-1})
\end{equation}
where $\ell_{\text{ssl}}$ is a self-supervised reconstruction objective over the current token $x_t$ for memorization. \revise{The resulting recurrent update has linear sequence complexity $\mathcal{O}(n)$.} 
\rthree{MIRAS~\cite{behrouz2025miras} provides a broader view that characterizes test-time memory through its architecture, internal objective, retention mechanism, and learning algorithm. Recent neural test-time memory architectures can accordingly be organized along four design directions:} 1) \rthree{\textit{Memory objectives and update rules}}~\cite{tandon2025end,hwang2026reinforced,feng2026place} that meta-learns initialization for end-to-end test-time updates~\cite{tandon2025end}, replaces next-token with next-sequence prediction via reinforcement learning~\cite{hwang2026reinforced}, and redesigns fast-weight updates for compatibility with existing LLM architectures such as LLaMA~\cite{feng2026place}, 2) \rthree{\textit{Memory capacity and management}~\cite{behrouz2025titans,behrouz2025atlas,swamynathan2026sr,wang2026allmem} that combines surprise-driven updates and adaptive forgetting in Titans~\cite{behrouz2025titans}, extends token-local optimization to a broader context window in ATLAS~\cite{behrouz2025atlas}}, and integrates neural memory with exact or sliding-window attention for more reliable local-global recall~\cite{swamynathan2026sr,wang2026allmem}, 3) \revise{\textit{Plug-and-play memory}}~\cite{li2026latent} that distills long contexts into compact portable buffer tokens via a disposable LoRA module~\cite{hu2022lora}, and 4) \rtwo{\textit{Modality and policy extensions}}~\cite{liu2026spatial} that adapt TTT layers to streaming video via 3D spatiotemporal
self-supervision. \rtwo{RoboTTT~\cite{jiang2026robottt} further extends fast-weight context memory to robot policies by compressing long visuomotor histories for long-horizon conditioning.}

\textbf{\textit{Evolving Memory for Experience Accumulation.}} A distinct paradigm updates memory to accumulate reusable experience across a stream of tasks
or interactions, enabling the agent to improve over
its deployment lifetime.
1) For \textit{LLM agents}~\cite{suzgun2026dynamic,wei2025evo,cheng2026tame}, 
\rthree{MemoryBank~\cite{zhong2024memorybank} maintains a retrievable interaction history and updates it through time- and recall-dependent forgetting, supporting memory reuse across sessions.}
Dynamic Cheatsheet~\cite{suzgun2026dynamic} endows a black-box language model with a persistent, evolving cheatsheet of problem-solving strategies that is distilled after each task, Evo-Memory~\cite{wei2025evo} formalizes this into a streaming benchmark revealing that most agents lack the dynamic memory management needed for genuine test-time learning, and TAME~\cite{cheng2026tame} introduces a dual-memory architecture with executor and evaluator to prevent degradation. \rthree{ReasoningBank~\cite{ouyang2026reasoningbank} distills generalizable strategies from self-judged successful and failed experiences, while generating diverse interactions as contrastive signals for refining this memory.}
2) For \textit{embodied agents}~\cite{li2026learning,huang2025adapower}, PhysMem~\cite{li2026learning} builds a structured memory of physical principles from robotic interaction via hypothesis generation and verification,
while AdaPower~\cite{huang2025adapower} combines temporal-spatial test-time training with memory persistence to retain task-specific knowledge across episodes.

\subsubsection{\rthree{Recursive Self-Improvement}}
\label{sec:recursive_self_improvement}

\rthree{Recursive self-improvement (RSI) extends TTL by making the agent, or the process that produces later updates writable. Ordinary TTL may repeatedly update memory, skills, or model parameters under a fixed rule. RSI instead forms a closed loop: the system proposes a change, evaluates it, retains successful modifications, and lets the modified system participate in the next cycle. The defining question is whether a retained modification changes how the next improvement is produced. Existing work can be understood as a progression from evolving artifacts under a fixed loop, to modifying the agent itself, and finally to improving the process that generates future modifications.}

\rthree{\textbf{\textit{Evolution under a fixed improvement loop.}} Evaluator-guided search provides the practical starting point for RSI. FunSearch~\cite{romeraparedes2024funsearch} and AlphaEvolve~\cite{novikov2025alphaevolve} repeatedly generate, evaluate, and retain executable programs. ADAS~\cite{hu2025adas} expands the search target from task solutions to agent code and workflows, while Frontis-MA1~\cite{yang2026frontisma1} applies execution-grounded program evolution to machine-learning engineering. These systems progressively enlarge what can be improved, from a solution to an agent design. However, the outer rules for proposing, selecting, and evaluating changes remain largely fixed, making these systems direct precursors to RSI rather than complete instances of it.}

\rthree{\textbf{\textit{Self-modifying agents.}} Existing systems differ in how much of the deployed agent becomes writable during the improvement cycle. STOP~\cite{zelikman2023stop} modifies an LM-based scaffold while leaving the underlying language model fixed. G\"odel Agent~\cite{yin2025godelagent} and SICA~\cite{robeyns2025sica} go further by modifying their runtime logic or agent implementation. The Darwin G\"odel Machine (DGM)~\cite{zhang2026dgm} validates self-modifications empirically and maintains an archive of diverse agents, allowing promising changes to continue along multiple evolutionary paths. The key transition is that the retained result is no longer a solution or memory item, but a modified agent that participates in the next improvement cycle. However, these systems may still rely on a fixed procedure for generating and evaluating modifications.}

\rthree{\textbf{\textit{Improving the improvement process.}} A deeper form of RSI also modifies how later changes are generated. Hyperagents~\cite{zhang2026hyperagents} jointly evolve a task agent and an editable meta-agent responsible for improving it. MetaSkill-Evolve~\cite{wang2026metaskill} similarly separates fast task-skill updates from slower updates to the meta-skill that controls the improvement pipeline. This creates a new evaluation problem: the best current agent is not necessarily the one most capable of producing stronger descendants. The Huxley--G\"odel Machine~\cite{wang2026hgm} captures this distinction through \emph{metaproductivity}, evaluating an agent by the quality of its future evolutionary lineage. Their central challenge is to verify that self-modifications improve subsequent cycles without causing objective drift or performance regression, which requires held-out evaluation, regression testing, version isolation, and rollback.}

\subsubsection{Personalization and User-Specific Learning}
\label{sec:personalization}
The adaptation methods in Section~\ref{sec:tta_adaptation} address distribution shift, where the model is degraded because the test distribution \textit{differs from} training. Personalization considers a different problem: the pretrained model \textit{has been trained} on diverse distributions and works well on average, but still fails to model individual-specific characteristics of a particular user, patient, or instance. Thus, the goal of test-time personalization is to \emph{specialize} a generic model to an individual. We organize existing methods by their target applications.

\textbf{\textit{Language and Content Generation.}} These methods~\cite{Hu2025TLM,bi2025customttt,kim2025instance,hu2025slot} \revise{personalize generative models to user-specified content or preferences}. TLM~\cite{Hu2025TLM} and SLOT~\cite{hu2025slot} perform next-token-prediction learning on user input prompts to better understand complex instructions and \revise{capture the user's linguistic preferences}. CustomTTT~\cite{bi2025customttt} jointly customizes motion and appearance from a single reference for video generation. TTT-Editor~\cite{kim2025instance} personalizes speech editing models to better preserve acoustic consistency and speaker identity.

\textbf{\textit{Human-Centered Perception and Recognition.}} Human-centered perception and recognition refers to tasks that infer human attributes, states, or individual-specific signals, such as body pose~\cite{li2021test,cui2023test}, gaze~\cite{wu2024ttagaze,liu2024test}, speech~\cite{shi2024examining}, lip movements~\cite{shen2024speaker}, handwriting~\cite{gu2025metawriter}, and daily activities~\cite{wang2024optimization}. Representative personalization strategies include self-supervised personalization to individual body proportions~\cite{li2021test,cui2023test}, meta-learned initializations for rapid per-user gaze and handwriting calibration~\cite{wu2024ttagaze,liu2024test,gu2025metawriter}, and pseudo-label-driven speaker adaptation for ASR and lip reading~\cite{shi2024examining,shen2024speaker}.

\textbf{\textit{Biomedical Applications.}} 
Personalization is critical in biomedicine due to inter-patient variability in anatomy, physiology, and pathology. Existing works personalize medical models across diverse granularities, including patient-specific segmentation via foundation-model pseudo-labels and \revise{information-geometric adaptation}~\cite{ravishankar2024tta, ravishankar2025information}, personalized brain-computer interfaces that eliminate costly per-subject EEG calibration~\cite{duan2025test,peng2025test}, adaptation to individual protein samples via masked protein reconstruction~\cite{bushuiev2024one}, and patient-aware clinical time series modeling through test-time adaptive mixture-of-experts~\cite{zhu2025tamer}.

\subsubsection{Environment-Specific and Task-Specific Specialization}
\label{sec:env_task_specialization}

Specialization aims to tailor a general-purpose model (\eg, a foundation model) to a particular deployment context, unlocking the potential that the general-purpose model already possesses, \revision{often with limited environment/task-specific data.}

\textbf{\textit{Autonomous Systems and Robotics.}}
Deploying general-purpose perception and action models in specific physical environments requires specialization to local geometry, dynamics, and task semantics. Representative works include terrain-specific locomotion adaptation for \textit{humanoid robots} to unseen parkour obstacles via test-time training on reconstructed geometry~\cite{zhu2026ttt}, \textit{world model} specialization that trains on the current scene to improve planning accuracy~\cite{wan2026worldagen,huang2025adapower}, policy-level adaptation that updates \textit{VLA models} via step-by-step environmental feedback with test-time reinforcement learning~\cite{liu2026fly}, reflective trial-and-error planning that learns from execution failures for \textit{embodied LLMs} during testing~\cite{hong2026learning}, and learning object-specific grasping skills at test time through embodied exploration~\cite{liu2025embodied}.

\textbf{\textit{Industrial and Scientific Environments.}}
Industrial and scientific deployments often involve unique equipment configurations, operating conditions, or physical phenomena. Test-time specialization enables models to self-calibrate to these contexts, including fault detection under evolving operating conditions in industrial systems~\cite{sun2026tard,tian2025continual,tian2025test,goodarzi2025practical}, underwater acoustic source localization to unseen ocean environments~\cite{kari2026joint}, semiconductor recipe generation adapted to specific fabrication equipment~\cite{gu2025few}, transient electromagnetic signal denoising specialized to new geological sites~\cite{yang2025dp}, and geospatial model adaptation across geographic regions through test-time multimodal reconstruction~\cite{gordon2026mmearth}.

\textbf{\textit{Foundation Model Task Specialization.}}
The rapid proliferation of foundation models creates a new and dominant use case: specializing a single general-purpose model to a specific downstream task at test time, without task-specific fine-tuning data. For vision-language models, test-time prompt tuning~\cite{shu2022test} and its variants~\cite{yoon2024c,sarkar2024active} specialize CLIP-style models to target visual domains by optimizing continuous prompts on unlabeled test data. For large language models, test-time specialization adapts general LLMs to specific domains through steering-vector-based activation manipulation~\cite{kang2026model}, retrieval-augmented TTT to better predict domain-specific content~\cite{sun2026predict}, and \revise{next-token-prediction learning} on inputs to incentivize reasoning~\cite{Hu2025TLM,hu2025slot}.

\subsubsection{\rthree{Interactive Improvement from User Feedback}}
\label{sec:interactive_feedback}
\rthree{Interactive improvement uses feedback supplied directly by human users to correct the active system or align its behavior with deployment-specific intent. Such feedback ranges from low-bandwidth evaluations to structured corrections that indicate how an output should change.}

\rthree{\textbf{\textit{Evaluative feedback.}} Binary judgments or sparsely queried labels provide a relatively inexpensive signal about whether an output is acceptable. BiTTA~\cite{lee2025test} learns from binary correct/incorrect feedback while balancing human-guided correction with self-adaptation on confident predictions. Active TTA~\cite{gui2024active} and TAPS~\cite{sarkar2025taps} instead query labels for selected uncertain samples to increase the adaptation benefit of each annotation. These signals reduce annotation effort, but provide limited information about how an incorrect output should be revised.}

\rthree{\textbf{\textit{Corrective and intent-bearing feedback.}} More structured interactions can localize an error or specify the desired result. HiTTA~\cite{hu2024towards} and ITTA~\cite{cao2024interactive} incorporate clinician corrections into medical-image adaptation, while SAM adaptation~\cite{schon2024adapting} learns from user scribbles and clicks. SplatPainter~\cite{zheng2025splatpainter} further uses user-provided 2D edits to update a Gaussian representation within an interactive 3D editing episode. Structured corrections provide more targeted supervision, but require task-specific interfaces and greater user effort. User feedback is one way to close the test-time learning loop. Other signals from environments, tools, and verifiers are also reviewed in Section~\ref{sec:external-feedback}. Their role depends on how they are used: verifier feedback that only selects the current output belongs to TTS (Section~\ref{sec:verifier-guidance}), whereas rollout-derived feedback that updates the active model is discussed under test-time reinforcement learning (Section~\ref{sec:test-time-rl}).}

\section{Test-Time Scaling}\label{tts}

\subsection{Problem Formulation}
Test-time scaling (TTS) refers to inference-time methods that improve model performance by allocating additional, controllable computation while keeping the pretrained model parameters frozen. Formally, let $f_{\boldsymbol{\theta}}:\mathcal{X}\rightarrow\mathcal{Y}$ denote a pretrained model with fixed parameters $\boldsymbol{\theta}$, where $\mathcal{X}$ and $\mathcal{Y}$ are the input and output spaces, respectively. For an input $x\in\mathcal{X}$, standard inference follows a fixed procedure $\mathcal{I}_0(\cdot)$:
\begin{equation}
\hat{y}_{\text{std}}=\mathcal{I}_0(f_{\boldsymbol{\theta}},x),
\end{equation}
Here, $\mathcal{I}_0$ uses a fixed decoding or decision rule at a baseline inference budget. TTS extends this fixed-budget procedure to a scalable inference function:
\begin{equation}
\hat{y}_{\text{tts}}=\mathcal{F}(f_{\boldsymbol{\theta}},x;\mathbf{c}),
\end{equation}
where $\mathbf{c}\in\mathcal{C}$ specifies a test-time compute budget or resource configuration, such as the number of samples, reasoning steps, search depth, verification rounds, or tool calls. Setting $\mathbf{c}=\mathbf{c}_0$ recovers standard inference at the baseline budget $\mathbf{c}_0$. In this survey, TTS concerns how additional computation beyond this baseline is controlled to improve the current inference process while $\boldsymbol{\theta}$ remains unchanged.

This formulation makes inference-time compute an additional scaling axis alongside model scaling. Empirical studies show that repeated sampling and compute-optimal allocation can substantially improve reasoning performance~\cite{snell2025scaling,brown2024large,wu2024comparative}; under matched or optimized inference-compute settings, smaller models can sometimes outperform much larger ones~\cite{snell2025scaling,liu2025can}. These findings suggest that the value of TTS depends not only on how much computation is added, but also on how it is used. Throughout this section, the model parameters remain frozen and the resulting improvement is realized within the active inference process.

\begin{figure*}[t]
\centering
\includegraphics[width=0.97\linewidth]{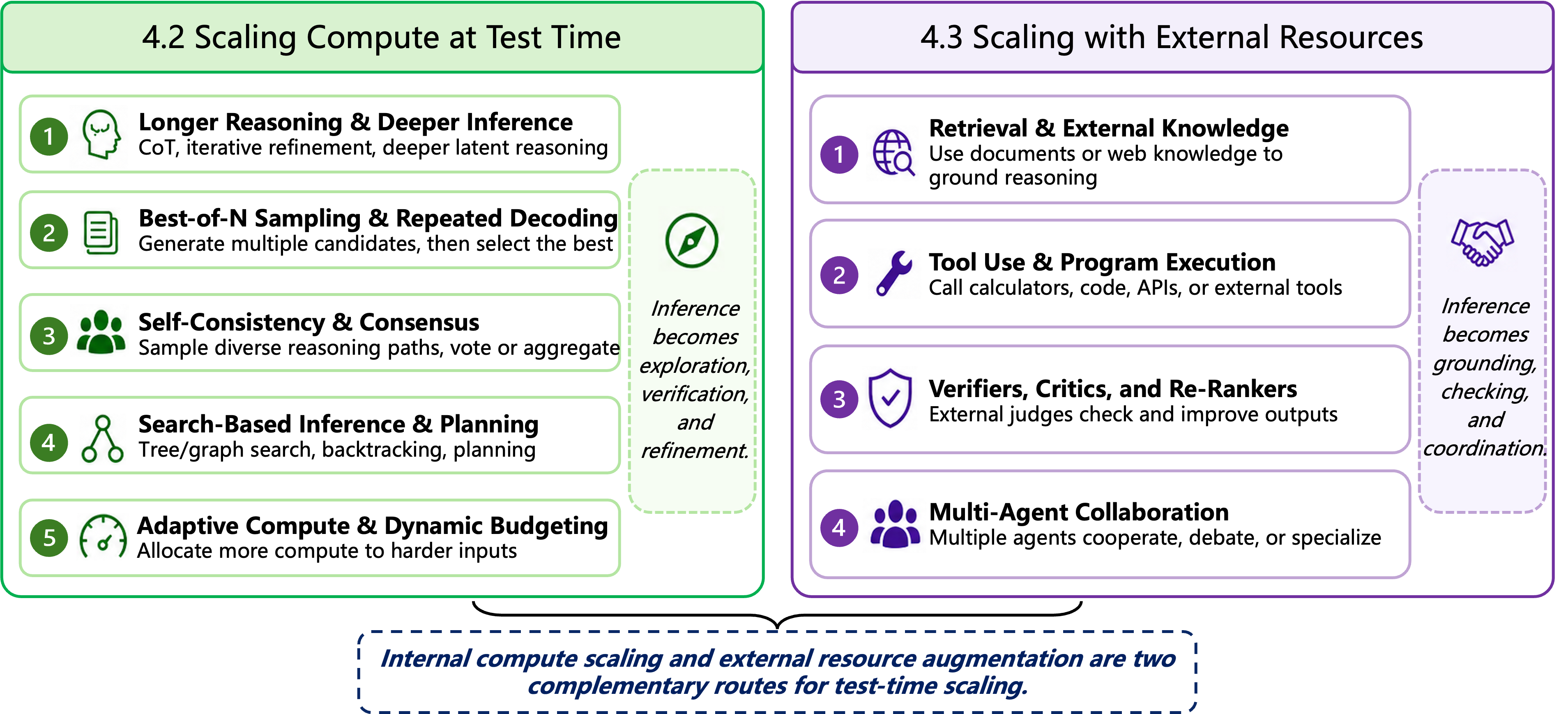}
\caption{Two major test-time scaling axes for improving inference-time performance without changing pretrained model parameters.}
\label{fig:tts-axes}
\end{figure*}

\subsection{Scaling Compute at Inference Time}

Additional inference-time computation can be used to extend one reasoning trajectory, generate multiple complete candidates, or explore a structured space of intermediate states. Adaptive methods complement these mechanisms by controlling how much computation each input or reasoning stage receives. Longer or iterative reasoning is natural when a solution can be improved step by step, whereas repeated sampling is useful when diverse complete answers can be generated and compared. Consensus offers a simple selection rule when answers can be matched reliably, while search is better suited to problems that support meaningful evaluation of partial states, lookahead, or backtracking. Adaptive control can further reduce average cost when input difficulty or the marginal value of computation can be estimated reliably. Table~\ref{tab:tts-compute-scaling-comparison} summarizes the resource requirements, applicable scenarios, and common failure modes of these families.

\begin{table*}[!t]
\centering
\caption{\revise{Decision-oriented comparison of inference-time compute-scaling families.}}
\label{tab:tts-compute-scaling-comparison}
\footnotesize
\setlength{\tabcolsep}{2pt}
\renewcommand{\arraystretch}{1.08}
\setlength{\emergencystretch}{3em}
\resizebox{\textwidth}{!}{%
\begin{tabularx}{1.18\textwidth}{@{}
>{\raggedright\arraybackslash}p{0.154\textwidth}
>{\raggedright\arraybackslash}p{0.201\textwidth}
>{\raggedright\arraybackslash}p{0.201\textwidth}
>{\raggedright\arraybackslash}p{0.13\textwidth}
>{\raggedright\arraybackslash}p{0.213\textwidth}
>{\raggedright\arraybackslash}X@{}}
\toprule
\revise{\textbf{Family}} & \revise{\textbf{Representative Works}} & \revise{\textbf{Required Resources}} & \revise{\textbf{Cost}} & \revise{\textbf{Applicable Scenarios}} & \revise{\textbf{Failure Modes}} \\
\midrule
\revise{Iterative Reasoning} & \revise{CoT~\cite{wei2022chain}; Self-Refine~\cite{madaan2023self}; Recurrent Depth~\cite{geiping2025scaling}} & \revise{Long context or recurrent state; stopping/self-feedback policy} & \revise{Moderate--high tokens; sequential latency} & \revise{Decomposable or open-ended tasks benefiting from stepwise refinement} & \revise{Early errors propagate; excessive depth causes overthinking} \\
\midrule
\revise{Best-of-$N$ Sampling} & \rtwo{Verifier-guided BoN~\cite{cobbe2021training}; LLMonkeys~\cite{brown2024large}; PairJudge RM~\cite{liu2025pairjudge}; MBR-BoN~\cite{jinnai2025regularized}} & \revise{Parallel decoding and a reliable selector} & \revise{High decoding/ranking cost, roughly proportional to $N$} & \revise{Parallelizable tasks with rankable outputs} & \revise{Biased selectors or low diversity make additional samples unhelpful} \\
\midrule
\revise{Consensus} & \revise{Self-Consistency~\cite{wang2023self}; CISC~\cite{taubenfeld2025confidence}; DSC~\cite{wang2025make}} & \revise{Multiple stochastic samples and an answer-equivalence rule} & \revise{Moderate--high decoding; low-cost aggregation} & \revise{Stable answer spaces with diverse reasoning paths} & \revise{Correlated samples produce confident majority errors} \\
\midrule
\revise{Search and Planning} & \rtwo{ToT~\cite{yao2023tree}; RAP~\cite{hao2023rap}; LATS~\cite{zhou2024lats}; ETS~\cite{hooper2025ets}} & \revise{Intermediate-state representation, evaluator, and frontier memory} & \revise{Very high, irregular compute, memory, and latency} & \revise{Planning or combinatorial tasks with evaluable partial states} & \revise{Noisy evaluators prune promising branches; branching exhausts the budget} \\
\midrule
\revise{Adaptive Computation} & \revise{Compute-optimal scaling~\cite{snell2025scaling}; Meta-Reasoner~\cite{sui2025meta}; token-budget-aware reasoning~\cite{han2025token}} & \revise{Difficulty or uncertainty signal and budget controller} & \revise{Low control overhead; variable per-instance compute} & \revise{Mixed-difficulty or anytime workloads} & \revise{Difficulty errors underallocate compute to hard inputs} \\
\bottomrule
\end{tabularx}}
\end{table*}

\subsubsection{Longer Reasoning and Deeper Inference}

The most direct way to spend additional inference compute is to extend the reasoning performed before an answer is produced. \rthree{Although many methods in this category, particularly CoT-style approaches, do not rely on explicit feedback and thus fall outside our definition of feedback-driven TTI, they represent an important paradigm of test-time scaling (TTS) and are therefore briefly reviewed here.} CoT prompting~\cite{wei2022chain} showed that intermediate reasoning steps can substantially improve arithmetic, commonsense, and symbolic reasoning without test-time parameter updates. Yeo~\textit{et al.}~\cite{yeo2025demystifying} further study how long reasoning traces emerge, finding that reinforcement-learning incentives can stabilize their growth and elicit error-correction behaviors already present in the base model.

Beyond extending an initial trace, extra computation can be used to revisit and improve an existing answer. Self-Refine~\cite{madaan2023self} alternates self-feedback with revision, turning a single response into a multi-stage correction process. A related line increases effective depth without requiring an equally long visible trace: looped transformers~\cite{saunshi2025reasoning} repeatedly apply a recurrent block, recurrent-depth models~\cite{geiping2025scaling} reason through latent iterations, and filler-token experiments~\cite{pfau2024lets} show that transformers can use otherwise uninformative tokens for hidden computation. \revise{Together, these approaches are attractive when additional computation can productively refine a single evolving solution, especially when generating and ranking many alternatives is impractical.} However, more depth is not invariably beneficial. Wu~\textit{et al.}~\cite{wu2026more} observe an inverted U-shaped relation between reasoning length and acc, showing that early mistakes can propagate and excessive reasoning can lead to overthinking.

\rthree{Beyond simply extending a single trace, structured prompting offers complementary ways to organize additional inference-time computation: Least-to-Most~\cite{zhou2023leasttomost} solves a decomposition of simpler subproblems sequentially, AoT~\cite{sel2024aot} injects algorithmic search patterns in in-context demonstrations, and SELF-DISCOVER~\cite{zhou2024selfdiscover} composes a task-specific reasoning structure before decoding.}

\subsubsection{Best-of-N Sampling and Repeated Decoding}

Best-of-$N$ (BoN) sampling is a simple way to scale inference: it generates $N$ candidate outputs and selects one according to a scoring criterion. \rtwo{Cobbe~\textit{et al.}~\cite{cobbe2021training} established a canonical verifier-guided form by generating multiple mathematical solutions and using a learned outcome verifier to rank them.} Brown~\textit{et al.}~\cite{brown2024large} show that the fraction of problems solved by at least one sample scales approximately log-linearly with the sampling budget across several verifiable domains. Repeated sampling can therefore increase the chance that a correct solution is available, but it does not by itself ensure that this solution will be selected.
\rthree{An early code-generation example is AlphaCode~\cite{li2022alphacode}, which scales inference through large-scale program sampling followed by behavior-based filtering and clustering to produce a small submission set.}
The practical value of BoN therefore depends on how candidates are compared. Self-certainty~\cite{kang2026scalable} estimates response quality from the model's output distribution without an external reward model. PairJudge RM~\cite{liu2025pairjudge} instead uses pairwise judgments and a knockout tournament, while MBR-BoN~\cite{jinnai2025regularized} regularizes reward-model selection to reduce reward hacking. The accompanying decoding cost can also be reduced: Speculative Rejection~\cite{sun2024fast} terminates unpromising candidates early while preserving high-reward selection. \revise{Overall, increasing $N$ broadens candidate coverage, but turns that coverage into final-answer accuracy only when the candidates are sufficiently diverse and the selector is reliable.}

\begin{figure}[]
  \centering
  \includegraphics[width=0.7\linewidth]{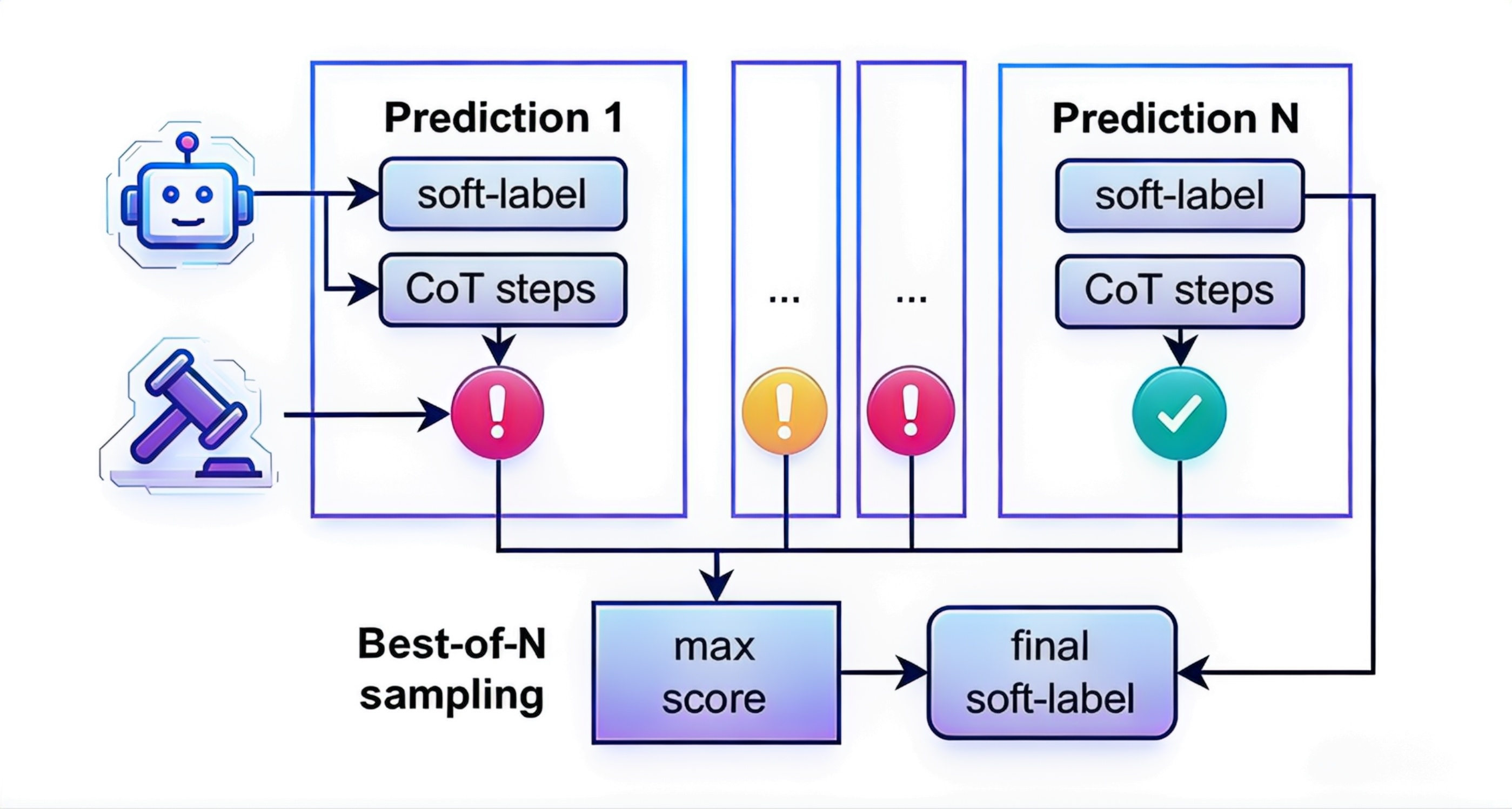}
  \caption{An illustration of the Best-of-N sampling process, where multiple predictions with reasoning chains are generated and the one with the highest score is selected as the final output.}
  \label{fig:Best_of_N}
\end{figure}

\subsubsection{Self-Consistency and Consensus Mechanisms}

Whereas BoN explicitly ranks its candidates, consensus methods primarily use agreement among sampled answers as the selection signal. The basic form, self-consistency~\cite{wang2023self}, marginalizes over diverse reasoning paths and returns the most frequent answer. \rthree{Subsequent variants soften this hard voting mechanism: CISC~\cite{taubenfeld2025confidence} weights samples by model-generated confidence, while Soft-SC~\cite{wang2024softsc} uses continuous likelihood-based scores to better aggregate sparse agreement in long-horizon agent tasks, both improving sampling efficiency.}

Agreement, however, is not a universal proxy for correctness. Chen~\textit{et al.}~\cite{chen2024more} show that majority-vote performance can first improve and then decline as the number of calls grows, while Nguyen~\textit{et al.}~\cite{nguyen2024consistent} find that reasoning length can be more informative than answer frequency in some settings. Whereas these works examine how sampled answers should be aggregated, difficulty-adaptive self-consistency (DSC)~\cite{wang2025make} addresses the complementary question of how many samples to draw by adjusting the sampling budget according to predicted difficulty. \revise{Overall, consensus is attractive when answers can be matched reliably, but correlated samples or ambiguous answer equivalence can turn additional votes into a confident shared error.}

\subsubsection{Search-Based Inference and Planning}

Repeated sampling treats complete candidates separately, whereas search retains the relations among partial states and uses intermediate evaluation to decide what to explore next. This structure supports deliberate lookahead, revision, and backtracking rather than selecting only after complete solutions have been generated. ToT~\cite{yao2023tree} organizes alternatives as a tree, while GoT~\cite{besta2024graph} allows more general graph relations, including the aggregation, reuse, and refinement of earlier thoughts.

Once a search space has been defined, an evaluator or constraint can guide which states are expanded. \rtwo{RAP~\cite{hao2023rap} evaluates MCTS branches through an LM world model and task rewards, whereas LATS~\cite{zhou2024lats} combines tree search with self-reflection and external environment feedback for interactive decision making.} TS-LLM~\cite{feng2023alphazero} learns a value function to guide tree-search decoding, while AlphaMath~\cite{chen2024alphamath} couples MCTS with a value model and uses the resulting search trajectories to derive step-level supervision. CMCTS~\cite{lin2026cmcts} instead restricts the action space and imposes partial-order rules to stabilize deep mathematical search. Maintaining and evaluating a search frontier can itself be expensive. ETS~\cite{hooper2025ets} reduces this overhead by pruning redundant trajectories and sharing KV caches while preserving semantic diversity. \revise{Search is most useful for planning or combinatorial problems whose partial states can be evaluated meaningfully. Its benefits can nevertheless disappear when noisy evaluators discard promising branches, or when branching, frontier memory, and sequential control exhaust the available budget.}

\subsubsection{Adaptive Computation and Dynamic Budgeting}

Not all inputs require the same amount or form of inference compute. Adaptive computation therefore acts as a control layer over the preceding mechanisms, deciding how much reasoning depth, sampling, or search each input receives rather than defining another way to generate answers. Compute-optimal scaling~\cite{snell2025scaling} selects among search and refinement strategies according to estimated prompt difficulty, and Liu~\textit{et al.}~\cite{liu2025can} show that effective allocation can allow smaller models to outperform substantially larger ones. DSC~\cite{wang2025make} similarly uses prior and posterior difficulty estimates to avoid unnecessary consensus samples on easier questions.

\rthree{Allocation can also operate directly on the reasoning trajectory.} Meta-Reasoner~\cite{sui2025meta} uses a contextual bandit to decide whether to continue, backtrack, switch strategies, or restart. Other methods control the length of the trajectory directly: \rthree{s1~\cite{muennighoff2025s1} introduces budget forcing, truncating generation at a prescribed budget or extending reasoning with repeated “Wait” tokens; LCPO~\cite{aggarwal2025l1} further learns to satisfy user-specified reasoning-length constraints through reinforcement learning, replacing heuristic budget forcing with learned length control,} while Han~\textit{et al.}~\cite{han2025token} adjust reasoning-token budgets according to problem complexity. \revise{Such controllers can reduce average cost across mixed-difficulty workloads, but they depend on informative difficulty or uncertainty estimates. When these estimates are poor, the controller may waste computation on easy inputs and stop too early on difficult ones.}

\subsubsection{\revise{Theoretical Understanding of Test-Time Scaling}}
\label{sec:tts_theory}

\rtwo{Additional inference-time computation does not by itself guarantee better reasoning. Current theoretical analyses are concentrated on language-model reasoning and provide conditional guarantees or empirical scaling laws for three questions: how sampling changes reasoning error, when verification converts candidate coverage into accuracy, and how computation should be allocated across prompts or reasoning stages.}

\rtwo{\textit{Sampling, confidence, and diversity.} RPC~\cite{zhou2025rpc} decomposes reasoning error into estimation error from finite sampling and confidence estimation, and model error from limitations of the base model that more samples alone cannot remove. It shows that self-consistency has relatively slow estimation-error convergence, whereas combining consensus with internal probabilities can improve this convergence from linear to exponential; pruning low-probability trajectories further controls model error. Thus, sampling addresses only the estimation component, and its gains can saturate when correct trajectories are rare, effective diversity is low, or confidence is misaligned with correctness.}

\rtwo{\textit{Verification and selection.} A complementary line asks when generated candidates can be converted into reliable gains. Huang~\textit{et al.}~\cite{huang2025bestofn} analyze BoN under an imperfect reward model, showing that its optimality requires strong policy coverage and that increasing $N$ can worsen reward hacking; their pessimistic alternative restores scaling monotonicity under the stated assumptions. Setlur~\textit{et al.}~\cite{setlur2025scaling} analyze the related question of training models to use additional test-time computation. Under heterogeneous correct-trace distributions and reward anti-concentration assumptions, they show that verifier-free imitation can have worse asymptotic suboptimality than verification-based reinforcement learning or search as the reasoning horizon and data budget grow. These results do not imply that an arbitrary verifier guarantees improvement: they rely on informative rewards and a policy that can exploit them. They instead distinguish candidate generation from reliable selection or reinforcement.}

\rtwo{\textit{Compute-optimal allocation and scaling laws.} Snell~\textit{et al.}~\cite{snell2025scaling} empirically compare verifier-guided search and sequential revision under fixed inference budgets. The preferred strategy depends on prompt difficulty and base-model capability: easier problems may benefit from sequential refinement, whereas harder but solvable problems may require broader search. Complementing this evidence, Plan-and-Budget~\cite{lin2026planbudget} models reasoning as subproblems with different uncertainty levels and explains how fixed token budgets can cause overthinking on easy steps and underthinking on difficult ones. These results motivate difficulty- and stage-aware allocation but depend on reliable difficulty estimates, verifiers or revision policies, and a capable base model. Overall, test-time scaling is conditional rather than monotonic: additional computation helps only when the model produces useful candidates and the system can evaluate and allocate them reliably. Comparable guarantees for external-resource scaling, multi-agent inference, and non-language modalities remain limited.}

\begin{table*}[!t]
\centering
\caption{\revise{Decision-oriented comparison of external-resource scaling families.}}
\label{tab:tts-external-resource-comparison}
\footnotesize
\setlength{\tabcolsep}{2pt}
\renewcommand{\arraystretch}{1.08}
\setlength{\emergencystretch}{3em}
\resizebox{\textwidth}{!}{%
\begin{tabularx}{1.18\textwidth}{@{}
>{\raggedright\arraybackslash}p{0.154\textwidth}
>{\raggedright\arraybackslash}p{0.201\textwidth}
>{\raggedright\arraybackslash}p{0.201\textwidth}
>{\raggedright\arraybackslash}p{0.13\textwidth}
>{\raggedright\arraybackslash}p{0.213\textwidth}
>{\raggedright\arraybackslash}X@{}}
\toprule
\revise{\textbf{Family}} & \revise{\textbf{Representative Works}} & \revise{\textbf{Required Resources}} & \revise{\textbf{Cost}} & \revise{\textbf{Applicable Scenarios}} & \revise{\textbf{Failure Modes}} \\
\midrule
\revise{Retrieval} & \revise{Self-RAG~\cite{asai2023self}; IRCoT~\cite{trivedi2023interleaving}; Adaptive-RAG~\cite{jeong2024adaptive}}; \rthree{FLARE~\cite{jiang2023flare}; Search-R1~\cite{jin2025searchr1}} & \revise{Retriever plus maintained corpus, index, or API} & \revise{Retrieval and reranking calls, index memory, context, and latency} & \revise{Knowledge-intensive or time-sensitive tasks with authoritative evidence} & \revise{Missing evidence leaves claims unsupported; poor evidence causes false grounding} \\
\midrule
\revise{Tool Use} & \revise{ReAct~\cite{yao2023react}; CRITIC~\cite{gou2024critic}; XoT~\cite{Liu2023plan}; TTE~\cite{lu2026beyond}} & \revise{Tool schemas, execution access, permissions, and runtime feedback} & \revise{Variable tool-call, execution, and debugging latency} & \revise{Executable verification, feedback-based correction, or dynamic tool construction} & \revise{Wrong tools or arguments cause failures or unsafe side effects} \\
\midrule
\revise{Verifier Guidance} & \revise{Process reward models~\cite{lightman2023let}; GenRM~\cite{zhang2024generative}; CLoud~\cite{ankner2024critique}; MAV~\cite{lifshitz2025multi}} & \revise{Candidate or intermediate-state access and a reliable evaluator} & \revise{Moderate--very high from repeated evaluator calls} & \revise{Selection or guided search with externally assessable correctness} & \revise{Evaluator bias favors wrong candidates; verification cannot recover absent solutions} \\
\midrule
\revise{Multi-Agent Inference} & \revise{MoA~\cite{wang2024mixture}; ReConcile~\cite{chen2024reconcile}; MacNet~\cite{qian2024scaling}; MAD~\cite{liang2024multiagentdebate}} & \revise{Multiple models or roles, communication, and an aggregator} & \revise{High--very high model-call, token, and coordination cost} & \revise{Decomposable or open-ended tasks benefiting from complementary expertise} & \revise{Correlated agents reinforce shared errors; coordination erases diversity gains} \\
\bottomrule
\end{tabularx}}
\end{table*}

\subsection{Scaling with External Resources}

The methods above improve inference by extending the model's own generation, sampling, or search. However, additional internal computation cannot by itself supply unavailable evidence, execute exact operations, or provide an independent assessment of candidate quality. TTS can therefore also couple the frozen model with external resources that broaden the information and capabilities available during inference.

The appropriate resource depends on the bottleneck being addressed. \textbf{Retrieval and external knowledge} provide missing, current, or domain-specific evidence when an accessible corpus is available, although irrelevant or conflicting evidence can mislead generation~\cite{asai2023self,trivedi2023interleaving}. \textbf{Tool use and program execution} support precise computation or executable actions, but depend on reliable interfaces and correct tool invocation~\cite{chen2023program,gou2024critic}. \textbf{Verifiers, critics, and re-rankers} improve selection or refinement when their judgments correlate with correctness, whereas biased evaluators may favor incorrect trajectories~\cite{lightman2023let,zhang2024generative}. \textbf{Multi-agent inference} introduces alternative proposals, interaction, or specialized roles when diverse perspectives are useful, but correlated errors and communication overhead can offset its gains~\cite{wang2024mixture,qian2024scaling}. These families can be combined; in practice, missing evidence, unreliable execution, weak evaluation, or insufficient reasoning diversity should guide the resource choice. Table~\ref{tab:tts-external-resource-comparison} provides a deployment-oriented comparison.

\subsubsection{Retrieval and External Knowledge}

\revise{Retrieval is relevant to feedback-driven TTI when it is invoked adaptively or coupled iteratively with reasoning, rather than merely supplying fixed context. Self-RAG~\cite{asai2023self} retrieves passages on demand and uses reflection tokens to assess retrieved evidence and generated responses, while FLARE~\cite{jiang2023flare} triggers retrieval from low-confidence predictions of upcoming content. IRCoT~\cite{trivedi2023interleaving} forms a more explicit retrieval--reasoning loop, where intermediate reasoning steps generate new queries and retrieved evidence informs subsequent reasoning. Adaptive-RAG~\cite{jeong2024adaptive} instead controls retrieval depth by selecting among no retrieval, single-step retrieval, and iterative retrieval according to question complexity. Together, these methods illustrate reflection-guided evidence use, uncertainty-triggered retrieval, iterative retrieval--reasoning interaction, and adaptive allocation of retrieval effort.}

\rthree{More recent methods extend this interaction toward agentic and structured evidence search. Search-o1~\cite{li2025searcho1} invokes search at knowledge-uncertain points during reasoning and explicitly reasons over retrieved documents, while Search-R1~\cite{jin2025searchr1} learns a multi-turn search policy that interleaves retrieval with stepwise reasoning. MIRAGE~\cite{wei2026mirage} further structures evidence acquisition through dynamic knowledge-graph retrieval, parallel reasoning paths, and cross-path verification for traceable medical question answering. These approaches allow inference-time effort to scale through additional search rounds, evidence paths, or verification, but remain sensitive to missing, irrelevant, or conflicting evidence and may incur substantial retrieval latency.}

\subsubsection{Tool Use and Program Execution}

Language models may reason about an operation yet still execute it unreliably through generation alone. External tools address this gap by delegating selected steps to external executors. \rtwo{They can also participate throughout a trajectory: ReAct~\cite{yao2023react} interleaves reasoning, actions, and observations so that tool or environment feedback informs later decisions.} Program of Thoughts (PoT)~\cite{chen2023program}, for example, generates a program whose execution produces the final answer, separating semantic reasoning from numerical computation. START~\cite{li2025start} uses lightweight hints to elicit tool calls for long-chain reasoning, computation, and self-debugging at inference time, although the full method also includes a self-training stage.

Tool outputs can serve not only as final answers, but also as feedback for checking and revising a model's reasoning. CRITIC~\cite{gou2024critic} uses search engines or code interpreters to validate parts of an initial response and revise it using execution feedback, while XoT~\cite{Liu2023plan} switches among reasoning strategies such as CoT and PoT when an external executor detects failure. Beyond fixed interfaces, Test-Time Tool Evolution (TTE)~\cite{lu2026beyond} synthesizes, verifies, and refines executable tools during inference for scientific reasoning. Unless explicitly written into memory or model state, tool outputs and corrections remain local to the current episode. \revise{Tool use is most effective when reliable executable interfaces are available; incorrect tool selection or invocation can otherwise offset its gains.}

\subsubsection{Verifiers, Critics, and Re-Rankers}
\label{sec:verifier-guidance}

Generating several candidates and recognizing the best one are distinct problems. Verifiers and critics address the latter by evaluating complete outputs or intermediate reasoning states, allowing additional candidates to support better selection, refinement, or search guidance. \rtwo{They may score final outcomes, as in verifier-guided BoN~\cite{cobbe2021training}, or provide step-level feedback, as in process reward models~\cite{lightman2023let}, which can rerank or guide reasoning paths.} AlphaMath~\cite{chen2024alphamath}, discussed with search-based inference above, illustrates how a learned value model can supply such intermediate signals within Monte Carlo Tree Search.

The evaluator need not be a single scalar-scoring model; it can itself reason or aggregate multiple judgments. GenRM~\cite{zhang2024generative} formulates verification as next-token prediction and aggregates verifier reasoning through majority voting, while Critique-out-Loud (CLoud)~\cite{ankner2024critique} produces a textual critique before assigning a scalar score. Multi-Agent Verification (MAV)~\cite{lifshitz2025multi} takes a complementary approach by combining multiple off-the-shelf verifiers with best-of-\(N\) sampling. \revise{These mechanisms are useful when correctness can be assessed more reliably than it can be generated. However, evaluation cannot recover a good solution that is absent from the candidate set, and a misaligned evaluator may still favor incorrect trajectories.}

\subsubsection{Multi-Agent Inference and Collaborative Reasoning}

When a single inference trajectory provides insufficient diversity or expertise, multiple agents can contribute alternative proposals and interact before producing a final answer. Agent Forest~\cite{li2024more} samples and votes across agents, while Mixture-of-Agents (MoA)~\cite{wang2024mixture} lets each layer aggregate outputs from the preceding layer. Agents can also revise their views after observing others: ReConcile~\cite{chen2024reconcile} exchanges answers and confidence scores to form a confidence-weighted consensus, whereas Multi-Agent Debate (MAD)~\cite{liang2024multiagentdebate} uses adversarial discussion to preserve divergent reasoning and mitigate degeneration of thought.

\rthree{A complementary line structures how agents communicate, specialize, or access external resources}. MacNet~\cite{qian2024scaling} arranges agents in directed acyclic graphs and observes saturating logistic scaling as the network grows, while METAL~\cite{li2025metal} assigns specialized roles for chart generation and improves with additional computational budget. \rthree{TUMIX~\cite{chen2025tumix} further combines agent specialization with tool diversity, running heterogeneous tool-equipped agents in parallel and iteratively sharing and refining their responses until sufficient confidence is reached}. \revise{Across these designs, gains arise from complementary information, expertise, \rthree{tools}, or error patterns rather than agent count alone. When agents make correlated errors, converge prematurely, or incur excessive communication and aggregation costs, adding agents may provide little benefit or even degrade performance.}

\subsection{What Scaling Improves and Enables}

The preceding sections describe how additional compute and external resources are used during inference. This section instead examines what these mechanisms improve or enable beyond end-task accuracy. Stronger reasoning~\cite{wang2023self}, more reliable selection and uncertainty assessment~\cite{gou2024critic}, and better decisions in control and planning~\cite{chen2025armap} are outcomes realized during active inference. In contrast, reasoning traces and search results can also be retained as artifacts for downstream learning~\cite{sessa2025bond}, with the resulting benefits realized through later amortization.

\subsubsection{Reasoning Quality}

Reasoning quality is one of the most direct benefits of test-time scaling. By moving beyond one-shot, single-path decoding, additional computation can expose intermediate steps and explore alternative reasoning trajectories. Chain-of-Thought prompting increases reasoning depth~\cite{wei2022chain}, while self-consistency and repeated sampling increase breadth by generating and aggregating multiple paths~\cite{wang2023self,brown2024large}.

Scaling can also improve how these trajectories are organized and evaluated. Tree- and graph-based search methods structure the alternatives and allow intermediate states to be revisited~\cite{yao2023tree,besta2024graph}, while process supervision and verifier-guided reasoning help filter or redirect them~\cite{lightman2023let,wang2024multi}. Adaptive allocation further concentrates computation on inputs or intermediate states where it is most useful~\cite{snell2025scaling}. Together, these mechanisms turn inference into a process that can be explored, compared, and selected.

\revise{The common lesson is that longer reasoning is not itself the source of improvement. Scaling works best when extra computation produces diverse, task-relevant trajectories and a sufficiently reliable mechanism can evaluate or aggregate them; otherwise, greater depth and breadth may only amplify redundant or incorrect reasoning.}

\subsubsection{Reliability and Calibration}

Additional inference-time computation can also help determine whether an output should be accepted, revised, or withheld. This goal involves three related but distinct roles: checking the reliability of an output, selecting among candidate outputs, and assessing whether confidence reflects actual correctness.

Verification and self-correction improve \emph{output reliability} by checking evidence or revising a candidate. For example, CRITIC uses external tools to examine and refine an output, while Self-RAG combines retrieval with reflection to decide whether further evidence or revision is needed~\cite{gou2024critic,asai2023self}. Confidence-aware aggregation improves \emph{selection} by weighting candidates, stopping early, or allocating samples according to estimated confidence~\cite{huang2025efficient,taubenfeld2025confidence}. Calibration and uncertainty estimation instead ask whether confidence tracks correctness; agreement across internal representations or query variations can provide useful signals beyond a single prediction~\cite{xie2024calibrating,feng2025rethinking}.

\revise{These roles are related but not interchangeable. More samples or stronger consensus may improve selection without calibrating confidence, especially when reasoning paths share correlated errors or agree confidently on an incorrect answer. Reliable scaling therefore requires both effective checking or aggregation and uncertainty signals that remain informative under such shared failures.}

\subsubsection{Decision Quality in Control and Planning}

In control and planning, a correct answer is not enough: the model must construct an actionable path and revise it as new feedback arrives. Additional computation can therefore be used not only to reason about a task, but also to organize, compare, and update candidate decisions over a longer horizon.

Several mechanisms contribute to this process. ADaPT decomposes a task when execution fails, whereas Plan-and-Act separates high-level planning from low-level execution~\cite{prasad2024adapt,erdogan2025plan}. ARMAP evaluates candidate action trajectories according to their longer-term value~\cite{chen2025armap}. Other approaches acquire new information through environment interaction or construct world models that predict future states and support replanning~\cite{shen2025thinking,yu2025generating}.

\revise{Together, these mechanisms help determine what to solve next, which trajectory to follow, and when the plan should change. They are most useful when the task admits meaningful decomposition, trajectory-level feedback, or sufficiently faithful environment models. When these signals are inaccurate, however, additional search can optimize the wrong action path rather than improve decision quality.}

\begin{table*}[!t]
\centering
\caption{\revise{Main limitations of pure test-time scaling and their deployment consequences.}}
\label{tab:tts_limitations}
\footnotesize
\setlength{\tabcolsep}{3.2pt}
\renewcommand{\arraystretch}{1.12}
\resizebox{\textwidth}{!}{%
\begin{tabularx}{1.111\textwidth}{@{}
>{\raggedright\arraybackslash}p{0.189\textwidth}
>{\raggedright\arraybackslash}p{0.30\textwidth}
>{\raggedright\arraybackslash}p{0.30\textwidth}
>{\raggedright\arraybackslash}X@{}}
\toprule
\revise{\textbf{Limitation}} & \revise{\textbf{Why It Arises}} & \revise{\textbf{Deployment Consequence}} & \revise{\textbf{Representative Evidence}} \\
\midrule
\revise{High Computational Cost} &
\revise{Repeated generation, search, verification, and refinement multiply model calls and tokens.} &
\revise{Higher latency, memory use, and serving cost can outweigh modest accuracy gains.} &
\revise{Repeated/compound scaling~\cite{brown2024large,chen2024more}; tree search~\cite{chen2024tree}; compute-optimal allocation~\cite{snell2025scaling}.} \\
\midrule
\revise{No Persistent Improvement by Default} &
\revise{Pure scaling does not write episode-specific results into reusable model state.} &
\revise{When reuse across queries is required, the same computation may be repeated unless gains are explicitly retained or amortized offline.} &
\revise{Self-improvement limits~\cite{huang2025selfimprovement}; MiND~\cite{song2024mind}; SEAL~\cite{zweiger2025self}; offline amortization~\cite{lin2025sleep}.} \\
\midrule
\revise{Diminishing Returns and Instability} &
\revise{Candidate diversity saturates, while long reasoning, unreliable critique, or poorly coordinated agents can add noise.} &
\revise{Additional budget yields little gain or can reduce accuracy and increase variance.} &
\revise{Compound scaling~\cite{chen2024more}; CoT length~\cite{wu2026more}; self-critique~\cite{stechly2024self}; agent scaling~\cite{li2026benchmark}.} \\
\bottomrule
\end{tabularx}}
\end{table*}

\subsection{Limitations of Pure Scaling}

Despite its empirical benefits, pure test-time scaling is not universally effective. The mechanisms that improve an inference result also add latency and resource cost; their gains usually remain local to the current episode; and additional computation may eventually become redundant or unstable~\cite{snell2025scaling,huang2025selfimprovement,chen2024more}. Table~\ref{tab:tts_limitations} traces these limitations from their causes to their consequences for deployment.

\subsubsection{High Computational Cost}

The same mechanisms that create scaling gains also create their cost. Repeated sampling and aggregation require more model calls as additional candidates are generated~\cite{brown2024large,chen2024more}. Search and verification introduce further overhead because they repeatedly expand, score, and revise intermediate states. For example, tree-search procedures can be substantially slower than simpler baselines while providing only limited incremental gains in some settings~\cite{chen2024tree}.

This trade-off makes the marginal improvement per unit of computation more important than the raw budget alone. Compute-optimal and adaptive allocation can avoid assigning the same budget to every input~\cite{snell2025scaling}, but they cannot remove the underlying cost of generation and evaluation. Practical deployment must therefore consider latency, memory use, and serving cost jointly with accuracy.

\subsubsection{No Persistent Improvement by Default}

Pure scaling usually improves the current inference episode without changing reusable model state. It may help the model find a better reasoning path or candidate for the current input, but the same search or verification may need to be repeated for later queries because the gain does not automatically accumulate~\cite{huang2025selfimprovement,song2024mind}. This episode-local behavior is not problematic for every use case, but it becomes a limitation when information or capability should be reused over time.

Cross-query persistence therefore requires an additional retention mechanism. Test-time results can be written into parameters or memory, as illustrated by SEAL~\cite{zweiger2025self}, or transferred through a later offline learning pipeline~\cite{lin2025sleep}. \rtwo{The latter is the training-time amortization setting discussed in Section~\ref{sec:scale_for_learn}.} Both routes can preserve useful results, but pure scaling alone does not provide this persistence.

\subsubsection{Diminishing Returns and Instability}

Additional computation may first encounter \emph{saturation}. As candidate coverage or diversity plateaus, more calls increasingly produce redundant outputs and smaller marginal gains~\cite{chen2024more}. Longer reasoning can likewise become counterproductive when additional steps add noise rather than useful progress~\cite{wu2026more}.

Scaling can also become unstable rather than merely saturate. Unreliable self-critique may reinforce incorrect judgments, while poorly coordinated agent scaling can increase variance and propagate shared errors~\cite{stechly2024self,li2026benchmark}. Additional computation remains useful only while generation is informative, evaluation can distinguish useful trajectories, and allocation stops before marginal gains are exhausted.

Taken together, these limitations show that deployment value depends on more than the available inference budget. Extra computation is worthwhile when its reliable marginal benefit exceeds its latency and resource cost, and when its episode-local or persistent effect matches the intended use.

\section{The Intersection of Learning and Scaling}\label{sec:ttl_tts}
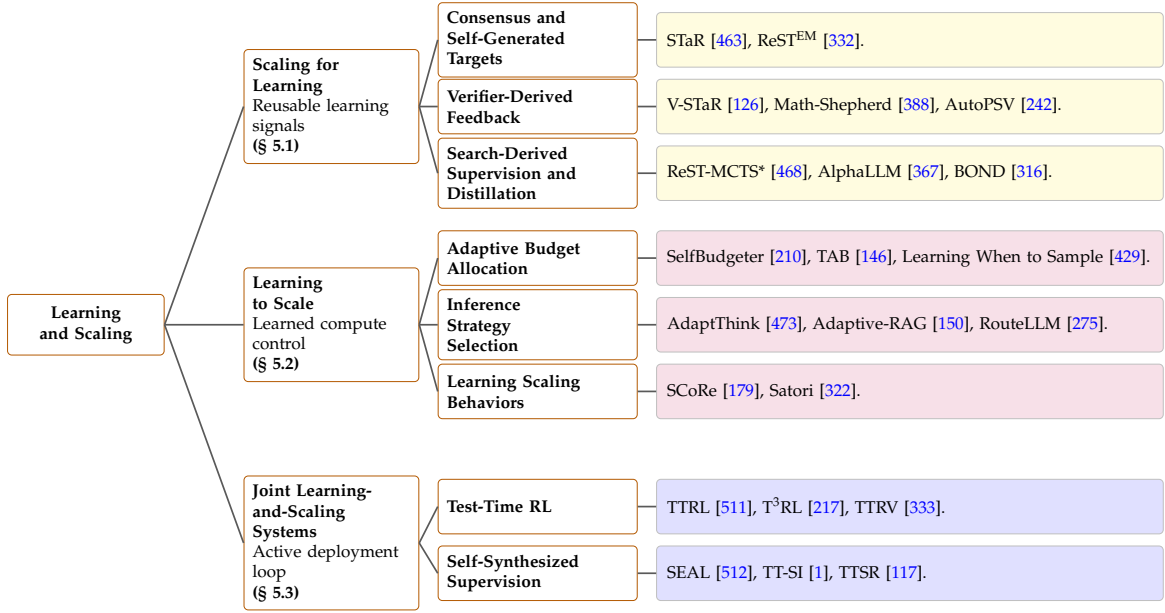
\begin{figure*}[!tbp]
\centering
\resizebox{0.9\textwidth}{!}{%
\begin{tikzpicture}[
x=1cm,
y=1cm,
line/.style={draw=black!65, line width=0.8pt},
root/.style={
    rectangle,
    rounded corners=2pt,
    draw=orange!70!black,
    fill=white,
    align=center,
    font=\bfseries\scriptsize,
    inner sep=4pt,
    text width=2.3cm,
    minimum height=1.0cm
},
group/.style={
    rectangle,
    rounded corners=2pt,
    draw=orange!70!black,
    fill=white,
    align=left,
    font=\bfseries\scriptsize,
    inner sep=4pt,
    text width=2.6cm,
    minimum height=1.15cm
},
sub/.style={
    rectangle,
    rounded corners=2pt,
    draw=orange!70!black,
    fill=white,
    align=left,
    font=\bfseries\scriptsize,
    inner sep=4pt,
    text width=3.0cm,
    minimum height=0.9cm
},
leaf1/.style={
    rectangle,
    rounded corners=2pt,
    draw=black!25,
    fill=yellow!15,
    align=left,
    font=\scriptsize,
    inner sep=5pt,
    text width=8.0cm,
    minimum height=0.9cm
},
leaf2/.style={
    rectangle,
    rounded corners=2pt,
    draw=black!25,
    fill=purple!12,
    align=left,
    font=\scriptsize,
    inner sep=5pt,
    text width=8.0cm,
    minimum height=0.9cm
},
leaf3/.style={
    rectangle,
    rounded corners=2pt,
    draw=black!25,
    fill=blue!12,
    align=left,
    font=\scriptsize,
    inner sep=5pt,
    text width=8.0cm,
    minimum height=0.9cm
}
]

\node[root] (root) at (0,0)
{\textbf{Learning}\\\textbf{and Scaling}};

\node[group, anchor=west] (g1) at (2.6,3.6)
{\textbf{Scaling for}\\\textbf{Learning}\\{\normalfont\scriptsize Reusable learning signals}\\(\S~\ref{sec:scale_for_learn})};

\node[group, anchor=west] (g2) at (2.6,0)
{\textbf{Learning}\\\textbf{to Scale}\\{\normalfont\scriptsize Learned compute control}\\(\S~\ref{sec:learn_to_scale})};

\node[group, anchor=west] (g3) at (2.6,-3.6)
{\textbf{Joint Learning-}\\\textbf{and-Scaling}\\\textbf{Systems}\\{\normalfont\scriptsize Active deployment loop}\\(\S~\ref{sec:joint-learning-and-scaling})};

\node[sub, anchor=west] (s11) at (5.8,4.7)
{\textbf{Consensus and}\\\textbf{Self-Generated}\\\textbf{Targets}};

\node[sub, anchor=west] (s12) at (5.8,3.6)
{\textbf{Verifier-Derived}\\\textbf{Feedback}};

\node[sub, anchor=west] (s13) at (5.8,2.5)
{\textbf{Search-Derived}\\\textbf{Supervision and}\\\textbf{Distillation}};

\node[sub, anchor=west] (s21) at (5.8,1.1)
{\textbf{Adaptive Budget}\\\textbf{Allocation}};

\node[sub, anchor=west] (s22) at (5.8,0)
{\textbf{Inference}\\\textbf{Strategy}\\\textbf{Selection}};

\node[sub, anchor=west] (s23) at (5.8,-1.1)
{\textbf{Learning Scaling}\\\textbf{Behaviors}};

\node[sub, anchor=west] (s31) at (5.8,-3.0)
{\textbf{Test-Time RL}};

\node[sub, anchor=west] (s32) at (5.8,-4.1)
{\textbf{Self-Synthesized}\\\textbf{Supervision}};

\node[leaf1, anchor=west] (l11) at (9.4,4.7)
{STaR~\citep{zelikman2022star}, ReST$^{\mathrm{EM}}$~\citep{singh2024beyond}.};

\node[leaf1, anchor=west] (l12) at (9.4,3.6)
{V-STaR~\citep{hosseiniv}, Math-Shepherd~\citep{wang2024mathshepherd}, AutoPSV~\citep{lu2024autopsv}.};

\node[leaf1, anchor=west] (l13) at (9.4,2.5)
{ReST-MCTS*~\citep{zhang2024rest}, AlphaLLM~\citep{tian2024toward}, BOND~\citep{sessa2025bond}.};

\node[leaf2, anchor=west] (l21) at (9.4,1.1)
{SelfBudgeter~\citep{li2025selfbudgeter}, TAB~\citep{jali2026not}, Learning When to Sample~\citep{xiong2026learning}.};

\node[leaf2, anchor=west] (l22) at (9.4,0)
{AdaptThink~\citep{zhang2025adaptthink}, Adaptive-RAG~\citep{jeong2024adaptive}, RouteLLM~\citep{ong2025routellm}.};

\node[leaf2, anchor=west] (l23) at (9.4,-1.1)
{SCoRe~\citep{kumar2025score}, Satori~\citep{shen2025satori}.};

\node[leaf3, anchor=west] (l31) at (9.4,-3.0)
{TTRL~\citep{zuo2025ttrl}, T$^3$RL~\citep{liao2026tool}, TTRV~\citep{singh2025ttrv}.};

\node[leaf3, anchor=west] (l32) at (9.4,-4.1)
{SEAL~\citep{zweiger2025self}, TT-SI~\citep{acikgoz2025self}, TTSR~\citep{he2026ttsr}.};

\draw[line] (root.east) -- (g1.west);
\draw[line] (root.east) -- (g2.west);
\draw[line] (root.east) -- (g3.west);

\draw[line] (g1.east) -- (s11.west);
\draw[line] (g1.east) -- (s12.west);
\draw[line] (g1.east) -- (s13.west);

\draw[line] (g2.east) -- (s21.west);
\draw[line] (g2.east) -- (s22.west);
\draw[line] (g2.east) -- (s23.west);

\draw[line] (g3.east) -- (s31.west);
\draw[line] (g3.east) -- (s32.west);

\draw[line] (s11.east) -- (l11.west);
\draw[line] (s12.east) -- (l12.west);
\draw[line] (s13.east) -- (l13.west);

\draw[line] (s21.east) -- (l21.west);
\draw[line] (s22.east) -- (l22.west);
\draw[line] (s23.east) -- (l23.west);

\draw[line] (s31.east) -- (l31.west);
\draw[line] (s32.east) -- (l32.west);

\end{tikzpicture}%
}
\caption{\rtwo{Taxonomy of the intersection between learning and scaling. Scaling for learning converts additional inference into reusable learning signals, learning to scale uses learned policies to control inference-time computation, and joint systems close the loop by updating the active system during deployment.}}
\label{fig:learning_scaling_intersection}
\end{figure*}

\rtwo{Learning and scaling interact through two directional relationships and one closed-loop composition. \textbf{Scaling for learning} uses additional inference-time computation to produce stronger learning signals, whereas \textbf{learning to scale} uses a learned controller to decide how much computation to allocate or which inference mechanism to invoke. \textbf{Joint learning-and-scaling systems} combine these directions by using scaling-derived feedback to update writable state and thereby influence subsequent inference within the same deployment process. We organize this section according to these relationships and the timing of the resulting improvement, as shown in Figure~\ref{fig:learning_scaling_intersection}.}

\subsection{Scaling for Learning}
\label{sec:scale_for_learn}
\label{sec:amortizable-benefits}

\rtwo{\emph{Scaling for learning} concerns how additional inference or inference-like computation generates signals for model learning, for example through generation, verification, or search. Such methods are typically performed during training or data construction, where their outputs are converted into reusable supervision or distilled into parametric capability. This corresponds to what we term \emph{training-time amortization} in Section~\ref{sec:definition}, and is therefore \emph{outside the scope of TTI under our definition}, which focuses on computation over test-time inputs. Nevertheless, this paradigm offers a useful perspective on converting inference computation into persistent model improvement, and may offer useful mechanisms and insights for future TTI research. We therefore briefly review this related line of work.}

\subsubsection{Consensus and Self-Generated Targets}
At the coarsest level, multiple generations can provide answer- or rationale-level targets when direct rationale or process supervision is unavailable.

\textit{Consensus targets} aggregate multiple reasoning paths into a pseudo-label, as in self-consistency~\cite{wang2023self}, which is itself an inference strategy. \textit{Rationale and sample bootstrapping} reuse selected generations for learning. STaR~\cite{zelikman2022star} retains rationales that lead to correct answers, whereas ReST$^{\mathrm{EM}}$~\cite{singh2024beyond} repeatedly generates samples, filters them with binary feedback, and fine-tunes on the accepted set. Related approaches reuse self-generated reasoning traces as supervision~\cite{li2025care} or expand prompts to construct additional fine-tuning examples~\cite{bsharat2025prompting}. Consensus provides a cheap but coarse signal, while iterative filtering produces reusable supervision at higher generation and update cost. Both can reinforce systematic errors when consensus or selection feedback is unreliable.

\subsubsection{Verifier-Derived Feedback}
Verification converts candidate generation into judgments that can filter, rank, or annotate self-generated outputs before they are reused for learning.

\textit{Answer-level verification} distinguishes correct from incorrect solutions, as in V-STaR~\cite{hosseiniv}. \textit{Process-level verification} derives denser step supervision from additional rollouts or intermediate assessments. Math-Shepherd~\cite{wang2024mathshepherd} labels steps through sampled continuations, OmegaPRM~\cite{luo2024omega} uses divide-and-conquer MCTS to locate early errors, and AutoPSV~\cite{lu2024autopsv} derives process labels from verifier-confidence changes. \textit{Preference and self-judgment feedback} instead ranks self-generated candidates for optimization~\cite{wang2024self}; Self-Rewarding Language Models~\cite{yuan2024selfrewarding} use the model itself as a judge in iterative DPO, and generation and verification can also be jointly learned~\cite{chen2026learning}. Finer feedback improves credit assignment but increases evaluation cost and exposure to verifier bias, which later updates may amplify.

\subsubsection{Search-Derived Supervision and Distillation}
Search and interaction produce trajectories containing intermediate decisions, action branches, and recovery paths. These structures are useful in long-horizon settings, where a final outcome alone provides weak credit assignment. The search-to-learning pattern predates LLM agents: Expert Iteration~\cite{anthony2017expertiteration} and AlphaGo Zero~\cite{silver2017alphagozero} generalize tree-search improvements into policy or value networks that guide later search. ReAct~\cite{yao2023react} illustrates the interleaved thought--action--observation loop to retrieve useful information for future reasoning and finetuning. These are, however, offline search-to-learning precedents which reuses search feedback rather than active TTL that updates state online.

\textit{Failure and recovery trajectories} turn unsuccessful exploration into contrastive or corrective supervision. ETO~\cite{song2024trial} pairs failed environment trajectories with successful ones for iterative policy optimization with DPO~\cite{rafailov2023direct}, A$^3$T~\cite{yang2024react} constructs contrastive ReAct-style data, and Agent-R~\cite{yuan2025agent} uses MCTS to recover correct trajectories from erroneous attempts. \textit{Search-derived policy and value supervision} reuses tree-search experience more broadly: ReST-MCTS*~\cite{zhang2024rest} collects reasoning traces and step values, AlphaLLM and AlphaLLM-CPL~\cite{tian2024toward,wang2024towards} distill behavior or preferences from MCTS, and rStar-Math~\cite{guan2025rstar} co-evolves a policy model and process preference model from verified rollouts. WebRL~\cite{qi2025webrl} extends this principle to web interaction through a self-evolving task curriculum and outcome-reward-guided RL. Trajectory supervision provides richer credit and broader exploration than answer-level targets, but depends on reliable evaluation and sufficiently diverse search; otherwise, the update can inherit the search process's errors and biases.

\rtwo{\textit{Search-gain distillation} compresses the benefit of expensive candidate selection or search into model parameters. BOND~\cite{sessa2025bond} and related work~\cite{yang2024faster,huang2025fast} distill gains from Best-of-$N$ sampling, iterative search, or explicit thought tokens, reducing future inference cost while preserving much of the improvement. Overall, scaling for learning can reduce annotation requirements and future inference costs by converting expensive inference into reusable supervision or parametric capability. In most existing methods, however, the improvement is realized only after offline learning; Section~\ref{sec:joint-learning-and-scaling} further considers the closed-loop setting in which scaling-derived feedback updates the active system during deployment.}

\subsection{Learning to Scale}\label{sec:learn_to_scale}

The reverse relationship is \emph{learning to scale}, where learning either controls when, where, and how additional computation is used or trains an inference behavior that can exploit it effectively. Adaptive computation has earlier precedents in learned recurrent halting and calibrated early exits~\cite{graves2016adaptive,schuster2022confident}; recent work extends this principle to reasoning tokens, samples, planning, retrieval, tools, and search. Here, a \emph{learned controller} denotes a predictor or policy trained before deployment or updated online, whereas fixed confidence rules and prompted difficulty estimates are adaptive heuristics. A fixed learned controller supports TTS deployment, whereas updating the controller during deployment constitutes TTL. Whereas Section~\ref{tts} surveys how inference can be scaled, this subsection focuses on how allocation, routing, and scaling behaviors are learned.

\subsubsection{Adaptive Budget Allocation}

A primary form of learning to scale is \emph{adaptive budget allocation}, which matches tokens, samples, or search effort to their expected utility rather than applying the same budget to every input. This avoids over-processing easy instances while retaining additional computation for difficult ones.

\textit{Upfront allocation} estimates the value or amount of compute before solving an input. Learning How Hard to Think~\cite{damani2025learning} learns the expected benefit of additional samples or a more expensive decoder, and SelfBudgeter~\cite{li2025selfbudgeter} learns to predict and follow a token budget. TALE~\cite{han2025token} provides both a training-free prompted estimator and a post-trained budget-aware variant. \textit{Sequential allocation} instead reacts to emerging evidence. TAB~\cite{jali2026not} learns turn-level budgets, while Strategic Scaling~\cite{zuo2026strategic} uses an online bandit to distribute samples across queries under a shared budget. \textit{Selective sampling} asks whether one trajectory is already sufficient: Learning When to Sample~\cite{xiong2026learning} trains such a gate from trajectory features. In contrast, CATTS~\cite{lee2026agentic} assigns extra agentic samples from vote uncertainty without learning the controller. Upfront policies have low control overhead but depend on initial calibration; sequential policies exploit intermediate evidence but incur monitoring or exploration costs.

\subsubsection{Learning When to Reason, Retrieve, or Search}

Beyond deciding how much computation to allocate, a controller may choose among qualitatively different inference mechanisms, such as direct answering, extended reasoning, planning, retrieval, tool use, or search. This turns test-time scaling into a routing problem rather than only a token- or sample-budgeting problem.

\textit{Internal-computation routing} determines how reasoning is organized. AdaptThink~\cite{zhang2025adaptthink} chooses between Thinking and NoThinking modes, Adaptive Parallel Reasoning~\cite{panlearning} learns when to branch into parallel computation, and Learning When to Plan~\cite{paglieri2025learning} teaches an agent when explicit planning is useful. \textit{External-mechanism routing} controls access to knowledge and tools. Adaptive-RAG~\cite{jeong2024adaptive} selects among no, single-step, and iterative retrieval, while Self-RAG~\cite{asai2023self} learns retrieval and reflection tokens. Toolformer~\cite{schick2023toolformer} learns when and how to invoke APIs, and RouteRAG~\cite{guo2026routerag} jointly routes reasoning and text- or graph-based retrieval. Internal routing trades reasoning depth or organization against computation, whereas external routing must additionally account for access cost and the reliability of retrieved evidence or tool outputs. Both require calibrated utility estimates and explicit cost objectives.

\textit{Model and cascade routing} instead chooses the inference engine itself. FrugalGPT~\cite{chen2024frugalgpt} learns cascades over heterogeneous model APIs, while RouteLLM~\cite{ong2025routellm} routes queries between stronger and weaker models from preference data. Upfront routing avoids unnecessary expensive calls but relies on query-level quality prediction; cascades can inspect earlier outputs before escalation but incur additional latency and cost.

\subsubsection{Learning Scaling Behaviors}

\rtwo{Learning can also prepare inference procedures that use additional computation effectively, rather than only deciding when to invoke them. SCoRe~\cite{kumar2025score} trains a model to perform reliable multi-turn self-correction, while Satori~\cite{shen2025satori} internalizes autoregressive search with reflection and exploration. These methods improve the scaling behavior itself, whereas the preceding families control its budget or selection. Here, learning prepares a fixed inference policy, while TTS is realized at deployment when that policy uses additional steps of correction, reflection, or search.}

\subsection{Joint Learning-and-Scaling Systems}
\label{sec:joint-learning-and-scaling}
\rtwo{\emph{Joint learning-and-scaling systems} use additional inference-time computation to construct feedback and write the resulting update into the active system's parameters, modules, or state. The updated state is then reused within the same deployment process. This distinguishes joint systems from pure scaling, which leaves no learned state, and from later offline training. We organize them by how the loop is closed: rollout-derived rewards for online reinforcement learning and self-generated adaptation data for parameter updates.}

\subsubsection{Test-Time Reinforcement Learning}
\label{sec:test-time-rl}

Test-time reinforcement learning is one of the clearest joint systems: repeated inference explores candidate solutions and converts their outcomes into an online reward for updating the active model on the same test stream. Existing work differs mainly in how this reward is constructed. \textit{1) Consensus-derived rewards} use agreement across sampled outputs. TTRL~\cite{zuo2025ttrl} applies majority voting to unlabeled reasoning data, while TTRV~\cite{singh2025ttrv} extends frequency-based rewards to vision--language models. \textit{2) Verified and stabilized rewards} seek to prevent frequent but incorrect outputs from reinforcing themselves. T$^3$RL~\cite{liao2026tool} adds tool verification, DARE~\cite{du2026distribution} and SCOPE~\cite{wang2025beyond} refine reward estimation using rollout distributions or step-wise confidence, and DDRL~\cite{yu2026ddrl} filters ambiguous samples and debiases advantage estimation.
\rtwo{\textit{3) Task-grounded rewards} become available when a problem provides a task-grounded objective or \rthree{executable} evaluator. TTT-Discover~\cite{yuksekgonul2026learning}, for example, updates the model from search experience on a single scientific or engineering problem while prioritizing high-reward candidates. \rthree{Alpha-RTL~\cite{zhou2026alphartl} closes the loop for RTL optimization by sampling design variants, obtaining executable EDA feedback, reusing high-reward candidates, and updating the per-design LLM policy online. TTT-Discover thus exemplifies search-derived task rewards, whereas Alpha-RTL/TTT-RTL represents execution-derived task rewards.}}

Overall, test-time RL is most reliable when the induced reward remains correlated with correctness. Consensus offers broad label-free coverage but is vulnerable to shared errors, whereas tool- or task-based rewards are stronger when executable feedback is available, at the cost of additional access and computation.

\subsubsection{Self-Synthesized Supervision}

Instead of compressing inference outcomes into a reward, this family scales inference-time generation to construct self-synthesized adaptation data and uses it to update the active model within deployment. The methods differ in what controls this data construction. \textit{Direct self-editing} lets the model specify both what to learn and how to update: SEAL~\cite{zweiger2025self} generates fine-tuning data, update directives, or optimization settings before applying persistent weight updates. \textit{Instance-conditioned synthesis} generates a small set of training examples relevant to the current query and uses them to adapt the model before producing the final answer. TT-SI~\cite{acikgoz2025self} applies this process to uncertain inputs by generating similar examples for lightweight fine-tuning. QueST~\cite{song2026quest} derives structurally related problem--solution pairs from the query for query-specific adaptation. MASS~\cite{kaya2026mass} further learns how to generate and weight a per-instance synthetic curriculum according to how much it improves post-update performance.
\textit{Reflection-guided curricula} instead use current failures or capability to decide what should be generated next. TTSR~\cite{he2026ttsr} diagnoses failed trajectories and synthesizes targeted variants, while TTCS~\cite{yang2026ttcs} co-evolves a question synthesizer and solver using self-consistency rewards. Direct self-editing jointly specifies the generated data and update configuration, targeted synthesis localizes supervision to the current problem, and reflective curricula adapt the supervision as the model changes. All three reduce dependence on external labels, but require reliable filtering because errors in synthetic data can be reinforced by the subsequent update.

\section{Applications}

\begin{table*}[!t]
  \centering
  \caption{Representative methods for test-time intelligence across application domains.}
  \label{tab:application_taxonomy}
  \footnotesize
  \setlength{\tabcolsep}{2pt}
  \renewcommand{\arraystretch}{1.08}
  \setlength{\emergencystretch}{3em}
  \resizebox{\textwidth}{!}{%
  \begin{tabularx}{1.06\textwidth}{@{}
  >{\raggedright\arraybackslash\hyphenpenalty=10000\exhyphenpenalty=10000}p{0.20\textwidth}
  >{\raggedright\arraybackslash\hyphenpenalty=10000\exhyphenpenalty=10000}p{0.25\textwidth}
  >{\raggedright\arraybackslash\hyphenpenalty=10000\exhyphenpenalty=10000}X@{}}
    \toprule
    {\textbf{Domain}} & {\textbf{Category}} & {\textbf{Representative Methods}} \\
    \midrule
    
    \multirow{4}{*}{\parbox{0.20\textwidth}{\raggedright\hyphenpenalty=10000\exhyphenpenalty=10000{Vision Perception}}}
    & {Classification} &
    {CoTTA \cite{wang2022continual}; EATA \cite{tan2025uncertainty}; SAR \cite{niu2025adapt}; FOA \cite{niu2024foa}} \\

    & {Segmentation} &
    {AuxAdapt \cite{zhang2022auxadapt}; TeSLA \cite{tomar2023tesla}; DIGA \cite{wang2023dynamically}; PromptCAL \cite{lei2025promptcal}; VPTTA \cite{chen2024each}; APCoTTA \cite{zhu2023uncertainty}} \\

    & {Detection \& Tracking} &
    {DARTH \cite{segu2023darth}; MonoTTA \cite{lin2024monotta}; DPO \cite{chen2024dpo}} \\

    & {Video Understanding} &
    {ViTTA~\cite{lin2023video}; MC-TTA \cite{xiong2024modality}; T3AL \cite{liberatori2024test}} \\
    
    \midrule
    
    \multirow{5}{*}{\parbox{0.20\textwidth}{\raggedright\hyphenpenalty=10000\exhyphenpenalty=10000{Generative Models}}}
    & {Image Restoration \& Synthesis} &
    {DIP \cite{ulyanov2018deep}; ZSSR \cite{shocher2018zero}; SRTTA~\cite{deng2023efficient}; TTT-MIM \cite{mansour2024ttt}} \\

    & {Video Generation} &
    {CustomTTT \cite{bi2025customttt}; SETA \cite{chen2023open}; TTOM \cite{qu2025ttom}} \\

    & {3D Generation \& Reconstruction} &
    {DreamFusion \cite{poole2023dreamfusion}; TTT3R \cite{chen2025ttt3r}; CloudFixer \cite{shim2024cloudfixer}} \\

    & {Inference-Time Scaling} &
    {Ma et al.\ \cite{ma2025inference}; SANA \cite{xie2025sana}; FK steering \cite{singhal2025general}} \\

    \midrule

    \multirow{5}{*}{\parbox{0.20\textwidth}{\raggedright\hyphenpenalty=10000\exhyphenpenalty=10000{Language \& Multimodal}}}
    & {Large Language Models} &
    {BoN~\cite{cobbe2021training}; Self-Consistency~\cite{wang2023self}; ToT~\cite{yao2023tree}; TLM \cite{Hu2025TLM}; Self-RAG \cite{asai2023self}} \\

    & {Vision-Language Models} &
    {TPT \cite{shu2022test}; DN \cite{zhou2023test}; RA-TTA \cite{lee2025ratta}; TTRV~\cite{singh2025ttrv}; VisualPRM~\cite{wang2025visualprm}} \\

    & {Spatial Reasoning} &
    {TangramSR \cite{zong2026tangramsr}; MindJourney \cite{yang2025mindjourney}; AVIC \cite{yu2026and}} \\

    & {Speech \& Audio} &
    {SUTA \cite{lin2022listen}; SGEM \cite{kim2023sgem}; E-BATS \cite{dong2025bats}} \\
    
    \midrule

    \multirow{5}{*}{\parbox{0.20\textwidth}{\raggedright\hyphenpenalty=10000\exhyphenpenalty=10000{Embodied AI}}}
    & {Policy Adaptation} &
    {PAD~\cite{hansen2021self}; AdaptFly~\cite{chen2026adaptfly}; T4P~\cite{park2024t4p}; BRIC~\cite{lim2026bric}; TT-VLA \cite{liu2026fly}; Centaur \cite{sima2025centaur}} \\

    & {Online Skill Improvement} &
    {Voyager~\cite{wang2024voyager}; LRLL~\cite{tziafas2024lifelong}; TAMP~\cite{mendez2023embodied}; Hong~et~al.~\cite{hong2026learning}} \\

    & {Planning-Time Scaling} &
    {RoboMonkey \cite{kwok2025robomonkey}; RoVer \cite{dai2025rover}; CoVer-VLA \cite{kwok2026scaling}; MG-Select~\cite{jang2025verifier}; TACO~\cite{yang2025steering}} \\

    \midrule

    \multirow{4}{*}{\parbox{0.20\textwidth}{\raggedright\hyphenpenalty=10000\exhyphenpenalty=10000{Agentic AI}}}
    & {Environment/System Adaptation} &
    {GTTA \cite{chen2025grounded}; MAS-on-the-Fly \cite{liu2026mas}} \\

    & {Active Self-Improvement} &
    {TT-SI \cite{acikgoz2025self}} \\

    \midrule

    \multirow{4}{*}{\parbox{0.20\textwidth}{\raggedright\hyphenpenalty=10000\exhyphenpenalty=10000{Healthcare}}}
& {Clinical Adaptation} &
{TTA-DAE~\cite{karani2021test}; Zhao et al.~\cite{zhao2026active}; CertainTTA~\cite{dong2025certaintta}} \\

& {Patient Adaptation} &
{Jang et al.~\cite{jang2025calibration}; Wang et al.~\cite{wang2025robust}; Bi-TTA~\cite{li2024bi}; Karpowicz et al.~\cite{karpowicz2025stabilizing}} \\

& {Privacy-Aware Personalization} &
{ATP~\cite{bao2023adaptive}; MSAFed~\cite{jin2025msafed}} \\

    \bottomrule
  \end{tabularx}%
  }
\end{table*}

\subsection{Vision Perception}

\rtwo{Vision perception is a mature deployment setting for TTI because sensor, weather, acquisition, and temporal shifts alter the observations available to a fixed model while labels are usually unavailable. }

\rtwo{\textbf{Image classification, segmentation, and detection} are the fundamental tasks addressed by the conventional TTL methods reviewed in Section~\ref{sec3}. To avoid repetition, we summarize only the representative methods in Table~\ref{tab:application_taxonomy}.}

\textbf{Video understanding.}
\rtwo{Video provides ordered test samples, allowing adaptation signals to be constructed across frames or temporal segments rather than treating each image independently. ViTTA~\cite{lin2023video} provides an early video-tailored route by aligning online spatio-temporal statistics with retained source statistics. ST2ST~\cite{fahim2024st2st} uses self-supervision for video action recognition, while MC-TTA~\cite{xiong2024modality} exploits modality collaboration or temporally synchronized prompt tuning. For open-set or zero-shot recognition and localization, T3AL~\cite{liberatori2024test} and Skeleton-Cache~\cite{zhu2025boosting} use temporal context or cached evidence. The application-specific issue is therefore how to exploit ordered and multimodal evidence without allowing errors to accumulate along the stream.}

\begin{figure}[t]
  \centering
  \includegraphics[width=0.78\linewidth]{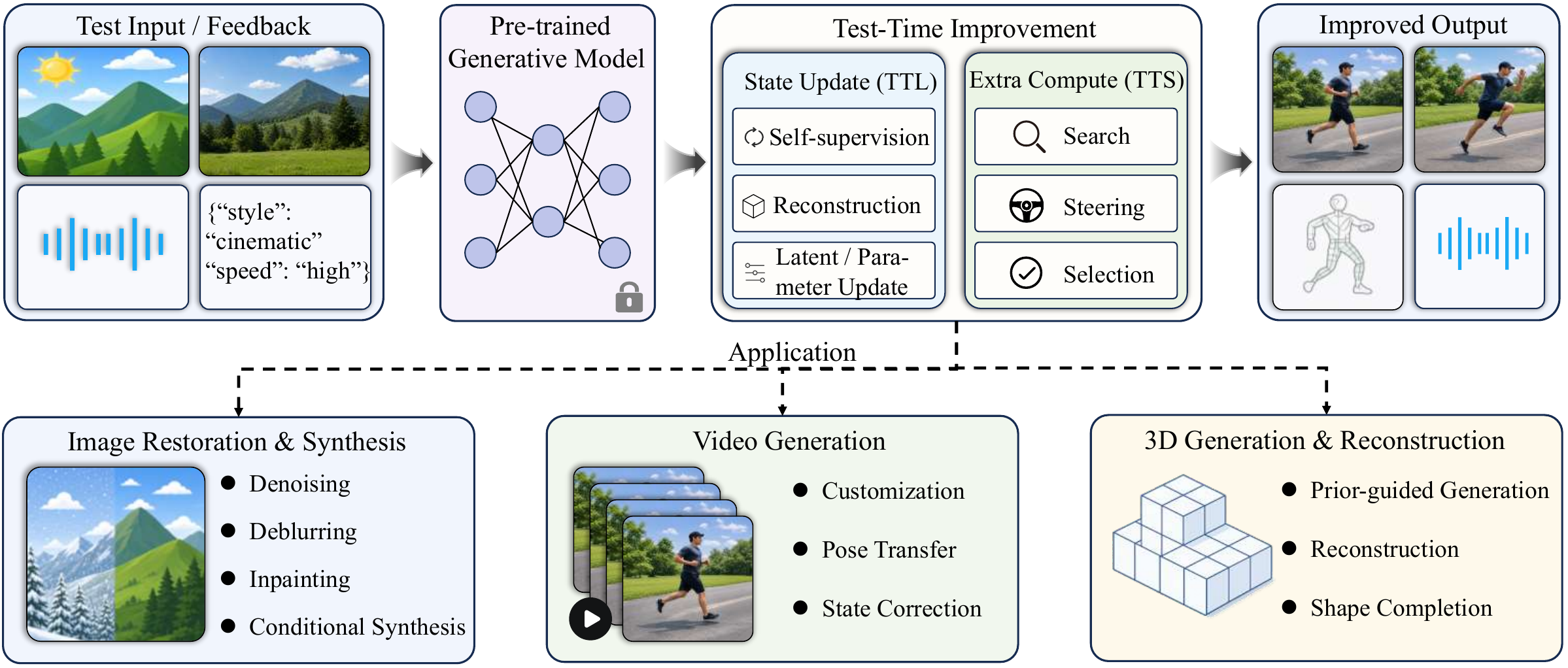}
  \caption{\rtwo{Test-time intelligence for generative models. Signals available at inference can refine image restoration and conditional synthesis, correct video states over time, adapt 3D reconstruction or completion, and guide search over stochastic generation trajectories.}}
  \label{fig:generative model}
\end{figure}

\subsection{Generative Models}
Generative models also benefit from test-time refinement and scaling, where extra computation is allocated at inference to iteratively improve quality, alignment, and temporal consistency without retraining the backbone.

\subsubsection{Image Restoration and Generation}

\rtwo{Earlier instance-internal methods optimize directly at test time: DIP~\cite{ulyanov2018deep} fits a randomly initialized generator so that its output matches the observed corrupted image, whereas ZSSR~\cite{shocher2018zero} trains an image-specific super-resolution model on self-supervised patch pairs derived from that image. Following them, recent methods can be grouped by where correction occurs. Model-side self-supervision or reconstruction adapts to the current degradation, as in CauSiam~\cite{cui2025continual}, SRTTA~\cite{deng2023efficient}, and TTT-MIM~\cite{mansour2024ttt}. Input- or degradation-side correction instead adjusts the input representation, uses a CLIP prior, performs collaborative model--data updates, or applies a diffusion corruption editor~\cite{shao2024adaptive,zhang2026collaborative,oh2024efficient}.}

\subsubsection{Video Generation}

\rtwo{In video generation, models must maintain visual quality, temporal coherence, motion consistency, and instance-specific control under changing test conditions. Representative TTI methods include CustomTTT for joint motion and appearance customization \cite{bi2025customttt}, SETA for open-world pose transfer through sequential adaptation \cite{chen2023open}, TTC for stabilizing long-video generation via reference-based calibration \cite{xiang2026pathwise}, and TTOM for improving compositional alignment through memory-guided optimization \cite{qu2025ttom}. These methods demonstrate the potential of TTI for controllable, stable, and adaptive video generation without full model retraining.}

\subsubsection{3D Generation and Reconstruction}

\rtwo{Test-time 3D methods follow two distinct routes. \textit{Prior-guided generation} leverages pretrained generative priors to optimize an instance-specific 3D representation. A representative example is DreamFusion~\cite{poole2023dreamfusion}, which distills knowledge from a frozen 2D diffusion model to optimize a text-conditioned NeRF. \textit{Observation-guided reconstruction} instead updates the model or 3D representation directly from incomplete test observations.  Existing methods span human-body methods that refine a test mesh~\cite{lumentut20223d,lumentut20233dhr}, shape-completion methods that adapt to observed structure~\cite{schopf20243d,jiang2025pointmac}, and point-cloud methods that correct inputs or representations~\cite{wei20253d,shim2024cloudfixer,wang2024backpropagation}. Moreover, TTT3R~\cite{chen2025ttt3r} explicitly formulates 3D reconstruction as test-time training. While prior-guided methods are constrained by the semantic alignment and geometric consistency of 2D generative priors, observation-guided methods depend heavily on the completeness and reliability of the available 3D measurements.}

\subsubsection{Inference-Time Scaling}\label{sec:diffusion_scaling}

\rtwo{Section~\ref{tts} reviewed the general mechanisms of inference-time compute scaling. Diffusion generation adds an application-specific search space: the stochastic denoising trajectory itself. Extra computation can compare initial noises, branch or resample intermediate states, and evaluate or reflect on partial generations while model parameters remain frozen. Representative methods use \rthree{initial noise-space optimization, where InitNO~\cite{guo2024initno} evaluates and optimizes initial noise using attention-derived scores and Ma et al.~\cite{ma2025inference} optimize noise candidates beyond simply increasing denoising steps}, classical or evolutionary search~\cite{zhang2025inference,he2025scaling}, particle-based steering with intermediate potentials (FK steering)~\cite{singhal2025general}, repeated sampling and selection(SANA)~\cite{xie2025sana}, reflection-based refinement (Reflect-DiT)~\cite{li2025reflect}, or $\epsilon$-greedy noise-trajectory search~\cite{ramesh2025test,dai2026guided}. This diffusion-specific structure makes compute actionable at several points along sampling, but its benefit depends on the quality of the evaluator or steering potential and comes with additional sampling cost.}
\rthree{The same scaling principle also extends beyond diffusion-based generation. ScalingAR~\cite{chen2026scalingar} applies TTS to next-token autoregressive image generation, using token-entropy confidence to adaptively prune trajectories and schedule guidance.}

\subsection{Language and Multimodal Models}

Many methods reviewed in previous sections already focus on conventional LLM/VLM reasoning (Sections~\ref{tts} and~\ref{sec:ttl_tts}) or CLIP-style VLM recognition (Section~\ref{sec:tta_adaptation}) tasks. To avoid repetition, we do not revisit them here, and only summarize some representative works in Table~\ref{tab:application_taxonomy}. Instead, we focus on more specialized scenarios, including spatial reasoning and speech/audio processing.

\begin{figure}[t]
    \centering
    \includegraphics[width=\linewidth]{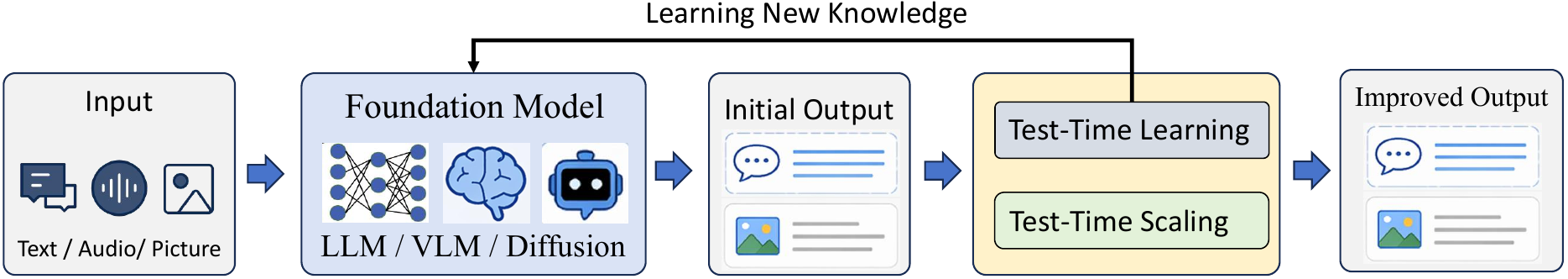}
    \caption{\rtwo{Test-time intelligence in language and multimodal foundation models. Given an input, a foundation model produces an initial output and then acquires task- and instance-specific knowledge at inference time through lightweight adaptation, such as prompt or external memory updates. The adapted model subsequently generates an improved output. The available feedback signals and adaptation objectives vary across modalities, including textual verification for language tasks, cross-modal alignment for vision--language tasks, and temporal consistency for speech tasks.}}
    \label{fig:tti_lmm}
\end{figure}

\subsubsection{Spatial Reasoning}

\rtwo{Spatial reasoning requires models to understand spatial relationships, geometry, distances, and viewpoints from visual observations. It is fundamental to understanding and interacting with the physical world, yet remains challenging due to the need for precise visual grounding and geometric reasoning. Moreover, reliable supervision is difficult to obtain, as spatial relations and geometric quantities are often not directly observable or verifiable without additional information.}

\rtwo{To address this, TangramSR~\cite{zong2026tangramsr} uses a training-free verifier--refiner loop to recursively improve geometric predictions through inference-time feedback. MindJourney~\cite{yang2025mindjourney} formulates spatial reasoning as a world-model-based test-time scaling problem, iteratively exploring imagined viewpoints and reasoning over multi-view evidence. AVIC~\cite{yu2026and} further treats visual imagination as an adaptive test-time resource, controlling when and how much imagined evidence to generate. Together, these works highlight spatial reasoning as a promising testbed for test-time verification, self-correction, and adaptive scaling, while explicit test-time updates remain under-explored.}

\subsubsection{Speech Recognition and Audio Processing} 
Test-time adaptation has been extensively studied for speech recognition to address acoustic domain shifts stemming from speaker variability, background noise, and cross-corpus mismatches. While general-purpose methods such as entropy minimization \cite{wang2021tent}, sample-efficient adaptation \cite{niu2022eata,niu2023sar}, and CoTTA \cite{wang2022continual} have been applied to speech tasks, their efficacy is often limited by the unique sequential structure and temporal dynamics of audio data.

To better capture these characteristics, speech-specific approaches have emerged. SUTA \cite{lin2022listen} pioneered single-utterance adaptation for ASR by extending entropy minimization to CNN feature encoders with temperature smoothing. Building upon this foundation, SGEM \cite{kim2023sgem} introduced sequence-level generalized entropy minimization with beam search-based logit acquisition and negative sampling, while CEA \cite{liu2024advancing} advanced adaptation for wild acoustic environments through refined loss design. Subsequent works have further addressed operational challenges: AWMC \cite{lee2023awmc} prevents mode collapse during continual adaptation, while Lin et al. \cite{lin2024continual} propose CSUTA and DSUTA---continual and dynamic variants of SUTA---to handle persistent noise through fast-slow model architectures.

Despite these advances, existing speech TTA methods predominantly rely on backpropagation through the feature encoder, incurring substantial memory overhead that limits scalability to large foundation models. E-BATS \cite{dong2025bats} addresses this limitation by proposing a backpropagation-free framework that optimizes lightweight prompt vectors while keeping the backbone parameters frozen, achieving comparable adaptation efficacy with significantly reduced memory footprint.

\subsection{Embodied AI and Robotics}

\begin{figure}[t]
    \centering
    \includegraphics[width=0.8\linewidth]{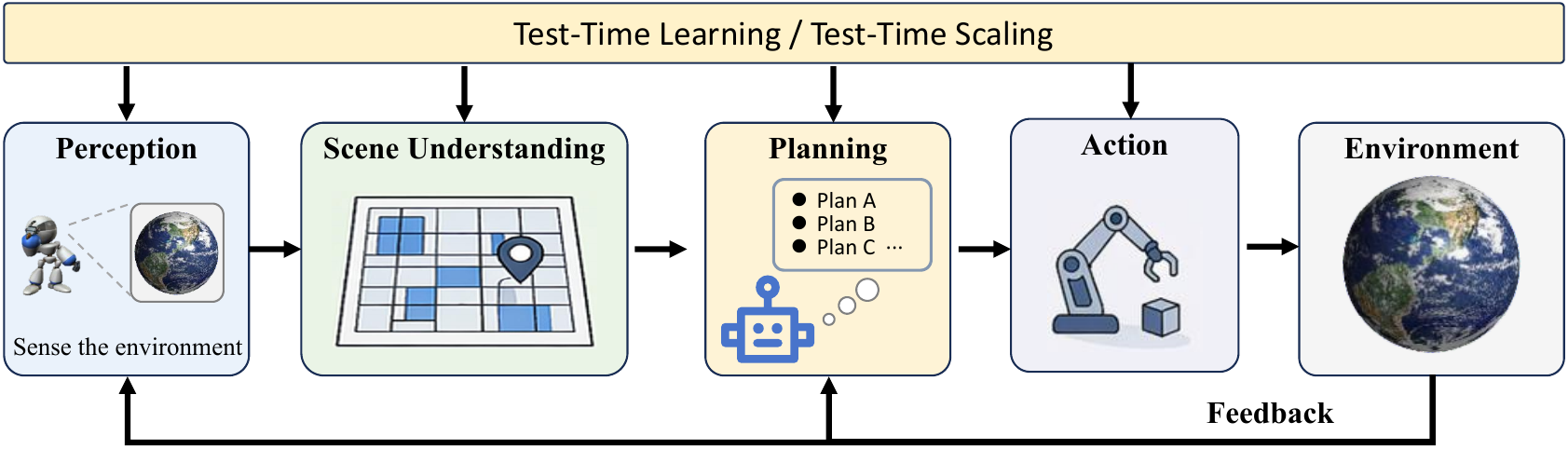}
    \caption{\rtwo{Test-time intelligence in embodied AI and robotics. Learning and scaling can intervene across the perception--planning--action loop through deployment feedback, persistent state updates, and additional computation over grounded candidate actions or plans.}}
    \label{fig:tti_robot}
\end{figure}

\rtwo{Embodied TTI operates inside a perception--planning--action loop: deployment feedback is delayed, action-dependent, and often safety-critical. We organize embodied TTI by three complementary mechanisms: adapting the current policy, accumulating reusable skills from interaction, and scaling computation for the current decision. Figure~\ref{fig:tti_robot} illustrates how learning and scaling intervene at different points in this loop.}

\subsubsection{Policy Adaptation to Unseen Conditions}

\rtwo{Policy adaptation to unseen conditions assumes that the deployed system already contains the required capability, but a change in observations, environments, or dynamics makes that capability unreliable. These methods differ primarily in which part of the decision process they recalibrate.}

\rtwo{\textit{1) Perception and representation adaptation} corrects upstream information while leaving the downstream policy largely intact. AdaptFly~\cite{chen2026adaptfly} handles weather, lighting, and viewpoint drift by retrieving or optimizing lightweight prompts for a frozen segmentation model. T4P~\cite{park2024t4p} instead uses a masked-autoencoder objective to update deeper trajectory representations and an actor-specific token memory to capture motion characteristics under driving distribution shifts. This route is suitable when performance degradation mainly arises from shifted observations or perception outputs, while the underlying policy remains capable; \textit{2) Planner and controller adaptation} localizes the update to a specific decision layer. BRIC~\cite{lim2026bric} adapts a physics controller to noisy diffusion-generated motion plans while regularizing against forgetting previously acquired skills. Centaur~\cite{sima2025centaur} updates an autonomous-driving planner using prior test-time observations to minimize uncertainty measured by Cluster Entropy. Such localized updates preserve more of the original system, although their benefit is limited by the adapted component; \textit{3) Direct policy adaptation} changes the action-producing policy itself. PAD~\cite{hansen2021self} continues a jointly trained inverse-dynamics objective during deployment, allowing the policy to adapt without test rewards or prior knowledge of the new environment. TT-VLA~\cite{liu2026fly} instead exploits richer feedback by converting stepwise task progress into dense rewards for test-time reinforcement learning while preserving the pretrained policy prior. Direct policy updates can correct behavior more substantially, but depend on reliable self-supervision or environment feedback to prevent harmful drift. Thus, the three families intervene at different levels but share the assumption that an existing capability must be made reliable under unseen deployment conditions.}

\subsubsection{Online Skill Improvement}
\rtwo{Online skill improvement uses deployment experience not only to recover an existing capability, but also to form reusable behaviors or improve how later tasks are solved. \textit{Skill-library expansion} stores successful experience as reusable skills. Voyager~\cite{wang2024voyager} improves action programs from environment feedback and execution errors, then retains successful programs for increasingly complex tasks. LRLL~\cite{tziafas2024lifelong} similarly combines self-guided exploration, experience memory, and skill abstraction to expand a composable robot skill library. These methods make improvement explicit through an expanding set of reusable skills. \textit{Experience-driven skill refinement} instead updates the models or policies that generate future behavior. Embodied Lifelong Learning for TAMP~\cite{mendez2023embodied} transfers planning experience across related tasks through shared and task-specific generative models. Hong~et~al.~\cite{hong2026learning} use failed trials as feedback to update both the reflection model and action policy, thereby improving subsequent behavior. Policy adaptation restores an existing capability under changed conditions, whereas online skill improvement uses deployment experience to expand or refine capabilities for later tasks.}

\subsubsection{Planning-Time Scaling}

\rtwo{Unlike policy adaptation or skill improvement, planning-time scaling does not require persistent updates to the agent, but improves the current decision by allocating additional deployment-time compute to search, refine, verify, or select grounded candidate actions and plans. RoboMonkey~\cite{kwok2025robomonkey} samples and verifies candidate actions; RoVer~\cite{dai2025rover} couples a robot process reward with candidate expansion; and CoVer-VLA~\cite{kwok2026scaling} extends scaling to hierarchical verification by jointly diversifying instructions and actions before selecting aligned behavior. Selection can also rely on internal confidence or a lightweight verifier, as in MG-Select~\cite{jang2025verifier} and TACO~\cite{yang2025steering}. Reflective Test-Time Planning further connects scaling with learning by evaluating candidates before execution and learning from post-execution reflection~\cite{hong2026learning}. Together, these methods characterize embodied planning-time scaling as additional deployment-time search, verification, and selection over grounded behavior.}

\subsection{Agentic AI}

\rtwo{Agentic systems provide a natural setting for feedback-driven TTI, as interactions with environments, tools, and other agents continuously produce feedback that can guide subsequent adaptation and improvement. Unlike ordinary execution of fixed agentic workflows, we focus on methods where such deployment-time feedback actively changes the agent or system to improve its behavior.}

\rtwo{GTTA~\cite{chen2025grounded} leverages environment-specific feedback for test-time adaptation. It learns lightweight adaptation vectors from test-time observations to align the agent with environment-specific syntax, and further explores state transitions to ground environment dynamics for subsequent decision making. MAS-on-the-Fly~\cite{liu2026mas} extends adaptation to multi-agent systems, where accumulated collaboration experience guides system instantiation and a dedicated watcher monitors runtime behavior to provide real-time interventions. Beyond environment-driven adaptation, TT-SI~\cite{acikgoz2025self} exploits self-assessed uncertainty as feedback: it identifies challenging test cases, generates targeted examples, and performs test-time updates for self-improvement. Together, these works illustrate how diverse feedback arising during agentic interaction---from environment observations and transitions to execution behavior and self-assessed uncertainty---can drive test-time adaptation and self-improvement.}

\subsection{Healthcare and Personalized AI}

\rtwo{Healthcare models may face variation shared across a clinical environment as well as variation specific to an individual patient. The former can arise from hospitals, scanners, or acquisition protocols, whereas the latter reflects physiological and behavioral differences across patients. Environment-level evidence may support updates shared across related cases, whereas updates that are patient-specific or retained over time require stronger evidence, privacy protection, and reliability controls.}

\rtwo{\textbf{Adapting to the clinical environment.} Changes in scanners, protocols, and image quality can degrade a model even when its clinical task remains unchanged. When the change can be localized to the current case, TTA-DAE~\cite{karani2021test} adapts image normalization under an anatomical prior. When the same shift persists across cases, active continual adaptation instead updates model state along a medical classification stream~\cite{zhao2026active}. In either setting, unreliable observations can misguide adaptation. CertainTTA~\cite{dong2025certaintta} therefore uses uncertainty to restrict unreliable segmentation updates. Case-specific correction limits the influence of an error, whereas a shared update is useful only when related cases provide consistent evidence of an environmental shift. Even after this environment is addressed, however, patients may differ under the same acquisition conditions.}

\rtwo{\textbf{Adapting to the individual patient.} Neural and physiological signals vary across patients, sessions, and time, making repeated supervised calibration impractical. Observations collected during deployment can instead establish and refine patient-specific state. Calibration-free EEG-based drowsiness detection reduces the need for newly labeled session data~\cite{jang2025calibration}, while methods for sEEG speech decoding~\cite{wang2025robust} and remote physiological measurement~\cite{li2024bi} track changes in neural and physiological signals. Over longer periods, latent-dynamics alignment compensates for neural drift without repeated labels~\cite{karpowicz2025stabilizing}. Although these methods reduce labeled calibration, they still require repeated observations from the current patient to distinguish stable individual characteristics from temporary measurement noise.}

\rtwo{\textbf{Privacy-aware personalization.} Patient-specific adaptation relies on sensitive deployment data, making where test-time updates occur and what information is exchanged central deployment constraints. TTPFL~\cite{bao2023adaptive} performs unsupervised test-time personalization locally at federated clients, while MSAFed~\cite{jin2025msafed} combines multi-center federated learning with test-time adaptation at unseen medical clients. These approaches reduce the need to centralize raw client or patient data while supporting adaptation to client- or site-specific shifts. However, federated execution alone does not provide a formal privacy guarantee for information encoded in the updated model state.}

\section{Open Challenges and Future Directions}\label{sec:open_challengs_future_directions}

\subsection{Open Challenges}

The open challenges of test-time intelligence largely center around one fundamental question: \textbf{How can test-time learning and test-time scaling be deployed in the real world at scale?} In relation to this question, several important dimensions still remain underexplored.

\vspace{6pt}\noindent\textbf{Long-Horizon Stability in the Wild} 
The stability issue primarily arises in learning-based methods. Their strong performance largely stems from explicit learning that modifies the model’s internal states. However, in open-world deployment, the environment can be arbitrary, and no assumptions should be made about the incoming test data stream. Under such conditions, the inherently noisy unsupervised signals used for TTL become more unreliable and uncertain, which places the model at a substantial risk of collapse~\cite{niu2023sar}. This issue becomes even more pronounced in long-horizon learning. To mitigate such collapse, prior works have explored various strategies, including unreliable-sample gradient filtering~\cite{niu2022eata,lee2024deyo,hu2025beyond}, sharpness-aware optimization~\cite{niu2023sar}, feature-diversity regularization~\cite{you2025test,niu2025adapt}, online risk monitoring for model recovery~\cite{schirmer2025monitoring}, and so on. Relative to the earliest TTL mechanisms~\cite{wang2021tent}, these advances have significantly improved stability. Nevertheless, there is still a room between current methods and the requirements of large-scale real-world deployment, particularly in complex and novel scenarios. Addressing this stability issue is important and in urgent demand since model collapse can lead to unacceptable failures once it occurs.

\vspace{6pt}\noindent\textbf{Effective Learning and Evaluation Objectives}
TTI fundamentally relies on an effective objective: learning-based methods require reliable unsupervised signals for model updates, while scaling-based methods require accurate criteria to assess, verify, or rank candidate outputs. The quality of these objectives largely determines whether a TTI method is effective in practice. However, despite the growing body of TTI methods~\cite{liang2025comprehensive}, most existing objectives are still designed for specific tasks, models, or modalities, and thus may generalize poorly beyond their original settings, \eg, applying Tent~\cite{wang2021tent} to segmentation~\cite{lee2025instance}. As new tasks, models, and deployment scenarios continue to emerge, developing test-time objectives that are not only effective, but also easy-to-use, computationally efficient, and broadly applicable, is still a central long-standing open challenge in the TTI literature. Examples include anti-collapse objectives for stable online test-time learning, versatile objectives that support various tasks, and low-cost evaluation mechanisms for reliably estimating output quality during test-time scaling.

\vspace{6pt}\noindent\textbf{Efficiency: Faster Convergence and Forward-Only}
As TTI introduces extra compute at inference, it is inherently less efficient than conventional ``standard inference". However, at test time, we hope the inference should be as efficient as possible, especially for latency-sensitive scenarios and resource-constrained edge devices. Prior works have made progress in this direction, for example in TTL literature: 1) filtering unreliable samples to reduce adaptation cost~\cite{niu2022eata,lee2024deyo}, 2) replacing standard optimization with learned optimizer for faster convergence, such as MGTTA~\cite{deng2025learning}, thereby achieving better performance under a limited computational budget, or 3) adopting forward-optimization strategies such as FOA~\cite{niu2024foa} and E-BATS~\cite{dong2025bats} to reduce memory and improve deployability. Nevertheless, these methods are still far from the ideal goal of ``achieving efficiency close to a single forward pass", which would greatly broaden the practicality of TTI in applications such as autonomous driving and online systems. To this end, important future trends include developing more efficient zeroth-order optimization methods that consider the noisy nature of unsupervised learning, better objectives tailored to such optimization, effective sampling or verification strategies for LLM-based systems, applications to quantized models, and so on.

\vspace{6pt}\noindent\textbf{Practical and Diverse Benchmarks} 
Much of the current literature on TTI still focuses on relatively narrow settings, such as synthetic corruptions~\cite{hendrycks2019benchmarking} for TTL, or benchmark-specific reasoning tasks for TTS. While useful for controlled analysis, these settings only partially reflect real deployment, where shifts can be mixed, evolving, multimodal, personalized, and coupled with temporal or interaction effects. More diverse and realistic benchmarks that capture these complexities are still needed. Moreover, current evaluation protocols often fail to jointly assess the key dimensions of TTI, including accuracy, latency, memory, compute overhead, calibration, and stability. Standardized evaluation under matched test-time budgets (\eg, \cite{alfarra2024evaluation}) and comparable feedback access is also still lacking. \revision{Although building such benchmarks remains difficult, they are essential for advancing deployable TTI.}

\vspace{6pt}\noindent\textbf{Auto Hyperparameter Setup}  
Many TTI methods rely on hyperparameters such as thresholds, learning rates, and memory sizes that are manually selected or tuned on validation data. In real deployment, however, a representative validation set from the target environment is often unavailable. Meanwhile, optimal setups may vary across models, domains, users, tasks, and time, making these methods fragile and hard to transfer.
Auto hyperparameter setup is therefore crucial for making TTI practical. Ideally, a deployable system should be able to calibrate itself online, including its hyperparameters, based on available signals such as proxy performance estimates~\cite{kim2023reliable} (an important tool), uncertainty, and resource constraints.
More broadly, this exposes a deeper limitation of current TTI: many methods still depend heavily on offline tuning, which weakens their claim of being truly adaptive and deployment-ready. Despite its importance, automatic hyperparameter setup remains long-neglected yet challenging.

\vspace{6pt}\noindent\textbf{Theoretical Foundation}
\revise{The current TTI literature remains largely empirical. As discussed in Sections~\ref{sec:tta_theory} and~\ref{sec:tts_theory}, existing theory provides conditional results on proxy--task alignment, mechanism-specific target risk, repeated-adaptation stability, non-stationary recovery, sampling error, verification, and budget allocation~\cite{sun2020test,zhang2023adanpc,hoang2024persistent,zhou2026ttalearnability,zhou2025rpc,setlur2025scaling,lin2026planbudget}. However, these results analyze isolated mechanisms under specific assumptions and do not yet provide a unified account of how feedback, writable state, and additional computation jointly determine deployment-time improvement. For TTL, theory should characterize when deployment signals are sufficiently informative, when proxy objectives align with task risk, and when updates to parameters, statistics, or memory improve performance rather than accumulate errors. For TTS, it should explain when additional sampling creates effective diversity, when verifiers reliably identify better candidates, and how finite compute should be allocated across generation, evaluation, and reasoning stages. For hybrid and long-term TTI, a broader framework should determine when to update state rather than spend additional computation, whether short-term gains persist, and how stability, regret, and risk evolve across deployment. Such a foundation would clarify when TTI can and cannot be expected to improve.}

\subsection{Future Directions}

\revise{Table~\ref{tab:future-directions-roadmap} summarizes the expected impact, current gaps, and qualitative near-term feasibility of the following research directions.}

\begin{table*}[!t]
\centering
\caption{\revise{Summary of future directions in TTI, including their expected impact, current gaps, and qualitative near-term feasibility.}}
\label{tab:future-directions-roadmap}
\renewcommand{\arraystretch}{1.2}
\small
\resizebox{0.98\textwidth}{!}{%
\begin{tabular*}{26.2cm}{@{\extracolsep{\fill}}>{\raggedright\arraybackslash}p{4.2cm} >{\raggedright\arraybackslash}p{8.8cm} >{\raggedright\arraybackslash}p{9.4cm} >{\centering\arraybackslash}p{2.0cm}}
\toprule
\revise{\textbf{Future Direction}} & \revise{\textbf{Expected Impact}} & \revise{\textbf{Current Gap}} & \revise{\textbf{Feasibility}} \\
\midrule
\revise{Test-Time Training for Long-Context Memory}
& \revise{\textbullet\ Enable TTI systems to operate over long-horizon interactions without unbounded context growth.}\par\revise{\textbullet\ Retain relevant information across long-horizon interactions and multimodal streams.}
& \revise{\textbullet\ Existing learnable-memory architectures have not demonstrated reliable long-term operation under bounded state and computation.}\par\revise{\textbullet\ Trade-offs among retention, interference, memory capacity, update stability, and inference efficiency remain insufficiently understood.}
& \revise{High} \\
\midrule
\revise{Safety under Adversarial Attack}
& \revise{\textbullet\ Make TTI systems deployable in open and adversarial real-world environments.}\par\revise{\textbullet\ Protect adaptation and memory from persistent corruption while preserving reliable inference-time scaling under attack.}
& \revise{\textbullet\ TTI lacks a unified threat model and evaluation protocol spanning adaptation- and scaling-based attacks.}\par\revise{\textbullet\ Mechanisms for detecting unreliable feedback, constraining harmful state changes, and recovering after compromise remain incomplete.}
& \revise{High} \\
\midrule
\revise{Black-Box Adaptation}
& \revise{\textbullet\ Extend TTI from internally accessible models to widely deployed closed model services.}\par\revise{\textbullet\ Enable users to specialize general-purpose APIs online without access to parameters, gradients, or source-domain statistics.}
& \revise{\textbullet\ Existing approaches often rely on supervised offline data, tunable prompts, source statistics, or other forms of internal access.}\par\revise{\textbullet\ Evidence for online unsupervised adaptation remains largely limited to classification-based vision APIs.}
& \revise{High} \\
\midrule
\revise{Test-Time Intelligent Diffusion Models}
& \revise{\textbullet\ Broaden TTI to iterative generative processes by enabling intervention throughout generation.}\par\revise{\textbullet\ Improve output quality per unit compute through reliable trajectory feedback and adaptive computation allocation.}
& \revise{\textbullet\ General test-time mechanisms that transfer across diffusion applications remain at an early stage.}\par\revise{\textbullet\ Intermediate-state feedback remains noisy or misaligned, while computation allocation across trajectories and sampling stages remains underdeveloped.}
& \revise{High} \\
\midrule
\revise{Human Interactive TTI}
& \revise{\textbullet\ Make TTI systems responsive to evolving user needs while preserving human control over adaptation.}\par\revise{\textbullet\ Enable personalized and reliable behavior from lightweight corrections, preferences, and instructions.}
& \revise{\textbullet\ TTI lacks a general interaction framework that balances user control with autonomous adaptation over time.}\par\revise{\textbullet\ Sparse feedback remains difficult to incorporate efficiently, while low interaction cost and stable improvement have not been jointly achieved.}
& \revise{High} \\
\midrule
\revise{TTI for Scientific Discovery}
& \revise{\textbullet\ Establish TTI as a domain-specific optimization framework for complex scientific and engineering problems.}\par\revise{\textbullet\ Enable continuous search and refinement using problem-specific objectives and feedback from simulators, experiments, or scientific constraints.}
& \revise{\textbullet\ Existing evidence remains concentrated on LLM-based discovery, limiting support for a general scientific optimization framework.}\par\revise{\textbullet\ Reliable integration of simulators, experimental measurements, scientific constraints, and non-LLM models has not been broadly demonstrated.}
& \revise{Medium} \\
\midrule
\revise{TTI for Deep Research}
& \revise{\textbullet\ Extend TTI from problem-level solution optimization to workflow-level improvement across the research loop.}\par\revise{\textbullet\ Enable auditable AI assistance for reproduction, debugging, experiment management, idea generation, and model discovery.}
& \revise{\textbullet\ Errors and invalid decisions can propagate across interdependent workflow stages, making end-to-end reliability difficult to establish.}\par\revise{\textbullet\ Evidence remains limited to early prototypes, while validity control and responsible human oversight remain underdeveloped.}
& \revise{Medium} \\
\bottomrule
\end{tabular*}%
}
\end{table*}

\vspace{6pt}\noindent\textbf{Test-Time Training (TTT) for Long-Context Memory} 
Sun \etal~\cite{sun2024learning} first introduced the TTT-Layer, which views the hidden state as a test-time learnable layer to memorize long contexts. This idea has since inspired a series of follow-up works that improve the training objective, enhance memory capacity and reliability, enable plug-and-play memory transfer, and extend to video modality \cite{tandon2025end,hwang2026reinforced,feng2026place,swamynathan2026sr,wang2026allmem,li2026latent,liu2026spatial}. MGTTA~\cite{deng2025learning} also leverages a similar idea to memorize the historical test-time learning gradients, and based on it to refine current gradients for faster adaptation. Looking forward, since long-context modeling is increasingly important while standard attention-based Transformers scale poorly with sequence length, extending this learnable hidden-state paradigm toward scalable long-term memory architectures is a highly promising direction.

\vspace{6pt}\noindent\textbf{Safety under Adversarial Attack}
\revise{TTI can introduce security risks beyond standard inference because test inputs may influence not only the current prediction but also subsequent model behavior. In adaptation-based methods, malicious samples can corrupt test-time feedback and thereby poison updated parameters, statistics, or memory, leading to error accumulation over time~\cite{park2024medbn,wu2023uncovering}. In scaling-based methods, attackers may instead constrain candidate diversity or manipulate the evaluators used for selection, causing additional computation to amplify unsafe or incorrectly scored outputs~\cite{nahin2026less,raina2024llmjudge}. An important future direction is therefore to develop a systematic framework for evaluating and defending TTI systems under adversarial attack. On the evaluation side, standardized protocols are needed to specify attacker access, attack budget, and attack timing, and to measure both immediate errors and persistent degradation after an attack ends. On the defense side, future TTI systems should develop robust mechanisms that identify unreliable feedback, limit harmful state changes, and recover when test-time adaptation or selection is compromised.}

\vspace{6pt}\noindent\textbf{Black-Box Adaptation}
In practice, many models are deployed as a service through closed APIs, such as GPT and Gemini, where parameters and gradients are inaccessible due to privacy, security, or commercial reasons. This makes black-box test-time adaptation both challenging and highly practical, as it could help users better adapt general-purpose APIs to their own tasks during deployment. Existing exploration: Some LLM/VLM-based studies~\cite{meng2025black,sun2024bbox} mainly focus on supervised offline adaptation. Forward-optimization methods~\cite{niu2024foa,dong2025bats} often rely on prompt tuning or access to source domain statistics, which are actually gray-box and not fully black-box. BETA~\cite{zhang2026adapting} takes an initial step toward online unsupervised black-box adaptation, but its scope still focuses on classification-based vision APIs. Overall, black-box adaptation remains largely underexplored and offers a promising direction for future research.

\vspace{6pt}\noindent\textbf{Test-Time Intelligent Diffusion Models}  
\revise{As discussed in Section~\ref{sec:diffusion_scaling}, diffusion models offer a distinctive setting for TTI because stochastic, iterative sampling exposes intermediate states and alternative trajectories, enabling intervention throughout generation~\cite{singhal2025general,li2025reflect}. This structure opens two complementary directions: improving feedback quality~\cite{ma2025inference,kim2026lookahead} and allocating computation adaptively~\cite{ramesh2025test,chun2025dynamic}. Reliable feedback should assess not only final outputs but also whether intermediate states are promising, while remaining robust to noisy or misaligned signals. Adaptive allocation should then use this feedback to focus extra exploration or refinement on the trajectories and stages most likely to improve the output. These advances could enable more general, efficient, and robust test-time mechanisms for diffusion models while keeping the additional inference cost manageable. Overall, this area remains at an early stage and offers broad potential for future research.}

\vspace{6pt}\noindent\textbf{Human Interactive TTI}
Another promising direction is human-interactive TTI, where models improve at test time not only from self-generated signals, but also from lightweight human feedback such as corrections, preferences, or instructions. This can make TTI more personalized, controllable, and reliable, especially for open-ended or user-specific tasks. In this case, the open questions include how to efficiently incorporate sparse human feedback (such as binary rewards~\cite{lee2025test,hubotter2026test}), how to balance user control with autonomous adaptation, and how to maintain low interaction cost and stable improvement over time.

\vspace{6pt}\noindent\textbf{TTI for Scientific Discovery}
\revise{Many scientific problems, such as drug discovery~\cite{askr2023deep}, material design~\cite{choudhary2022recent}, and combinatorial optimization~\cite{bengio2021machine}, have problem-specific objectives and can provide test-time feedback through simulators, experimental measurements, or scientific constraints. This makes them well suited to TTI.} TTI can help models explore large solution spaces, incorporate domain knowledge, use external tools, and adapt inference to both the problem structure and the evolving environment. A promising direction is therefore TTI + X, where TTI is combined with simulators, retrieval systems, optimization algorithms, or human expertise to build domain-specialized discovery systems. Learning to Discover~\cite{yuksekgonul2026learning} provides an early example, showing that models can continue learning during deployment to search for exceptional solutions to specific scientific problems, rather than merely improving average performance. However, it still focuses largely on LLM-only discovery, while broad scientific workflows often involve a wider range of models, such as diffusion models and VLMs. Advancing this direction could enable AI systems not only to improve predefined tasks, but also to assist with real scientific and engineering challenges under scarce supervision and common domain shifts.

\vspace{6pt}\noindent\textbf{TTI for Deep Research}
\revise{While TTI for scientific discovery focuses on improving candidate solutions within a predefined problem, deep research extends the same test-time improvement process to the research workflow itself.} AutoSOTA~\cite{li2026autosota} develops an auto-system that goes beyond high-level text-format solution discovery to the full research loop, including reproduction, debugging, experiment management, idea generation, and validity control. Given a research paper and its SOTA model, such a system could even discover models that outperform this SOTA. Building such a system is highly challenging, and TTI is naturally suited to support it by leveraging adaptive inference with additional test-time computation and scaling. This points to a future where TTI is used not only to improve outputs within a task, but also to discover solutions for complex scientific problems and advance SOTA models for system-level research challenges.
However, we emphasize that this direction should primarily serve as a tool to assist researchers and engineers in solving problems, rather than a mechanism for automatically generating papers for direct submission. We strongly discourage such irresponsible misuse.

\section{Conclusions}\label{sec:conclusion}

In this survey, we presented Test-Time Intelligence (TTI) as a unified perspective for understanding how AI systems improve during deployment. By organizing existing methods around \revision{state update and inference-time compute}, TTI connects test-time learning, test-time adaptation, and test-time scaling under a common framework. We reviewed representative methods across learning-based adaptation, inference-time scaling, and their intersection, and discussed applications in \revision{vision, language, multimodal learning, generative models, robotics, and healthcare}. Despite rapid progress, TTI remains an emerging area. Key challenges remain in designing reliable test-time objectives, ensuring long-horizon stability, improving efficiency, establishing realistic evaluation protocols, developing theoretical foundations, addressing safety risks, \etc~We hope this survey provides a coherent foundation and useful roadmap for future research toward adaptive, scalable, and self-improving AI systems at test time.

\bibliographystyle{abbrvnat}
\small
\setlength{\bibsep}{3pt}
\bibliography{arxiv_references}

\end{document}